\documentclass{article}

    \PassOptionsToPackage{numbers, compress}{natbib}

    \usepackage[final]{neurips_2025}
    
\usepackage[utf8]{inputenc} 
\usepackage[T1]{fontenc}    
\usepackage{hyperref}       
\usepackage{url}            
\usepackage{booktabs}       
\usepackage{amsfonts}       
\usepackage{nicefrac}       
\usepackage{microtype}      
\usepackage{xcolor}         
\usepackage[dvipsnames]{xcolor} 
\hypersetup{
  breaklinks,
  colorlinks,
  linkcolor=BlueViolet,
  citecolor=BlueViolet,  
}
\usepackage{graphicx}

\usepackage{amsthm}
\usepackage{amsmath}
\theoremstyle{plain}

\newtheorem{proposition}{Proposition}
\theoremstyle{remark}

\usepackage[capitalize]{cleveref}
\usepackage{algorithm}      
\usepackage{algpseudocode}  
\usepackage{multirow}
\usepackage{subcaption}
\usepackage{enumitem}
\usepackage{todonotes}

\newcommand{\modelname}[0]{NicheFlow}

\crefname{section}{Sec.}{Secs.}
\crefname{table}{Tab.}{Tabs.}
\crefname{appendix}{App.}{Apps.}
\crefname{algorithm}{Alg.}{Algs.}

\Crefname{table}{Table}{Tables}

\usepackage{bm}

\def\1{\bm{1}}

\def\rd{{\textnormal{d}}}

\def\vc{{\bm{c}}}

\def\vf{{\bm{f}}}

\def\vm{{\bm{m}}}

\def\vr{{\bm{r}}}

\def\vx{{\bm{x}}}

\def\vz{{\bm{z}}}

\def\vgamma{{\boldmath{\vgamma}}}

\def\vgamma{{\bm{\gamma}}}

\DeclareMathAlphabet{\mathsfit}{\encodingdefault}{\sfdefault}{m}{sl}
\SetMathAlphabet{\mathsfit}{bold}{\encodingdefault}{\sfdefault}{bx}{n}

\def\gL{{\mathcal{L}}}
\def\gM{{\mathcal{M}}}
\def\gN{{\mathcal{N}}}

\def\gU{{\mathcal{U}}}

\newcommand{\E}{\mathbb{E}}

\usepackage{wrapfig}

\title{Modeling Microenvironment Trajectories on Spatial Transcriptomics with NicheFlow}

\author{%
  Kristiyan Sakalyan\thanks{Equal contribution}\:\:$^{,1}$, Alessandro Palma\footnotemark[1]\:\:$^{,1,2,3}$, Filippo Guerranti\footnotemark[1]\:\:$^{,1,2,4}$, \\ \textbf{Fabian J. Theis}$^{1,2,3,4,5}$\textbf{, Stephan Günnemann}$^{1,2,4}$ \\
  $^1$School of Computation, Information and Technology, Technical University of Munich\\
  $^2$Munich Centre for Machine Learning (MCML) \\
  $^3$Institute of Computational Biology, Helmholtz Munich \\ 
  $^4$Munich Data Science Institute (MDSI), Technical University of Munich \\
  $^5$TUM School of Life Sciences Weihenstephan, Technical University of Munich \\
  Correspondence to: \texttt{\{k.sakalyan, a.palma, f.guerranti\}@tum.de} \\
}

\begin{document}

\maketitle

\begin{abstract}

\setcounter{footnote}{0}

Understanding the evolution of cellular microenvironments in spatiotemporal data is essential for deciphering tissue development and disease progression. While experimental techniques like spatial transcriptomics now enable high-resolution mapping of tissue organization across space and time, current methods that model cellular evolution operate at the single-cell level, overlooking the coordinated development of cellular states in a tissue. We introduce \modelname{}, a flow-based generative model that infers the temporal trajectory of cellular microenvironments across sequential spatial slides. By representing local cell neighborhoods as point clouds, \modelname{} jointly models the evolution of cell states and spatial coordinates using optimal transport and Variational Flow Matching. Our approach successfully recovers both global spatial architecture and local microenvironment composition across diverse spatiotemporal datasets, from embryonic to brain development\footnote{Project page: \url{https://www.cs.cit.tum.de/daml/nicheflow}}.

\end{abstract}
\section{Introduction}

Uncovering the principles governing tissue organization across space and time remains one of the most fundamental challenges in biology, with profound implications for evolutionary and developmental studies \citep{mayr2019exploring,zinner2020design}. 
While individual cells form the basic units of biological systems, they operate not in isolation but as integral parts of spatially organized \textit{microenvironments}, functionally distinct neighborhoods, or \textit{niches}, shaped by cell-to-cell interactions and extracellular components \citep{ren2023spatial,schaar2024nicheformer,liu2025single}. These spatial microenvironments influence crucial biological processes, from tumor progression to immune infiltration and tissue regeneration \citep{moncada2020integrating,larsson2021spatially,chen2022spatiotemporal}.

Spatial transcriptomics (ST) has transformed our ability to investigate these tissue architectures by providing single-cell resolution mapping of gene expression while preserving spatial context \citep{staahl2016visualization,rodriques2019slide,eng2019transcriptome,moses2022museum}. This technological breakthrough has enabled researchers to examine the molecular underpinnings of tissue organization with unprecedented detail. However, ST provides only static snapshots of inherently dynamic biological systems.
\textit{Time-resolved} spatial analysis extends ST by capturing how gene expression patterns and cellular arrangements evolve across developmental stages or experimental time \citep{briggs2018dynamics,wagner2018single,ren2023spatiotemporally}. This temporal dimension offers critical insights into the development of tissue organization in health and disease \citep{papanicolaou2022temporal} that cannot be inferred from static observations alone. 

Despite these technological advances, modeling trajectories on time-resolved spatial slides is complicated, as no direct correspondence exists between cells across slides due to the destructive nature of acquisition practices. Moreover, current computational methods fall short in modeling the evolution of tissue organization at the level of cellular microenvironments. Most approaches infer trajectories by modeling single-cell dynamics using velocity-based models \citep{abdelaal2024sirv,long2025spvelo,shen2025inferring} or optimal transport between individual cells \citep{klein2025mapping,bryan2025accurate}. While effective at capturing cell evolution, these cell-centric methods fundamentally miss the coordinated evolution of structured niches within tissues.

This limitation presents a critical research gap that we address with the following question:
\begin{center}
\textit{How can we model the spatiotemporal evolution of cellular microenvironments preserving both local neighborhood relationships and cellular state transitions?}
\end{center}
\begin{wrapfigure}[20]{r}{0.54\textwidth} 
    \centering
    \vspace{-6pt}
    \includegraphics[width=\linewidth]{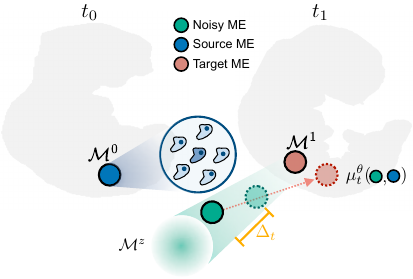}
    \caption{\textbf{Overview of \modelname{}.}  
    At time $t_1$, we generate a target microenvironment $\mathcal{M}^1$ by transforming Gaussian noise $\mathcal{M}^z$ using a Variational Flow Matching model with a posterior $\mu_t^{\theta}$ conditioned on a source microenvironment $\mathcal{M}^0$ at $t_0$. Source-target pairs are identified via entropic OT over pooled microenvironment coordinates and gene expression profiles.}
    \label{fig: main_figure}
\end{wrapfigure}

To address this question, we directly model the dynamics of cellular neighborhoods as cohesive units rather than focusing on isolated cell trajectories. This approach aligns naturally with tissue-scale biological processes and enables principled learning of dynamics in structured, high-dimensional, and variably sized spatial domains.


We introduce \textbf{Niche} \textbf{Flow} Matching (\textbf{\modelname{}}) (\cref{fig: main_figure}), a generative model for learning spatiotemporal dynamics of cellular niches from time-resolved spatial transcriptomics data. \modelname{} builds on recent advances in Flow Matching (FM) and Optimal Transport (OT) to operate over distributions of microenvironments, which we represent as point clouds. \modelname{} enables accurate modeling of global spatial architecture and local microenvironment composition within evolving tissues.

Our contributions include:
\begin{itemize}[leftmargin=*, itemsep=2pt, topsep=0pt]
    \item \textbf{A microenvironment-centered trajectory inference paradigm} that shifts from modeling individual cells in time to modeling niches as point clouds, enabling simultaneous prediction of spatial coordinates and gene expression profiles while preserving local tissue context.
    \item \textbf{A factorized Variational Flow Matching (VFM) approach} with distributional families (Laplace for spatial coordinates, Gaussian for gene expression) that jointly trains on spatial and cell state dynamics using a factorized loss, modeling spatial reconstruction and biological fidelity with tailored distributional assumptions.
    \item \textbf{A spatially-aware sampling strategy} using OT between niche representations, enabling scalable training on large tissue sections while ensuring comprehensive coverage of heterogeneous regions.
\end{itemize}
Our approach consistently outperforms baselines in recovering cell-type organization and spatial structure across embryonic, brain development, and ageing datasets. \modelname{} enables principled learning of dynamics in structured, high-dimensional, and variably sized spatial domains, a challenge with parallels in other spatiotemporal modeling domains beyond biology.

\section{Related Work}
\modelname{} is at the interface between generative models and spatiotemporal transcriptomic data. 

\textbf{FM and single-cell transcriptomics.} We propose a model based on FM, a framework introduced by several seminal works \citep{liu2023flow, lipman2023flow, albergo2023building}. Specifically, we adopt a variational view of the FM objective, following \citet{eijkelboom2024variational}, but extend it to mixed-factorized distributions for point cloud generation. Our method, \modelname{}, combines FM with OT, a pairing that has proven effective in modeling cellular data \citep{tong2024improving, pmlr-v238-tong24a, atanackovic2025meta, klein2024genot}. Unlike these models, however, we focus on point clouds of spatially-resolved transcriptomic profiles. Closest to our approach is Wasserstein FM for point cloud generation \citep{haviv2024wasserstein}, applied to reconstruct cellular niches. Yet, that work does not address joint generation of spatial coordinates and cellular states, nor OT-based temporal trajectory prediction, both of which are central to our contribution.

\textbf{Generative models for spatial transcriptomics.} Generative models have been key to spatial tasks such as gene expression prediction from histology slides \citep{zhu2025diffusion, wan2023integrating}, integration with dissociated single-cell data \citep{wan2023integrating}, spatial imputation \citep{haviv2025covariance, li2024stdiff}, and perturbation \cite{akbarnejad2025mapping}. More recently, LUNA \citep{yu2025tissue} demonstrated strong performance in predicting single-cell spatial coordinates using diffusion models \citep{ho2020denoising} conditioned on transcription data. While related, our model addresses the distinct task of inferring niche trajectories, enabling the simultaneous generation of coordinates and cellular states.

\textbf{Trajectory inference for spatial transcriptomics.} Previous work has explored learning trajectories from spatial slides. \citet{pham2023robust} proposed a graph-based spatiotemporal algorithm for pseudotime inference, while others leveraged tissue-resolved transcriptomics to estimate cell velocity \citep{abdelaal2024sirv, long2025spvelo, shen2025inferring}. Closer to our approach, \citet{klein2025mapping} and \citet{bryan2025accurate} use discrete OT to link cells across time and infer the evolution of cell states from spatially-resolved gene expression. Similarly, DeST-OT \citep{halmos2025dest} aligns spatial slides with semi-relaxed OT couplings, preserving transcriptomic and spatial proximity between ancestor and descendant cells, while SpaTrack \citep{shen2025inferring} uses Fused Gromov-Wasserstein OT \citep{vayer2020fused}, balancing transcriptomic and spatial differences based on spatial autocorrelation of features. Unlike our mini-batch deep learning model, these methods do not operate on entire microenvironments and rely on exact OT at the single-cell level, resulting in limitations in scalability and generalization.
\section{Background}\label{sec: background}
\subsection{Optimal Transport with FM}\label{sec: fm_ot}
FM~\citep{lipman2023flow} is a generative model that transforms a source density $p_0$ into a target density $p_1$. It operates by learning a time-dependent velocity field $u_t(\vx)$ for $t \in [0,1]$ and $\vx\in\mathbb{R}^D$, which generates a probability path $\{p_t\}_{t \in [0,1]}$. This path is constructed such that the marginals at time $t=0$ and $t=1$ match the source and target distributions, i.e., $p_0$ and $p_1$, respectively. The velocity field induces an Ordinary Differential Equation (ODE), whose solution $\phi_t(\vx)$ defines a \textit{flow map} that transports samples from the source to the target distribution.

In practice, FM approximates $u_t(\vx)$ with a time-conditioned neural network $v^\theta_t(\vx)$. While the exact marginal velocity field $u_t(\vx)$ is intractable, it can be expressed in terms of data-conditioned velocity fields and a joint distribution $\pi(\vx_0, \vx_1)$ over the source and target samples $\vx_0 \sim p_0$ and $\vx_1 \sim p_1$:
\begin{equation}\label{eq: marginal_field_fm}
u_t(\vx) = \int u_t(\vx \mid \vx_0, \vx_1)\, \frac{p_t(\vx \mid \vx_0, \vx_1) \pi(\vx_0, \vx_1)}{p_t(\vx)} \, \rd\vx_0 \, \rd\vx_1 \: ,
\end{equation}
where $p_t(\vx \mid \vx_0, \vx_1)$ is a pre-defined interpolating probability path. Here, we consider the tractable probability path $p_t(\vx \mid \vx_0, \vx_1) = \delta(\vx - g_t(\vx_0, \vx_1))$, where $g_t(\vx_0, \vx_1) = (1-t)\vx_0 + t\vx_1$ is a linear interpolation and $\delta$ denotes a Dirac delta function, representing a deterministic conditional path.

\citet{lipman2023flow} show that regressing the conditional field $u_t(\vx \mid \vx_0, \vx_1)$ is equivalent to learning the marginal field $u_t(\vx)$ in expectation. Hence, the FM objective becomes the task of learning the velocity along the conditional probability path between any pair of source and target data points. For a linear conditional probability path, the velocity $u_t(\vx \mid \vx_0, \vx_1)$ has a closed form, and the FM loss is:\looseness=-1
\begin{equation}
\mathcal{L}_{\textrm{FM}}(\theta) = \mathbb{E}_{t \sim \mathcal{U}[0,1],\, (\vx_0, \vx_1) \sim \pi}\left[ \left\| v^\theta_t\left(g_t(\vx_0, \vx_1)\right) - \frac{\partial}{\partial t}g_t(\vx_0, \vx_1)  \right\|^2_2 \right].
\end{equation}
In practice, the coupling $\pi(\vx_0, \vx_1)$ is instantiated using sample pairs drawn from a mini-batch estimate. When one chooses $\pi^{\star}$ as the OT coupling under a squared Euclidean cost between samples from $p_0$ and $p_1$, FM approximates the dynamic OT map between source and target densities \citep{pmlr-v202-pooladian23a, tong2024improving}. 
Thus, the solution samples $(\vx_0, \vx_1) \sim \pi^{\star}$ from the joint distribution approximately follow:
\begin{equation}
    \vx_0 \sim p_0, \: \vx_1 \sim \delta\left(\vx_1 - \phi^{\theta}_1(\vx_0)\right),
\end{equation}
where $\phi^{\theta}_t$ is the solution of the ODE with velocity field $v^{\theta}_t$.

\subsection{Generative OT on incomparable source and target spaces}\label{sec: genot}
\citet{klein2024genot} generalize the OT FM formulation to settings where the source and target distributions are defined on incomparable spaces and propose an approach to generative entropic OT using FM. Given a standard normal noise distribution with samples $\vz \sim \mathcal{N}(\mathbf{0}, \mathbf{I}_D)$, the authors show that the following sampling procedure:\looseness=-1
\begin{equation}
    \vx_0 \sim p_0, \: \vx_1 \sim \delta\left(\vx_1 - \phi^{\theta}_1(\vz \mid \vx_0)\right),
\end{equation}
defines a generative model that implicitly samples from an Entropic OT (EOT) coupling, where $\smash{\phi^{\theta}_t(\vz \mid \vx_0)}$ is a FM model that maps noise to target samples, conditioned on source points. To achieve this, $\phi^{\theta}_t$ is trained using source-target pairs $(\vx_0, \vx_1)$ drawn from the EOT coupling $\pi^{\star}_{\epsilon}$, with $\epsilon$ denoting the entropic regularization parameter, which the model aims to approximate.

Crucially, this formulation enables OT between distinct source and target spaces, \textit{as $\vx_0$ does not flow directly into $\vx_1$, but instead conditions the generation of target samples from noise}.

\subsection{Source-conditioned VFM}\label{sec: vfm}
Consider the source-conditioned FM formulation in \cref{sec: genot}. Given a conditioning source $\vx_0$ and noise-based generation, the marginal field in \cref{eq: marginal_field_fm} can be written as: 
\begin{align}
    \label{eq: source_cond_posterior}
    u_t(\vx \mid \vx_0) = \E_{p_t(\vx_1 \mid \vx, \vx_0)} \left[ u_t(\vx \mid \vx_1) \right] \: , 
\end{align}
where we drop the conditioning on $\vx_0$ in the velocity field, as \(u_t\) is entirely determined by the target $\vx_1$ when generating from noise under linear probability paths.

Since $u_t(\vx \mid \vx_1)$ is tractable \citep{lipman2023flow}, one can recast FM as a variational inference problem, following \citet{eijkelboom2024variational}, by introducing a parameterized approximation $q^{\theta}_t(\vx_1 \mid \vx, \vx_0)$ to the true posterior $p_t(\vx_1 \mid \vx, \vx_0)$. Integrating the expected velocity in \cref{eq: source_cond_posterior} over $t \in [0,1]$ enables the generation of target points $\vx_1$ from noise, conditioned on $\vx_0$.\looseness=-1

During training, the source-conditioned Variational Flow Matching (VFM) loss is:
\begin{equation}\label{eq: vfm_loss}
    \mathcal{L}_{\textrm{SC-VFM}}(\theta) = -\E_{t \sim \gU[0, 1],\, (\vx_0,\vx_1) \sim \pi^{\star}_{\epsilon},\, \vx \sim p_t(\vx \mid \vx_1)}\left[ \log q^{\theta}_t(\vx_1 \mid \vx, \vx_0) \right] \: ,
\end{equation}
where $\pi^{\star}_{\epsilon}$ is an entropic OT coupling modeling the joint distribution over source and target samples, and $p_t(\vx \mid \vx_1)$ interpolates between target samples and noise. In the generation phase, one samples $\vx_0 \sim p_0$ and noise $\vz \sim \gN(\mathbf{0}, \mathbf{I}_D)$, then simulates the marginal field in \cref{eq: source_cond_posterior} starting from $\phi_0(\vx) = \vz$ to generate a target sample $\vx_1$.

Crucially, under the assumption that $u_t(\vx \mid \vx_1)$ is linear in $\vx_1$, which holds when using straight-line interpolation paths, the marginal field in \cref{eq: source_cond_posterior} only depends on the posterior's first moment on $\vx_1$:
\begin{equation}
u_t(\vx \mid \bm{x}_0) = u_t\left(\vx \mid \mathbb{E}_{p_t(\vx_1 \mid \vx, \vx_0)}\left[ \vx_1 \right]\right)
\end{equation}
This implies that the VFM objective reduces to matching the first moment of the approximate posterior $q_t^\theta(\vx_1 \mid \vx, \vx_0)$ to that of the true posterior $p_t(\vx_1 \mid \vx, \vx_0)$. As a result, the approximate posterior can be chosen fully factorized under a \textit{mean field assumption}, since each dimension can be matched independently if the mean of the true posterior is preserved; see \cref{app: theoretical_factorized_posterior} for more details.

\section{\modelname{}}
\label{sec: nicheflow}
We introduce a flow-based generative OT model to infer the temporal evolution of spatially resolved cellular microenvironments. More specifically, given a spatial microenvironment represented as a point cloud of cell states with their coordinates, \modelname{} predicts the corresponding tissue structure at a later time point. To delineate our approach, we define a list of desiderata.

\textbf{Generative model on structured data.} Similar to prior work \citep{haviv2024wasserstein,yu2025tissue}, we consider a generative model over structured point cloud data representing cellular microenvironments. This approach implicitly accounts for spatial correlations between cells, in contrast to models that study the evolution of spatial trajectories at the single-cell level \citep{klein2025mapping}.

\textbf{Sub-regions and variable location.} Crucially, for better memory efficiency, we do not consider an entire spatial slide for trajectory inference, but instead learn the dynamics of variably located sub-regions. This design choice enables scalability and flexibility in modeling functional regions across different parts of the screened tissue.

\textbf{Changes in the number of nodes.} To model the temporal evolution and densification of microenvironments, we allow the source and target regions to differ in the number of nodes. We adopt a formulation similar to \cref{sec: genot}, implementing OT FM between non-comparable spaces.

\textbf{Flexible generative models for features and coordinates.} We allow flexibility in the choice of generative models for the features and coordinates. To this end, we implement the approach described in \cref{sec: vfm}, factorizing features and coordinates into separate posteriors from different families.

\subsection{Data description and problem statement}\label{sec: single_cell_microenv}
We are given a sequence of time-resolved spatial transcriptomic measurements across biological processes such as development or ageing. For simplicity, we formulate our problem in terms of two consecutive time points indexed by $s$, such that $s \in \{0,1\}$, though the model can be extended to collections of more consecutive discrete temporal measurements. Each dataset at a time point is a full tissue slide that can be represented as an \textit{attributed point cloud}:
\begin{equation}
    \mathcal{P}_s = \{ (\vc_i^s, \vx_i^s) \mid i = 1, \dots, N_s \},
\end{equation}
where $\vc_i^s \in \mathbb{R}^2$ denotes the 2D spatial coordinate of cell $i$ at time $s$, and $\vx_i^s \in \mathbb{R}^D$ denotes its associated feature vector, typically corresponding to its gene expression profile or a low-dimensional representation thereof. Thus, each dataset is an \textit{attributed point set} in two-dimensional space. Note that each slide can have a variable number of cells, and no direct correspondence exists between single cells across subsequent slides. 

To capture spatial context beyond individual cells, we define local \textit{microenvironments} as fixed-radius neighborhoods. Specifically, for each cell $(\vc_i^s, \vx_i^s)$ at time $s$, we construct a neighborhood $\mathcal{M}_i^s$ consisting of all neighboring cells within a spatial radius $r$:
\begin{equation}
    \mathcal{M}_i^s = \left\{ (\vc_j^s, \vx_j^s) \,\middle|\, \|\vc_j^s - \vc_i^s\| \leq r \right\}.
\end{equation}
Let $\smash{\{\gM_i^0\}_{i=1}^{N_0}}$ and $\smash{\{\gM_j^1\}_{j=1}^{N_1}}$ be collections of source and target microenvironments at consecutive time points. Our goal is to train a parameterized flow model $\phi_t^\theta$, with $t \in [0,1]$, that generates target microenvironments conditioned on source point clouds. Specifically, to sample a target microenvironment with $k$ cells conditioned on a variably sized source $\gM^0$, we define sampling as:
\begin{align}
    \gM^z &= \left\{(\vc_i^z, \vz_i) \mid \vc_i^z \sim \gN(\mathbf{0}, \mathbf{I}_2),\ \vz_i \sim \gN(\mathbf{0}, \mathbf{I}_D),\ \forall i = 1,\dots,k \right\}, \\
    \gM^1 &= \phi_1^\theta(\gM^z \mid \gM^0),\label{eq: pc_sampling}
\end{align}
where $\gM^z$ is a point cloud composed of noisy coordinates and features, and $\gM^1$ is a generated prediction for the evolution of $\gM^0$ at the next time point. As explained in \cref{sec: fm_ot} and \cref{sec: genot}, we want our generative model to parameterize some notion of optimal entropic coupling $\pi^\star_{\epsilon}$ between microenvironments across slides (see \cref{sec: pc_ot}).

\subsection{OT formulation}\label{sec: pc_ot}
To train the flow map $\phi_t^{\theta}$ to perform conditional EOT, we define a cost function that induces an optimal entropic coupling $\pi^\star_{\epsilon}$ between source and target point clouds. This coupling is used to sample pairs of source and target microenvironment mini-batches during training. While there is no established notion of optimal cost in this setting, we propose to compute OT using source and target microenvironment representations based on the weighted average of features and coordinates.

Specifically, we compute a pooled representation for each microenvironment in the source and target slides by averaging spatial coordinates and gene expression features, weighted by a tunable hyperparameter $\lambda \in [0, 1]$ that balances spatial versus cellular state information:
\begin{equation}
    \label{eq: ot-weighted}
    \bar{\vm}_i^s = \left[ 
        \frac{1 - \lambda}{|\mathcal{M}_i^s|} \sum_{(\vc_j^s, \vx_j^s) \in \mathcal{M}_i^s} \vc_j^s \quad \Big\| \quad
        \frac{\lambda}{|\mathcal{M}_i^s|} \sum_{(\vc_j^s, \vx_j^s) \in \mathcal{M}_i^s} \vx_j^s
    \right],
\end{equation}
where $\|$ denotes concatenation, $\bar{\vm}_i^s \in \mathbb{R}^{2 + D}$, and $s \in \{0, 1\}$. We then apply EOT on the sets $\smash{\{\bar{\vm}_i^0\}_{i=1}^{N_0}}$ and $\smash{\{\bar{\vm}_j^1\}_{j=1}^{N_1}}$ using a squared Euclidean cost and regularization parameter $\epsilon$, yielding the coupling $\pi^\star_{\epsilon, \lambda}$. During training, we sample matched pairs $(\mathcal{M}^0, \mathcal{M}^1) \sim \pi^\star_{\epsilon, \lambda}$ computed over mini-batches, and use them as supervision for learning the conditional generative model. A higher value of $\lambda$ prioritizes feature similarity, while lower values favor proximity in coordinates (see \cref{fig: moscot-ot-lambdas}).

\subsection{Mixed-factorized VFM}\label{sec: mixture_vfm}
Once we have established a strategy for performing mini-batch OT, we proceed to describe our approach for learning the flow model $\phi^{\theta}_t$ and simulating \cref{eq: pc_sampling}. To this end, we adopt a variant of VFM (\cref{sec: vfm}), originally developed for graph generation, and adapt it to our point cloud setting.

In line with \cref{sec: genot} and \ref{sec: vfm}, we delineate an objective to train a parameterized, source-conditioned posterior over target point clouds $q^{\theta}_t(\gM^{1} \mid \gM, \gM^{0})$, where $\gM^0$ and $\gM^1$ represent source and target niches, and $\gM$ denotes a noisy point cloud at interpolation time $t$. Importantly, our posterior comes with the following characteristics: \textbf{(i)} the posterior is factorized across the single points in a point cloud; \textbf{(ii)} the posterior is factorized across cellular features and coordinate dimensions; and \textbf{(iii)} the family of posteriors can be chosen differently between cellular features and coordinates.

Following \textbf{(i)}, we model the variational distribution over \( \mathcal{M}^1 \) by factorizing it across individual points \( (\vc_1, \vx_1) \in \mathcal{M}^1 \). Moreover, we tackle \textbf{(ii)} and \textbf{(iii)} using the mean-field VFM assumption, modeling cellular state and positions separately (see \cref{app: theoretical_vfm} and \ref{app: theoretical_factorized_posterior} for theoretical justifications):
\begin{equation}
    q_t^{\theta} (\mathcal{M}^1 \mid \mathcal{M}, \mathcal{M}^0)
    = \prod_{(\vc_1, \vx_1) \in \mathcal{M}^1} 
    \left(\prod_{k = 1}^{2}  f_t^{\theta} (\ c_1^k \mid \mathcal{M}, \mathcal{M}^0) \prod_{d = 1}^{D} r_t^{\theta} (x_1^d \mid \mathcal{M}, \mathcal{M}^0) \right) \: .
\end{equation}
Here, $f^{\theta}_t$ and $r^{\theta}_t$ denote distinct approximate posterior families for cellular states ($x_1^d$) and spatial positions ($\smash{c_1^k}$), respectively. We use a Laplace distribution for $\smash{f^{\theta}_t}$ due to its concentration around the mean, which supports precise modeling of coordinate features, while a Gaussian distribution is used for $r^{\theta}_t$.\looseness=-1

As explained in \cref{sec: vfm}, if one uses FM with straight probability paths, only the first moment of $q_t^{\theta}$ is required to simulate the generative field in \cref{eq: source_cond_posterior}. Therefore, the posterior is replaced by a time-dependent predictor $\smash{(\bar{\vf}_t^{\theta}, \bar{\vr}_t^{\theta}) = \mu_t^{\theta}(\mathcal{M}, \mathcal{M}^0)}$ of the mean features $\smash{\bar{\vr}_t^{\theta}}$ and coordinates $\smash{\bar{\vf}_t^{\theta}}$ of the target point clouds, learned minimizing the following loss:\looseness=-1
\begin{align}\label{eq: stcmflow_obj}
    &\mathcal{L}_{\text{\modelname{}}}(\theta) =\ 
    \mathop{\mathbb{E}}\limits_{\substack{
        t \sim \gU[0, 1] \\
        (\mathcal{M}^0, \mathcal{M}^1) \sim \pi^\star_{\epsilon, \lambda} \\
        \mathcal{M} \sim p_t(\mathcal{M} \mid \mathcal{M}^1)
    }} \left[ \sum_{(\vc_1, \vx_1) \in \mathcal{M}^1}
     \Big( \| \vc_1 - \bar{\vf}_t^{\theta} \|_1 
    + \frac{1}{2} \| \vx_1 - \bar{\vr}_t^{\theta} \|_2^2 \Big) \right]  \: .
\end{align}
We derive the objective in \cref{app: objective_derivation} and provide algorithms in \cref{appendix: algorithms}. Here, $\mu_t^{\theta}$ inputs a noisy point cloud $\gM$ and a source $\gM^0$ and provides a mean prediction vector for coordinates and feature dimensions, respectively indicated as $\bar{\vf}_t^{\theta}$ and $\bar{\vr}_t^{\theta}$. We implement it as a point cloud transformer (see \cref{sec: pc_transformer}).

We highlight a crucial aspect about \cref{eq: stcmflow_obj}. While the single dimensions are fully factorized in the predictions from $\mu_t^{\theta}$, every feature's mean is a function of the whole noisy point cloud $\gM$ as well as the target microenvironment $\gM^0$. In other words, the predictions exploit structural information in the point cloud to predict the individual posterior mean of each dimension. Like most approaches, our method has modeling limitations that we outline in \cref{app: limitations}.

\subsection{Backbone architecture: Microenvironment transformer}\label{sec: pc_transformer}

To parameterize the conditional posterior mean $\mu_t^{\theta}$ from \cref{sec: mixture_vfm}, we use a \textit{permutation-invariant} transformer architecture designed for point clouds with variable size.

\textbf{Encoder--decoder structure.} The model follows an encoder-decoder layout, processing the source microenvironment \( \gM^0 \) via the encoder and predicting the posterior mean from a noisy target \( \gM \sim p_t(\cdot \mid \gM^1) \) via the decoder conditioned on the encoder's output.\looseness=-1 

\textbf{Input embeddings.} Each point is represented by features \( \vx \in \mathbb{R}^D \) and spatial coordinates \( \vc \in \mathbb{R}^2 \), embedded separately and concatenated. Time \( t \) is encoded using sinusoidal embeddings, linearly projected and broadcast across points.

\textbf{Cross-attention conditioning.} The encoder processes the embedded source $\gM^0$ microenvironment using self-attention. The decoder operates on the noisy target $\gM$ using self-attention, followed by cross-attention to condition on the encoder's outputs. The cross-attention mechanism allows each target point to attend to all source points.

\textbf{Output projection.} The decoder outputs are linearly projected to yield posterior mean estimates $(\bar{\vf}_t^{\theta}, \bar{\vr}_t^{\theta})$ for expression and coordinates.
\section{Experiments}
We propose quantitative and qualitative evaluations of our algorithm. Quantitatively, we test whether source-conditioned samples generated by our model preserve the biological structure and shape of future tissue states. Qualitatively, we demonstrate that our approach accurately captures compositional shifts in substructural components and developmental trajectories across time. 

\begin{table}[t]
    \caption{
    Performance comparison across three biological datasets for SPFlow, RPCFlow and \modelname{}. Models are trained using the Conditional Flow Matching (CFM), Gaussian VFM (GVFM), or Gaussian-Laplacian VFM (GLVFM) strategies. Results are reported as mean ± standard deviation over five evaluation runs on mouse embryonic development (MED), axolotl brain development (ABD), and mouse brain aging (MBA). For all experiments, we use a fixed value of $\lambda=0.1$, enabling spatial location preservation across time.
    }
    \label{table: main-results}
    \centering
    \resizebox{\textwidth}{!}{%
    \begin{tabular}{ll
        ccc
        ccc
        ccc
    }
        \toprule
        \multicolumn{2}{l}{} &
        \multicolumn{3}{c}{MED} &
        \multicolumn{3}{c}{ABD} &
        \multicolumn{3}{c}{MBA} \\
        \cmidrule(lr){3-5} \cmidrule(lr){6-8} \cmidrule(lr){9-11}
        Model & Obj. & 1NN-F1~$\uparrow$ & PSD~$\downarrow$ ($10^2$) &  SPD~$\downarrow$ ($10^2$) & 1NN-F1~$\uparrow$ & PSD~$\downarrow$ ($10^2$) &  SPD~$\downarrow$ ($10^2$) & 1NN-F1~$\uparrow$ & PSD~$\downarrow$ ($10^2$) &  SPD~$\downarrow$ ($10^2$) \\
        \midrule
        LUNA        & —         &  0.540 {\scriptsize{± 0.004}} & — & — & 0.331 {\scriptsize{± 0.003}} & — & — & 0.222 {\scriptsize{± 0.000}} & — & — \\
        \midrule
        SPFlow & CFM & 0.272 {\scriptsize{± 0.0011}} & 1.681 {\scriptsize{± 0.0087}} & 0.602 {\scriptsize{± 0.0013}} & 0.190 {\scriptsize{± 0.0005}} & 2.494 {\scriptsize{± 0.0051}} & 1.119 {\scriptsize{± 0.0037}} & 0.205 {\scriptsize{± 0.0003}} & 1.836 {\scriptsize{± 0.0022}} & 0.824 {\scriptsize{± 0.0006}} \\
        SPFlow & GVFM & 0.259 {\scriptsize{± 0.0009}} & 2.383 {\scriptsize{± 0.0082}} & 0.582 {\scriptsize{± 0.0009}} & 0.175 {\scriptsize{± 0.0010}} & 3.373 {\scriptsize{± 0.0103}} & 1.104 {\scriptsize{± 0.0023}} & 0.181 {\scriptsize{± 0.0001}} & 2.585 {\scriptsize{± 0.0029}} & 0.834 {\scriptsize{± 0.0011}} \\
        SPFlow & GLVFM & 0.251 {\scriptsize{± 0.0008}} & 2.249 {\scriptsize{± 0.0114}} & 0.592 {\scriptsize{± 0.0015}} & 0.173 {\scriptsize{± 0.0013}} & 2.870 {\scriptsize{± 0.0238}} & 1.093 {\scriptsize{± 0.0037}} & 0.195 {\scriptsize{± 0.0005}} & 2.320 {\scriptsize{± 0.0009}} & 0.853 {\scriptsize{± 0.0008}} \\
        
        \midrule

        RPCFlow & CFM & 0.546 {\scriptsize{± 0.0012}} & 0.981 {\scriptsize{± 0.0024}} & 0.564 {\scriptsize{± 0.0015}} & 0.524 {\scriptsize{± 0.0020}} & \textbf{2.051 {\scriptsize{± 0.0039}}} & 1.015 {\scriptsize{± 0.0036}} & 0.271 {\scriptsize{± 0.0004}} & \textbf{1.543 {\scriptsize{± 0.0016}}} & 0.810 {\scriptsize{± 0.0010}} \\
        RPCFlow & GVFM & 0.503 {\scriptsize{± 0.0013}} & 1.155 {\scriptsize{± 0.0044}} & 0.578 {\scriptsize{± 0.0007}} & 0.477 {\scriptsize{± 0.0008}} & 2.260 {\scriptsize{± 0.0077}} & 1.036 {\scriptsize{± 0.0031}} & 0.249 {\scriptsize{± 0.0003}} & 1.753 {\scriptsize{± 0.0020}} & 0.784 {\scriptsize{± 0.0010}} \\
        RPCFlow & GLVFM & 0.586 {\scriptsize{± 0.0016}} & 0.979 {\scriptsize{± 0.0021}} & 0.586 {\scriptsize{± 0.0012}} & 0.554 {\scriptsize{± 0.0007}} & 2.053 {\scriptsize{± 0.0044}} & 1.038 {\scriptsize{± 0.0025}} & 0.265 {\scriptsize{± 0.0004}} & 1.723 {\scriptsize{± 0.0015}} & 0.779 {\scriptsize{± 0.0011}} \\
        
        \midrule

        NicheFlow & CFM & 0.609 {\scriptsize{± 0.0030}} & 0.979 {\scriptsize{± 0.0228}} & 0.402 {\scriptsize{± 0.0036}} & 0.604 {\scriptsize{± 0.0018}} & 2.086 {\scriptsize{± 0.0058}} & \textbf{0.568 {\scriptsize{± 0.0030}}} & 0.283 {\scriptsize{± 0.0003}} & 1.557 {\scriptsize{± 0.0014}} & 0.556 {\scriptsize{± 0.0028}} \\
        NicheFlow & GVFM & 0.596 {\scriptsize{± 0.0027}} & 0.991 {\scriptsize{± 0.0137}} & 0.406 {\scriptsize{± 0.0025}} & 0.574 {\scriptsize{± 0.0015}} & 2.220 {\scriptsize{± 0.0107}} & 0.594 {\scriptsize{± 0.0046}} & 0.268 {\scriptsize{± 0.0003}} & 1.661 {\scriptsize{± 0.0033}} & \textbf{0.531 {\scriptsize{± 0.0010}}} \\
        NicheFlow & GLVFM & \textbf{0.664 {\scriptsize{± 0.0014}}} & \textbf{0.883 {\scriptsize{± 0.0094}}} & \textbf{0.398 {\scriptsize{± 0.0023}}} & \textbf{0.628 {\scriptsize{± 0.0013}}} & 2.079 {\scriptsize{± 0.0043}} & 0.576 {\scriptsize{± 0.0055}} & \textbf{0.285 {\scriptsize{± 0.0003}}} & 1.554 {\scriptsize{± 0.0021}} & 0.532 {\scriptsize{± 0.0009}} \\
        \bottomrule
       
    \end{tabular}%
    }
\end{table}

\subsection{Quantitative evaluation}\label{sec: quant_eval}
Our first research question is to assess the impact of two core modeling choices in \modelname{}: \textbf{(i)} learning trajectories over spatial microenvironments, rather than independently for each cell and \textbf{(ii)} restricting source and target point clouds to spatially co-localized neighborhoods of cells, instead of sampling them randomly across the slide.

We use \modelname{} and baseline FM approaches that do not incorporate \textbf{(i)} and \textbf{(ii)} to simulate spatial trajectories conditioned on early time point observations. Assuming that spatial arrangements and the biological composition at later stages evolve from earlier slides, the global correspondence between predicted and true slides indicates the quality of the generative trajectory. 

\subsubsection{Training setup}\label{sec: experimental-setup}

\textbf{Datasets.} We assess model performance across three spatiotemporal datasets: \textbf{(i)} Mouse embryogenesis \citep{klein2025mapping, chen2022spatiotemporal} and \textbf{(ii)} the axolotl brain development \citep{axolotl}, two Stereo-seq datasets profiling the spatially-resolved cellular development of a mouse embryo and axolotl brain across three (E9.5, E10.5 and E11.5) and five time points, respectively. We also consider the \textbf{(iii)} mouse brain ageing dataset \citep{Sun2025aging}, profiled with MERFISH \citep{chen2015spatially} across twenty time points (see \cref{appendix: dataset_description,appendix: dataset-preprocessing}). 

\textbf{Dataset construction.} For each dataset and time point $s \in \mathcal{S}$, we construct a set of cellular microenvironments by applying the fixed-radius neighborhood definition introduced in \cref{sec: single_cell_microenv}. Each microenvironment \( \mathcal{M}_i^s \) is centered at cell \( i \) and contains all cells within a fixed radius \( r \). This results in:\looseness=-1
$$
\boldsymbol{\mathcal{M}}^s := \{ \mathcal{M}_i^s \mid i = 1, \dots, N_s \},
$$
where \( N_s \) is the number of cells in the tissue at time \( s \), and each \( \mathcal{M}_i^s \) is an attributed point cloud encoding both spatial and gene expression information. We standardize coordinates for cross-time comparability and reduce the normalized gene expression to its top 50 Principal Components (PC).

\textbf{Batching.} We train \modelname{} with mini-batches of source and target cellular point clouds. To ensure spatial diversity during training, we sample individual batches uniformly from within discrete regions of the slides computed with $K$-Means clustering over the 2D coordinates (see \cref{fig: kmeans-regions} for a visualization with different $K$ values). From these regions, we collect $M$ source and target microenvironments and resample $N \leq M$ matching pairs from the entropic OT coupling $(\smash{{\boldsymbol{\mathcal{M}}}^{0}, {\boldsymbol{\mathcal{M}}}^{1}) \sim \pi^\star_{\epsilon, \lambda}}$ as described in \cref{sec: fm_ot}, where \( \smash{\boldsymbol{\mathcal{M}}^{0}} \) and \( \smash{\boldsymbol{\mathcal{M}}^{1}} \) denote the sampled sets (see \cref{sec: pc_ot} for details on our OT coupling).

\textbf{Evaluation data.} For consistent and reproducible evaluation, we discretize each tissue into a fixed 2D grid and define evaluation microenvironments as fixed-radius neighborhoods around the nearest cells to each grid point. This guarantees full spatial coverage and ensures deterministic comparison across methods. See \cref{appendix: discretized-microenvironments} for details.

\textbf{Multiple time-point.} \modelname{} predicts piecewise trajectories between subsequent time points. Instead of learning one flow for each couple of subsequent slides, we train a single model with additional conditioning on source and target labels (see \cref{appendix: microenvironment-transformer}).

\subsubsection{Quantitative evaluation metrics}

\textbf{Spatial structure.} We quantify coordinate generation accuracy using two asymmetric distance metrics. The \textit{point-to-shape distance} (PSD) measures how far predicted coordinates deviate from the true structure by computing the mean squared distance from each generated point to its nearest ground truth counterpart. In contrast, the \textit{shape-to-point distance} (SPD) evaluates how well the generated points cover the target region by averaging the squared distance from each ground truth point to its nearest generated point (see \cref{app: quantitative_evaluation_metrics}
 for a mathematical formulation of the metrics).

\textbf{Cell-type organization.} To assess how well the model reconstructs the spatial organization of different cell types, we use a \textit{1-nearest-neighbor} (1NN) classification setup. Since the model generates only gene expression profiles and spatial coordinates, we assign cell type labels to generated cells using a classifier trained on ground truth gene expression data (see \cref{appendix: cell-type-classifier}). Each predicted cell is then matched to its nearest real cell, and we report the weighted F1 score (1NN-F1).

\subsubsection{Models and results}
\begin{figure}[t]
    \centering
    \includegraphics[width=1\linewidth]{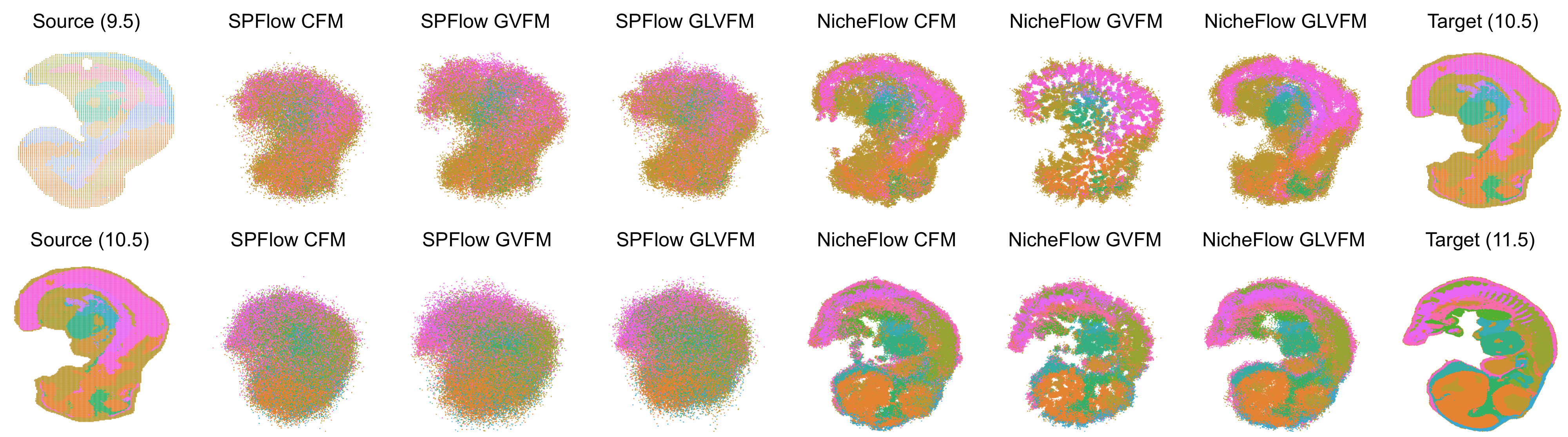}
    \caption{Qualitative comparison of generated samples on the embryonic development dataset (9.5-11.5 days). We show source and target samples alongside predictions from SPFlow and \modelname{} with different objectives.
}
    \label{fig: moscot-figure}
\end{figure}

\textbf{Baselines.} We compare against what we call \textit{SPFlow} (Single-Point Flow), a standard FM-based model that predicts temporal trajectories across slides at a single-cell level using an MLP-based velocity field. We also consider \textit{RPCFlow} (Random Point Cloud Flow), which has the same backbone as \modelname{}, but conditions on randomly sampled point clouds instead of radius neighborhoods. Additionally, we include \textit{LUNA} \citep{yu2025tissue}, a diffusion model for spatial reconstruction from dissociated cells. Note that LUNA does not model temporal dynamics and only generates coordinates from noise with their respective biological annotations. Therefore, we use it as a reference for spatial generation accuracy via the 1NN-F1 metric rather than a proper baseline.

\textbf{Ablations.} We assess different training objectives by comparing standard Conditional Flow Matching (CFM) \cite{tong2024improving} with two variational formulations modeling posteriors over coordinates and features: Gaussian-only (GVFM) and Gaussian-Laplace (GLVFM). The former uses Gaussian posteriors for coordinates and features. The latter uses the factorized formulation in \cref{sec: mixture_vfm}. For the point-cloud-based methods, we use a fixed value of $\lambda=0.1$ and sampled batches of $64$ regions chosen from the ablations in \cref{appendix: ot-ablation-study} and \ref{appendix: kmeans-ablation-study}. We present results with a single value for $\lambda$ as this experiment compares different models on reconstructing fixed tissue structures over time, for which we enforce spatial preservation (see \cref{sec: pc_ot}). However, we also include model comparisons across multiple values of $\lambda$ in \cref{table: lambda-full-ablation} for completion.

\textbf{Results.} Our quantitative evaluation (\cref{table: main-results}) demonstrates that \textit{\modelname{}} trained with the GLVFM objective consistently achieves strong performance across both spatial and semantic metrics. It outperforms all baselines in reconstructing spatial structure (PSD, SPD) and cell-type organization (1NN-F1) on developmental datasets, while remaining competitive on ageing data. These results highlight the importance of structured microenvironment modeling for capturing the spatiotemporal dynamics of complex tissues. We complement our quantitative results visually in \cref{fig: moscot-figure} and \cref{fig: axolotl-figure} and \ref{fig: aging-figure} in the Appendix, where we show that SPFlow fails to capture tissue-level organization, producing blurry and spatially incoherent samples. In contrast, \modelname{} generates predictions that preserve spatial structure and cell-type organization, despite learning only from local microenvironments. In \cref{appendix: conditioanl-generation} and \cref{appendix: stcmflow-vs-rpcflow}, we additionally demonstrate that \modelname{} produces conditionally consistent outputs with the source, while RPCFlow generates very diffused mappings across the slide, failing to preserve spatial consistency across time. 

\subsection{Qualitative evaluation and biological analysis}\label{sec: bio_eval}
We explore the capabilities of \modelname{} on the spatial trajectory inference task through qualitative and biological assessments. Specifically, we focus on validating whether our model captures compositional changes within fixed spatial structures and developmental trajectories.

\textbf{Experimental setup.} We train the model as described in \cref{sec: experimental-setup}. Using the mouse embryonic dataset \cite{chen2022spatiotemporal}, we select specific microenvironments as source niches for which we want to study the trajectory over time. Specifically, we propose two scenarios for the application of \modelname{} depending on the choice of the OT parameter $\lambda$ (see \cref{sec: pc_ot}):
\begin{enumerate}[label=\arabic*., itemsep=0pt, topsep=1pt, leftmargin=12pt]
    \item \textit{Compositional changes in fixed structures across time}. We choose the evolution of the \textit{spinal cord} of the embryo from E10.5 to E11.5 as an example and set a low value of $\lambda$ to prioritize the preservation of the spatial location in the trajectory.
    \item \textit{Spatial and cellular development of immature cells}. Developing cells may displace to different areas of the embryo, requiring a higher value of $\lambda$ to account for gene expression. As a case study, we inspect how neural crest cells in the head evolve into mesenchymal and cranial structures.
\end{enumerate}

\begin{figure}[t]
    \centering
    \includegraphics[width=1\linewidth]{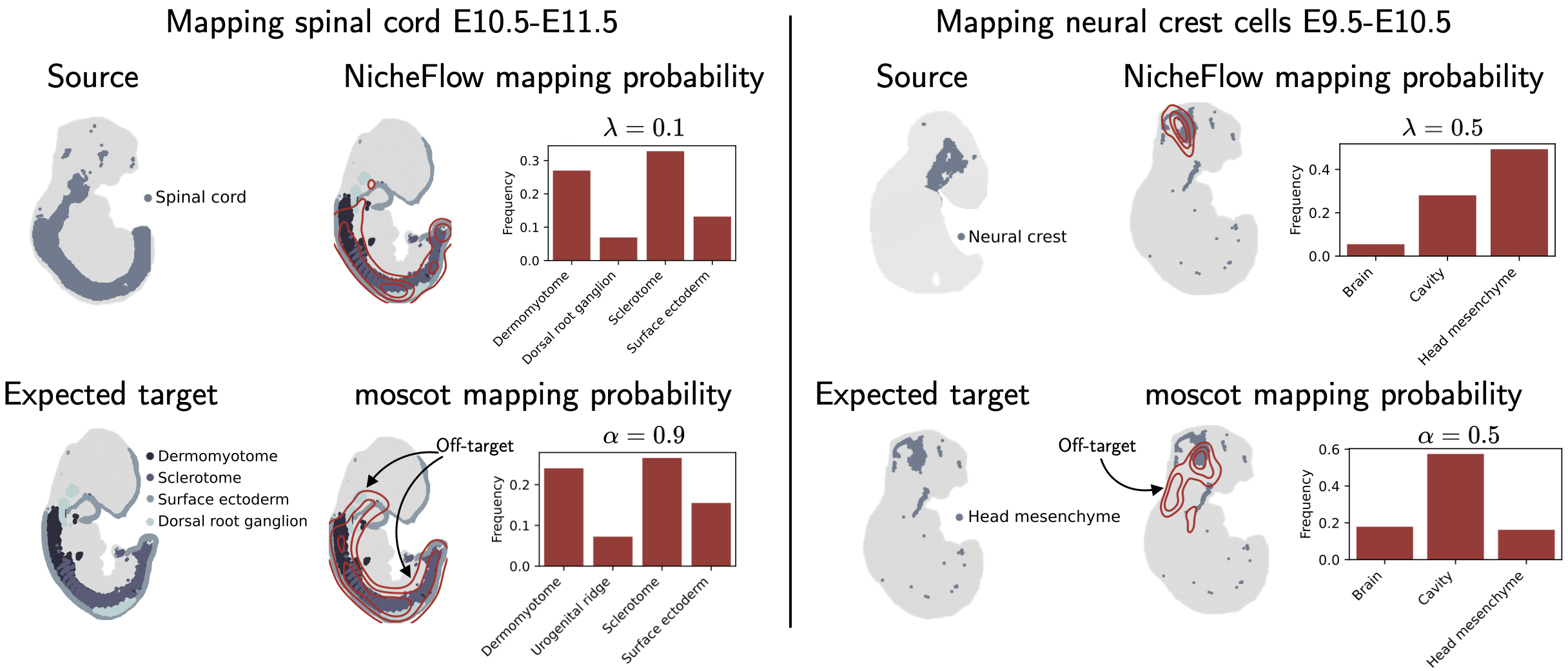}
    \caption{Left and right panels show the mapping of spinal cord (E10.5 to E11.5) and head neural crest cells (E9.5 to E10.5). In each panel, the left column shows source cells and expected targets, and the right column shows density contours of the most likely mapped regions. Bar plots display transition probabilities to the most likely descendant cell types. For \modelname{}, contours represent the proportion of samples in generated point clouds assigned to real cell coordinates across 10 samples.}

    \label{fig: bio_analysis}
\end{figure}

\textbf{Baseline.} We compare \modelname{} with the spatiotemporal framework in moscot \citep{klein2025mapping}, which models spatial trajectories at the single-cell level. In contrast to \modelname{}, moscot integrates spatial coordinates directly into the OT formulation using a Fused Gromov-Wasserstein cost \citep{vayer2019optimal}, where the hyperparameter~$\alpha$ controls the trade-off between spatial and feature-based distances (see \cref{app: biological experiments}). A high~$\alpha$ increases the influence of spatial distances, whereas in our framework this role is played by the hyperparameter~$\lambda$ (see \cref{sec: pc_ot}). We provide more details on the selection of the hyperparameter $\alpha$ for our experiments in \cref{sec: moscot_sweep}. Notably, moscot relies on exact OT and learns a transition matrix between source and target samples. As such, it is not a generative model over point clouds like \modelname{}. However, given the overlap in downstream tasks, we consider the comparison relevant.

\textbf{Evaluation and results.} For both scenarios~(1) and~(2), we have prior knowledge of the ground truth regions that the source microenvironments are expected to occupy at later developmental stages, as well as their corresponding biological compositions. For both moscot and \modelname{}, we assess whether the transported mass of source samples concentrates within the correct anatomical region at the target time point, and whether the predicted descendant cell types are biologically consistent (see \cref{app: biological experiments}). Results are summarized in \cref{fig: bio_analysis}. When modeling the evolution of the spinal cord, moscot assigns considerable mass to unrelated regions such as the urogenital ridge and branchial arches, whereas \modelname{} correctly maps source niches to the maturing spinal cord. Similarly, \modelname{} captures the differentiation of neural crest cells into mesenchymal and cranial tissues within the head region, while moscot exhibits substantial off-target leakage towards lower regions.

\section{Conclusions and Discussion}
\label{sec: conclusions}
We introduce \modelname{}, a point-cloud-based generative model designed to capture the spatiotemporal dynamics of cellular niches in time-resolved spatial transcriptomics data. Unlike methods that model single-cell trajectories independently, \modelname{} implicitly captures spatial correlations between cells by learning trajectories on variably sized local neighborhoods. To this end, we combine OT with a new version of VFM that factorizes features and coordinates into distinct posteriors from different distribution families. We showed that \modelname{} outperforms standard FM approaches at reconstructing spatial context from previous time points and improves the mapping of biological structures in time over established exact OT approaches. With the expected increase in the volume and quality of spatial data, modeling coordinated cellular state translations with generative models is a promising avenue for generalization beyond spatiotemporal inference. We envision that models like \modelname{} will enable spatial perturbation prediction and modality translation tasks requiring principled parameterized maps beyond discrete OT to extrapolate and drive biological hypotheses. 
    
\section*{Acknowledgments}
A.P. is supported by the Helmholtz Association under the joint research school Munich School for Data Science (MUDS). A.P., F.G., S.G. and F.J.T. also acknowledge support from the German Federal Ministry of Education and Research (BMFTR) through grant numbers 031L0289A and 031L0289C. F.J.T. acknowledges support from the Helmholtz Association’s Initiative and Networking Fund via the CausalCellDynamics project (grant number Interlabs-0029) and the European Union (ERC, DeepCell, grant number 101054957). The authors of this work take full responsibility for its content.

\bibliographystyle{unsrtnat}
\bibliography{bibliography}

@article{briggs2018dynamics,
    title = {The dynamics of gene expression in vertebrate embryogenesis at single-cell resolution},
    volume = {360},
    number = {6392},
    journal = {Science},
    publisher = {American Association for the Advancement of Science (AAAS)},
    author = {Briggs,  James A. and Weinreb,  Caleb and Wagner,  Daniel E. and Megason,  Sean and Peshkin,  Leonid and Kirschner,  Marc W. and Klein,  Allon M.},
    year = {2018},
    month = jun 
}

@article{wagner2018single,
    title = {Single-cell mapping of gene expression landscapes and lineage in the zebrafish embryo},
    volume = {360},
    number = {6392},
    journal = {Science},
    publisher = {American Association for the Advancement of Science (AAAS)},
    author = {Wagner,  Daniel E. and Weinreb,  Caleb and Collins,  Zach M. and Briggs,  James A. and Megason,  Sean G. and Klein,  Allon M.},
    year = {2018},
    month = jun,
}

@article{larsson2021spatially,
    title = {Spatially resolved transcriptomics adds a new dimension to genomics},
    volume = {18},
    number = {1},
    journal = {Nature Methods},
    publisher = {Springer Science and Business Media LLC},
    author = {Larsson,  Ludvig and Frisén,  Jonas and Lundeberg,  Joakim},
    year = {2021},
    month = jan,
    pages = {15–18}
}

@article{liu2025single,
    title = {Single-cell and spatial transcriptomic profiling revealed niche interactions sustaining growth of endometriotic lesions},
    volume = {5},
    number = {1},
    journal = {Cell Genomics},
    publisher = {Elsevier BV},
    author = {Liu,  Song and Li,  Xiaoyan and Gu,  Zhiyue and Wu,  Jiayu and Jia,  Shuangzheng and Shi,  Jinghua and Dai,  Yi and Wu,  Yushi and Yan,  Hailan and Zhang,  Jing and You,  Yan and Xue,  Xiaowei and Liu,  Lulu and Lang,  Jinghe and Wang,  Xiaoyue and Leng,  Jinhua},
    year = {2025},
    month = jan,
    pages = {100737}
}

@article{ren2023spatial,
    title = {Spatial transcriptomics reveals niche-specific enrichment and vulnerabilities of radial glial stem-like cells in malignant gliomas},
    volume = {14},
    number = {1},
    journal = {Nature Communications},
    publisher = {Springer Science and Business Media LLC},
    author = {Ren,  Yanming and Huang,  Zongyao and Zhou,  Lingling and Xiao,  Peng and Song,  Junwei and He,  Ping and Xie,  Chuanxing and Zhou,  Ran and Li,  Menghan and Dong,  Xiangqun and Mao,  Qing and You,  Chao and Xu,  Jianguo and Liu,  Yanhui and Lan,  Zhigang and Zhang,  Tiejun and Gan,  Qi and Yang,  Yuan and Chen,  Tengyun and Huang,  Bowen and Yang,  Xiang and Xiao,  Anqi and Ou,  Yun and Su,  Zhengzheng and Chen,  Lu and Zhang,  Yan and Ju,  Yan and Zhang,  Yuekang and Wang,  Yuan},
    year = {2023},
    month = feb 
}

@article{mayr2019exploring,
    title={Exploring single cells in space and time during tissue development, homeostasis and regeneration},
    author={Mayr, Urs and Serra, Denise and Liberali, Prisca and Klein, Allon and Treutlein, Barbara},
    journal={Development},
    volume={146},
    number={12},
    year={2019},
    publisher={The Company of Biologists}
}

@article{zinner2020design,
    title={Design principles of tissue organisation: How single cells coordinate across scales},
    author={Zinner, Marietta and Lukonin, Ilya and Liberali, Prisca},
    journal={Current opinion in cell biology},
    volume={67},
    pages={37--45},
    year={2020},
    publisher={Elsevier}
}

@article{schaar2024nicheformer,
    title={Nicheformer: a foundation model for single-cell and spatial omics},
    author = {Schaar,  Anna Christina and Tejada-Lapuerta,  Alejandro and Palla,  Giovanni and Gutgesell,  Robert and Halle,  Lennard and Minaeva,  Mariia and Vornholz,  Larsen and Dony,  Leander and Drummer,  Francesca and Bahrami,  Mojtaba and Theis,  Fabian J},
    journal={bioRxiv},
    pages={2024--04},
    year={2024},
    month = apr, 
    publisher={Cold Spring Harbor Laboratory}
}

@article{ren2023spatiotemporally, 
    title = {{Spatiotemporally resolved transcriptomics reveals the subcellular RNA kinetic landscape}},
    volume = {20},
    number = {5},
    journal = {Nature Methods},
    publisher = {Springer Science and Business Media LLC},
    author = {Ren,  Jingyi and Zhou,  Haowen and Zeng,  Hu and Wang,  Connie Kangni and Huang,  Jiahao and Qiu,  Xiaojie and Sui,  Xin and Li,  Qiang and Wu,  Xunwei and Lin,  Zuwan and Lo,  Jennifer A. and Maher,  Kamal and He,  Yichun and Tang,  Xin and Lam,  Judson and Chen,  Hongyu and Li,  Brian and Fisher,  David E. and Liu,  Jia and Wang,  Xiao},
    year = {2023},
    month = apr,
    pages = {695–705}
}

@article{moncada2020integrating,
    title = {{Integrating microarray-based spatial transcriptomics and single-cell RNA-seq reveals tissue architecture in pancreatic ductal adenocarcinomas}},
    volume = {38},
    number = {3},
    journal = {Nature Biotechnology},
    publisher = {Springer Science and Business Media LLC},
    author = {Moncada,  Reuben and Barkley,  Dalia and Wagner,  Florian and Chiodin,  Marta and Devlin,  Joseph C. and Baron,  Maayan and Hajdu,  Cristina H. and Simeone,  Diane M. and Yanai,  Itai},
    year = {2020},
    month = jan,
    pages = {333–342}
}

@article{chen2022spatiotemporal,
    title = {Spatiotemporal transcriptomic atlas of mouse organogenesis using {DNA} nanoball-patterned arrays},
    volume = {185},
    number = {10},
    journal = {Cell},
    publisher = {Elsevier BV},
    author = {Chen,  Ao and Liao,  Sha and Cheng,  Mengnan and Ma,  Kailong and Wu,  Liang and Lai,  Yiwei and Qiu,  Xiaojie and Yang,  Jin and Xu,  Jiangshan and Hao,  Shijie and Wang,  Xin and Lu,  Huifang and Chen,  Xi and Liu,  Xing and Huang,  Xin and Li,  Zhao and Hong,  Yan and Jiang,  Yujia and Peng,  Jian and Liu,  Shuai and Shen,  Mengzhe and Liu,  Chuanyu and Li,  Quanshui and Yuan,  Yue and Wei,  Xiaoyu and Zheng,  Huiwen and Feng,  Weimin and Wang,  Zhifeng and Liu,  Yang and Wang,  Zhaohui and Yang,  Yunzhi and Xiang,  Haitao and Han,  Lei and Qin,  Baoming and Guo,  Pengcheng and Lai,  Guangyao and Muñoz-Cánoves,  Pura and Maxwell,  Patrick H. and Thiery,  Jean Paul and Wu,  Qing-Feng and Zhao,  Fuxiang and Chen,  Bichao and Li,  Mei and Dai,  Xi and Wang,  Shuai and Kuang,  Haoyan and Hui,  Junhou and Wang,  Liqun and Fei,  Ji-Feng and Wang,  Ou and Wei,  Xiaofeng and Lu,  Haorong and Wang,  Bo and Liu,  Shiping and Gu,  Ying and Ni,  Ming and Zhang,  Wenwei and Mu,  Feng and Yin,  Ye and Yang,  Huanming and Lisby,  Michael and Cornall,  Richard J. and Mulder,  Jan and Uhlén,  Mathias and Esteban,  Miguel A. and Li,  Yuxiang and Liu,  Longqi and Xu,  Xun and Wang,  Jian},
    year = {2022},
    month = may,
    pages = {1777--1792.e21}
}

@article{eng2019transcriptome,
    title = {{Transcriptome-scale super-resolved imaging in tissues by RNA seqFISH+}},
    volume = {568},
    number = {7751},
    journal = {Nature},
    publisher = {Springer Science and Business Media LLC},
    author = {Eng,  Chee-Huat Linus and Lawson,  Michael and Zhu,  Qian and Dries,  Ruben and Koulena,  Noushin and Takei,  Yodai and Yun,  Jina and Cronin,  Christopher and Karp,  Christoph and Yuan,  Guo-Cheng and Cai,  Long},
    year = {2019},
    month = mar,
    pages = {235–239}
}

@article{rodriques2019slide,
    title = {{Slide-seq: A scalable technology for measuring genome-wide expression at high spatial resolution}},
    volume = {363},
    number = {6434},
    journal = {Science},
    publisher = {American Association for the Advancement of Science (AAAS)},
    author = {Rodriques,  Samuel G. and Stickels,  Robert R. and Goeva,  Aleksandrina and Martin,  Carly A. and Murray,  Evan and Vanderburg,  Charles R. and Welch,  Joshua and Chen,  Linlin M. and Chen,  Fei and Macosko,  Evan Z.},
    year = {2019},
    month = mar,
    pages = {1463–1467}
}

@article{staahl2016visualization,
    title = {Visualization and analysis of gene expression in tissue sections by spatial transcriptomics},
    volume = {353},
    number = {6294},
    journal = {Science},
    publisher = {American Association for the Advancement of Science (AAAS)},
    author = {Ståhl,  Patrik L. and Salmén,  Fredrik and Vickovic,  Sanja and Lundmark,  Anna and Navarro,  José Fernández and Magnusson,  Jens and Giacomello,  Stefania and Asp,  Michaela and Westholm,  Jakub O. and Huss,  Mikael and Mollbrink,  Annelie and Linnarsson,  Sten and Codeluppi,  Simone and Borg,  Åke and Pontén,  Fredrik and Costea,  Paul Igor and Sahlén,  Pelin and Mulder,  Jan and Bergmann,  Olaf and Lundeberg,  Joakim and Frisén,  Jonas},
    year = {2016},
    month = jul,
    pages = {78–82}
}

@article{moses2022museum,
    title = {Museum of spatial transcriptomics},
    volume = {19},
    number = {5},
    journal = {Nature Methods},
    publisher = {Springer Science and Business Media LLC},
    author = {Moses,  Lambda and Pachter,  Lior},
    year = {2022},
    month = mar,
    pages = {534–546}
}

@inproceedings{lipman2023flow,
    title={Flow Matching for Generative Modeling},
    author={Yaron Lipman and Ricky T. Q. Chen and Heli Ben-Hamu and Maximilian Nickel and Matthew Le},
    booktitle={The Eleventh International Conference on Learning Representations },
    year={2023}
}

@article{tong2024improving,
    title={Improving and generalizing flow-based generative models with minibatch optimal transport},
    author={Alexander Tong and Kilian FATRAS and Nikolay Malkin and Guillaume Huguet and Yanlei Zhang and Jarrid Rector-Brooks and Guy Wolf and Yoshua Bengio},
    journal={Transactions on Machine Learning Research},
    year={2024},
}

@InProceedings{pmlr-v202-pooladian23a,
    title = 	 {Multisample Flow Matching: Straightening Flows with Minibatch Couplings},
    author =       {Pooladian, Aram-Alexandre and Ben-Hamu, Heli and Domingo-Enrich, Carles and Amos, Brandon and Lipman, Yaron and Chen, Ricky T. Q.},
    booktitle = 	 {Proceedings of the 40th International Conference on Machine Learning},
    pages = 	 {28100--28127},
    year = 	 {2023},
    volume = 	 {202},
    series = 	 {Proceedings of Machine Learning Research},
    month = 	 {23--29 Jul},
    publisher =    {PMLR},
}

@inproceedings{eijkelboom2024variational,
    title={Variational Flow Matching for Graph Generation},
    author={Floor Eijkelboom and Grigory Bartosh and Christian A. Naesseth and Max Welling and Jan-Willem van de Meent},
    booktitle={The Thirty-eighth Annual Conference on Neural Information Processing Systems},
    year={2024}
}

@article{klein2024genot,
    title={GENOT: Entropic (Gromov) Wasserstein flow matching with applications to single-cell genomics},
    author={Klein, Dominik and Uscidda, Th{\'e}o and Theis, Fabian and Cuturi, Marco},
    journal={Advances in Neural Information Processing Systems},
    volume={37},
    pages={103897--103944},
    year={2024}
}

@article{axolotl,
    author = {Xiaoyu Wei  and Sulei Fu  and Hanbo Li  and Yang Liu  and Shuai Wang  and Weimin Feng  and Yunzhi Yang  and Xiawei Liu  and Yan-Yun Zeng  and Mengnan Cheng  and Yiwei Lai  and Xiaojie Qiu  and Liang Wu  and Nannan Zhang  and Yujia Jiang  and Jiangshan Xu  and Xiaoshan Su  and Cheng Peng  and Lei Han  and Wilson Pak-Kin Lou  and Chuanyu Liu  and Yue Yuan  and Kailong Ma  and Tao Yang  and Xiangyu Pan  and Shang Gao  and Ao Chen  and Miguel A. Esteban  and Huanming Yang  and Jian Wang  and Guangyi Fan  and Longqi Liu  and Liang Chen  and Xun Xu  and Ji-Feng Fei  and Ying Gu },
    title = {Single-cell Stereo-seq reveals induced progenitor cells involved in axolotl brain regeneration},
    journal = {Science},
    volume = {377},
    number = {6610},
    pages = {eabp9444},
    year = {2022},
}

@article{yu2025tissue,
    title={Tissue reassembly with generative AI},
    author={Yu, Tingyang and Ekbote, Chanakya and Morozov, Nikita and Fan, Jiashuo and Frossard, Pascal and d’Ascoli, St{\'e}phane and Brbi{\'c}, Maria},
    journal={bioRxiv},
    pages={2025--02},
    year={2025},
    publisher={Cold Spring Harbor Laboratory}
}

@article{haviv2024wasserstein,
    title={Wasserstein Flow Matching: Generative modeling over families of distributions},
    author={Haviv, Doron and Pooladian, Aram-Alexandre and Pe'er, Dana and Amos, Brandon},
    journal={arXiv preprint arXiv:2411.00698},
    year={2024}
}

@article{klein2025mapping,
    title={Mapping cells through time and space with moscot},
    author={Klein, Dominik and Palla, Giovanni and Lange, Marius and Klein, Michal and Piran, Zoe and Gander, Manuel and Meng-Papaxanthos, Laetitia and Sterr, Michael and Saber, Lama and Jing, Changying and others},
    journal={Nature},
    pages={1--11},
    year={2025},
    publisher={Nature Publishing Group UK London}
}

@article{Sun2025aging,
    author={Sun, Eric D.
    and Zhou, Olivia Y.
    and Hauptschein, Max
    and Rappoport, Nimrod
    and Xu, Lucy
    and Navarro Negredo, Paloma
    and Liu, Ling
    and Rando, Thomas A.
    and Zou, James
    and Brunet, Anne},
    title={Spatial transcriptomic clocks reveal cell proximity effects in brain ageing},
    journal={Nature},
    year={2025},
    month={Feb},
    day={01},
    volume={638},
    number={8049},
    pages={160-171},
}

@inproceedings{liu2023flow,
    title={{Flow Straight and Fast: Learning to Generate and Transfer Data with Rectified Flow}},
    author={Xingchao Liu and Chengyue Gong and Qiang Liu},
    booktitle={The Eleventh International Conference on Learning Representations },
    year={2023}
}

@inproceedings{albergo2023building,
    title={Building Normalizing Flows with Stochastic Interpolants},
    author={Michael Samuel Albergo and Eric Vanden-Eijnden},
    booktitle={The Eleventh International Conference on Learning Representations },
    year={2023}
}

@InProceedings{pmlr-v238-tong24a,
    title = 	 {Simulation-Free {S}chrödinger Bridges via Score and Flow Matching},
    author =       {Tong, Alexander Y. and Malkin, Nikolay and Fatras, Kilian and Atanackovic, Lazar and Zhang, Yanlei and Huguet, Guillaume and Wolf, Guy and Bengio, Yoshua},
    booktitle = 	 {Proceedings of The 27th International Conference on Artificial Intelligence and Statistics},
    pages = 	 {1279--1287},
    year = 	 {2024},
    volume = 	 {238},
    series = 	 {Proceedings of Machine Learning Research},
    month = 	 {02--04 May},
    publisher =    {PMLR},
}

@inproceedings{atanackovic2025meta,
    title={Meta Flow Matching: Integrating Vector Fields on the Wasserstein Manifold},
    author={Lazar Atanackovic and Xi Zhang and Brandon Amos and Mathieu Blanchette and Leo J Lee and Yoshua Bengio and Alexander Tong and Kirill Neklyudov},
    booktitle={The Thirteenth International Conference on Learning Representations},
    year={2025}
}

@inproceedings{zhu2025diffusion,
    title={Diffusion Generative Modeling for Spatially Resolved Gene Expression Inference from Histology Images},
    author={Sichen Zhu and Yuchen Zhu and Molei Tao and Peng Qiu},
    booktitle={The Thirteenth International Conference on Learning Representations},
    year={2025}
}

@article{wan2023integrating,
    title={Integrating spatial and single-cell transcriptomics data using deep generative models with SpatialScope},
    author={Wan, Xiaomeng and Xiao, Jiashun and Tam, Sindy Sing Ting and Cai, Mingxuan and Sugimura, Ryohichi and Wang, Yang and Wan, Xiang and Lin, Zhixiang and Wu, Angela Ruohao and Yang, Can},
    journal={Nature Communications},
    volume={14},
    number={1},
    pages={7848},
    year={2023},
    publisher={Nature Publishing Group UK London}
}

@article{li2024stdiff,
    title={stDiff: a diffusion model for imputing spatial transcriptomics through single-cell transcriptomics},
    author={Li, Kongming and Li, Jiahao and Tao, Yuhao and Wang, Fei},
    journal={Briefings in Bioinformatics},
    volume={25},
    number={3},
    pages={bbae171},
    year={2024},
    publisher={Oxford University Press}
}

@article{ho2020denoising,
    title={Denoising diffusion probabilistic models},
    author={Ho, Jonathan and Jain, Ajay and Abbeel, Pieter},
    journal={Advances in neural information processing systems},
    volume={33},
    pages={6840--6851},
    year={2020}
}

@article{haviv2025covariance,
    title={The covariance environment defines cellular niches for spatial inference},
    author={Haviv, Doron and Rem{\v{s}}{\'\i}k, J{\'a}n and Gatie, Mohamed and Snopkowski, Catherine and Takizawa, Meril and Pereira, Nathan and Bashkin, John and Jovanovich, Stevan and Nawy, Tal and Chaligne, Ronan and others},
    journal={Nature Biotechnology},
    volume={43},
    number={2},
    pages={269--280},
    year={2025},
    publisher={Nature Publishing Group US New York}
}

@article{pham2023robust,
    title={Robust mapping of spatiotemporal trajectories and cell--cell interactions in healthy and diseased tissues},
    author={Pham, Duy and Tan, Xiao and Balderson, Brad and Xu, Jun and Grice, Laura F and Yoon, Sohye and Willis, Emily F and Tran, Minh and Lam, Pui Yeng and Raghubar, Arti and others},
    journal={Nature communications},
    volume={14},
    number={1},
    pages={7739},
    year={2023},
    publisher={Nature Publishing Group UK London}
}

@article{abdelaal2024sirv,
    title={SIRV: Spatial inference of RNA velocity at the single-cell resolution},
    author={Abdelaal, Tamim and Grossouw, Laurens M and Pasterkamp, R Jeroen and Lelieveldt, Boudewijn PF and Reinders, Marcel JT and Mahfouz, Ahmed},
    journal={NAR genomics and bioinformatics},
    volume={6},
    number={3},
    pages={lqae100},
    year={2024},
    publisher={Oxford University Press}
}

@article{long2025spvelo,
    title={spVelo: RNA velocity inference for multi-batch spatial transcriptomics data},
    author={Long, Wenxin and Liu, Tianyu and Xue, Lingzhou and Zhao, Hongyu},
    journal={bioRxiv},
    pages={2025--03},
    year={2025},
    publisher={Cold Spring Harbor Laboratory}
}

@article{bryan2025accurate,
    title={Accurate trajectory inference in time-series spatial transcriptomics with structurally-constrained optimal transport},
    author={Bryan, John Peterson and Farhi, Samouil L and Cleary, Brian},
    journal={bioRxiv},
    pages={2025--03},
    year={2025},
    publisher={Cold Spring Harbor Laboratory}
}

@article{shen2025inferring,
    title={Inferring cell trajectories of spatial transcriptomics via optimal transport analysis},
    author={Shen, Xunan and Zuo, Lulu and Ye, Zhongfei and Yuan, Zhongyang and Huang, Ke and Li, Zeyu and Yu, Qichao and Zou, Xuanxuan and Wei, Xiaoyu and Xu, Ping and others},
    journal={Cell Systems},
    volume={16},
    number={2},
    year={2025},
    publisher={Elsevier}
}

@article{chen2015spatially,
    title={Spatially resolved, highly multiplexed RNA profiling in single cells},
    author={Chen, Kok Hao and Boettiger, Alistair N and Moffitt, Jeffrey R and Wang, Siyuan and Zhuang, Xiaowei},
    journal={Science},
    volume={348},
    number={6233},
    pages={aaa6090},
    year={2015},
    publisher={American Association for the Advancement of Science}
}

@inproceedings{vayer2019optimal,
  title     = {{Optimal Transport for Structured Data with Application on Graphs}},
  author    = {Vayer, Titouan and Courty, Nicolas and Tavenard, Romain and Flamary, R{\'e}mi},
  booktitle = {Proceedings of the 36th International Conference on Machine Learning},
  pages     = {6275--6284},
  year      = {2019},
  volume    = {97},
  series    = {Proceedings of Machine Learning Research},
  publisher = {PMLR}
}

@inproceedings{chamferdistance,
    author = {Haoqiang Fan and Hao Su and Leonidas Guibas},
    title = {A Point Set Generation Network for 3D Object Reconstruction from a Single Image},
    booktitle = {Computer Vision and Pattern Recognition},
    year = {2016}
}

@Article{gw,
    author={M{\'e}moli, Facundo},
    title={Gromov--Wasserstein Distances and the Metric Approach to Object Matching},
    journal={Foundations of Computational Mathematics},
    year={2011},
    month={Aug},
    day={01},
    volume={11},
    number={4},
    pages={417-487},
}

@article{vayer2020fused,
    title = {Fused Gromov-Wasserstein Distance for Structured Objects},
    author={Vayer, Titouan and Chapel, Laetitia and Flamary, R{\'e}mi and Tavenard, Romain and Courty, Nicolas},
    journal={Algorithms},
    volume={13},
    number={9},
    pages={212},
    year={2020},
    publisher={MDPI}
}

@inproceedings{
    wasserstein-ex-deepruot,
    title={Learning stochastic dynamics from snapshots through regularized unbalanced optimal transport},
    author={Zhenyi Zhang and Tiejun Li and Peijie Zhou},
    booktitle={The Thirteenth International Conference on Learning Representations},
    year={2025},
}

@InProceedings{wasserstein-ex-trajectorynet,
  title = 	 {{T}rajectory{N}et: A Dynamic Optimal Transport Network for Modeling Cellular Dynamics},
  author =       {Tong, Alexander and Huang, Jessie and Wolf, Guy and Van Dijk, David and Krishnaswamy, Smita},
  booktitle = 	 {Proceedings of the 37th International Conference on Machine Learning},
  pages = 	 {9526--9536},
  year = 	 {2020},
  editor = 	 {III, Hal Daumé and Singh, Aarti},
  volume = 	 {119},
  series = 	 {Proceedings of Machine Learning Research},
  publisher =    {PMLR},
}

@inproceedings{
    wasserstein-ex-mioflow,
    title={Manifold Interpolating Optimal-Transport Flows for Trajectory Inference},
    author={Guillaume Huguet and Daniel Sumner Magruder and Alexander Tong and Oluwadamilola Fasina and Manik Kuchroo and Guy Wolf and Smita Krishnaswamy},
    booktitle={Advances in Neural Information Processing Systems},
    editor={Alice H. Oh and Alekh Agarwal and Danielle Belgrave and Kyunghyun Cho},
    year={2022},
}

@article{papanicolaou2022temporal,
  title={Temporal profiling of the breast tumour microenvironment reveals collagen XII as a driver of metastasis},
  author={Papanicolaou, Michael and Parker, Amelia L and Yam, Michelle and Filipe, Elysse C and Wu, Sunny Z and Chitty, Jessica L and Wyllie, Kaitlin and Tran, Emmi and Mok, Ellie and Nadalini, Audrey and others},
  journal={Nature communications},
  volume={13},
  number={1},
  pages={4587},
  year={2022},
  publisher={Nature Publishing Group UK London}
}

@article{halmos2025dest,
  title={DeST-OT: Alignment of spatiotemporal transcriptomics data},
  author={Halmos, Peter and Liu, Xinhao and Gold, Julian and Chen, Feng and Ding, Li and Raphael, Benjamin J},
  journal={Cell Systems},
  volume={16},
  number={2},
  year={2025},
  publisher={Elsevier}
}

@article{akbarnejad2025mapping,
  title={Mapping and reprogramming human tissue microenvironments with MintFlow},
  author={Akbarnejad, Amir and Steele, Lloyd and Jafree, Daniyal J and Birk, Sebastian and Sallese, Marta Rosa and Rademaker, Koen and Boxall, Adam and Rumney, Benjamin and Tudor, Catherine and Patel, Minal and others},
  journal={bioRxiv},
  pages={2025--06},
  year={2025},
  publisher={Cold Spring Harbor Laboratory}
}

@article {LUNA,
	author = {Yu, Tingyang and Ekbote, Chanakya and Morozov, Nikita and Fan, Jiashuo and Frossard, Pascal and d{\textquoteright}Ascoli, St{\'e}phane and Brbi{\'c}, Maria},
	title = {Tissue reassembly with generative AI},
	elocation-id = {2025.02.13.638045},
	year = {2025},
	doi = {10.1101/2025.02.13.638045},
	publisher = {Cold Spring Harbor Laboratory},
	eprint = {https://www.biorxiv.org/content/early/2025/02/17/2025.02.13.638045.full.pdf},
	journal = {bioRxiv}
}

\newpage
\appendix
\section{Broader impacts}\label{app: impact_statement}
This work addresses fundamental challenges in spatial transcriptomics by modeling complex spatial and compositional changes in developing tissues. We demonstrate how efficient representations of high-dimensional spatial cellular data can advance the understanding of developmental trajectories and microenvironment dynamics. We anticipate releasing \modelname{} as an open-source, user-friendly tool to enable broad application in spatial biology studies. Given its use with biological data, \modelname{} may also be applied in sensitive contexts involving clinical or patient information.

\section{Limitations}\label{app: limitations} Our approach relies on fixed OT feature weighting by a parameter $\lambda$ during training (see \cref{sec: pc_ot}), limiting flexibility at inference and potentially constraining certain biological analyses. Moreover, radius-based niche definitions may also be sub-optimal for small or irregularly shaped microenvironments, where the radius captures excessive spatial context and does not allow fine-grained modeling of the functional region's evolution. 

The model assumes that spatial slides can be aligned with respect to each other in time and requires normalization-based pre-processing. Future work will be directed towards rotational and translational invariant spatial constraints and the incorporation of cell-to-cell communication priors in the neighborhood definition. While the learned flow models cell population dynamics, it does not explicitly capture biological events such as division or death. 

Finally, this study focuses on in-distribution testing and does not consider generalization to unseen slides or full anatomical regions excluded from the training process. To achieve prediction in unseen settings, we foresee the need for technical replicates of the same slide across time points, as the model cannot extrapolate spatial arrangements without prior exposure to the associated region. We leave these analyses to future work when spatio-temporal measurements across multiple replicates become increasingly available.

\section{Mixed-factorized Variational Flow Matching}
\label{appendix:mixture_vfm}

\subsection{Theoretical aspects of Variational Flow Matching}\label{app: theoretical_vfm}
Variational Flow Matching (VFM) \citep{eijkelboom2024variational} relies on the observation that one can write the time-resolved marginal vector field $u_t(\vx)$ in FM as the expected conditional field $u_t(\vx \mid \vx_1)$ under the posterior $p_t(\vx_1 \mid \vx)$ as:
\begin{equation}\label{eq: marginal_field_original_paper}
u_t(\vx) = \E_{p_t(\vx_1 \mid \vx)}\left[ u_t(\vx \mid \vx_1) \right] \: .
\end{equation}
Since $u_t(\vx \mid \vx_1)$ has a closed form and $u_t(\vx)$ is all that we need to generate the probability path $p_t$ from noise to data, this opens the door to a new interpretation of the objective as a variational inference problem, where we approximate $p_t(\vx_1 \mid \vx)$ with a variational posterior $q^{\theta}_t(\vx_1 \mid \vx)$. In other words, one can optimize the following objective:
$$
\mathcal{L}_{\text{VFM}}(\theta) = -\mathbb{E}_{t \sim \gU[0,1], \vx_1\sim p_1(\vx_1), \vx \sim p_t(\vx \mid \vx_1)} \left[ \log q_t^\theta(\vx_1 \mid \vx) \right] \: ,
$$
where $p_1(\vx)$ is the data distribution and $p_t(\vx \mid \vx_1)$ a straight probability path. When $u_t(\vx \mid \vx_1)$ is linear in $\vx_1$, this model formulation acquires convenient properties listed below. 

\textbf{Mean parameterization.} The expected conditional field under the posterior only depends on the posterior mean: 
$$
\mathbb{E}_{p_t(\vx_1 \mid \vx)}\left[u_t(\vx \mid \vx_1)\right] = u_t(\vx_1 \mid \E_{p_t(\vx_1 \mid \vx)}\left[ \vx_1 \right]) \: , 
$$

suggesting that it is sufficient to parameterize the posterior mean to simulate data under the marginal flow. The posterior mean can be regressed against real samples $\vx_1$ during training. 

\textbf{Equivalence between posterior and approximate posterior formulation.} From the previous point, it follows that the expectation of the conditional field is the same under the true and approximate posterior, as long as their first moments match. 

\textbf{Efficient simulation.} Given a parameterized posterior mean $\mu_t^{\theta}$, simulating the generative field in \cref{eq: marginal_field_original_paper} is efficient under the linearity condition. For example, in the standard FM setting with straight paths \citep{lipman2023flow}, the marginal generative field becomes:
\begin{align}
u_t(\vx) 
&= \mathbb{E}_{q^{\theta}_t(\vx_1 \mid \vx)}\left[u_t(\vx \mid \vx_1)\right] \\
&= u_t\left(\vx \mid \mathbb{E}_{q^{\theta}_t(\vx_1 \mid \vx)}\left[ \vx_1 \right]\right) \\
&= \frac{\mu_t^{\theta}(\vx) - \vx}{1 - t} \: .
\end{align}
which can be easily simulated in the range $t \in [0,1]$.

\subsection{Factorized posterior}\label{app: theoretical_factorized_posterior}
Similar to \citet{eijkelboom2024variational}, in our work, we use a fully factorized posterior, where individual dimensions can follow different families of distributions with finite moments (see \cref{sec: mixture_vfm}). Notably, a factorized approximate posterior over $\vx_1$ is allowed as a choice for $q^{\theta}_t$, since the only requirement to simulate $u_t(\vx)$ is for $q_t^{\theta}(\vx_1 \mid \vx)$ to match the expectation of $p_t(\vx_1 \mid \vx)$ over $\vx_1$, irrespectively of higher moments or correlations between factors. 

In this regard, it is useful to consider the following proposition.

\begin{proposition}
Let $\vx_1 \in \mathbb{R}^D$ be a $D$-dimensional target data point, $p_t(\vx_1 \mid \vx)$ the posterior probability path conditioned on a noisy point $\vx \sim p_t(\vx)$, and $u_t(\vx \mid \vx_1)$ the conditional velocity field. Assume that $u_t(\vx \mid \vx_1)$ is linear in $\vx_1$. Then, for any dimension $d \in \{1, \ldots, D\}$, the following holds:
\begin{align}
    &\mathbb{E}_{p_t(\vx_1 \mid \vx)}[x_1^d] = \mathbb{E}_{p_t(x_1^d \mid \vx)}[x_1^d] \label{eq:dim-marginal} \\
    &u_t(x^d) = u_t\left(x^d \mid \mathbb{E}_{p_t(x_1^d \mid \vx)}[x_1^d]\right) \label{eq:velocity-field} \: ,
\end{align}
where $x^d$ refers to the $d^{\mathrm{th}}$ scalar dimension of the vector $\vx$.
\end{proposition}

\textit{Proof.}
We begin by proving \cref{eq:dim-marginal} using marginalization:
\begin{align}
    \mathbb{E}_{p_t(\vx_1 \mid \vx)}[x_1^d] 
    &= \int x_1^d \, p_t(\vx_1 \mid \vx) \, \mathrm{d}\vx_1 \nonumber \\
    &= \int x_1^d \left( \int p_t(\vx_1 \mid \vx) \, \mathrm{d}\vx_1^{\setminus d} \right) \mathrm{d}x_1^d \nonumber \\
    &= \int x_1^d \, p_t(x_1^d \mid \vx) \, \mathrm{d}x_1^d \label{eq:equality_proof} \: .
\end{align}
Next, we prove \cref{eq:velocity-field}. Under the assumption that the conditional velocity field $u_t(\vx \mid \vx_1)$ is linear in $\vx_1$, we have:
\begin{align}
    u_t(x^d) 
    &= \mathbb{E}_{p_t(\vx_1 \mid \vx)}\left[ u_t(x^d \mid \vx_1) \right] \nonumber \\
    &\overset{(1)}{=} \mathbb{E}_{p_t(\vx_1 \mid \vx)}\left[ u_t(x^d \mid x_1^d) \right] \nonumber \\
    &= \mathbb{E}_{p_t(\vx_1 \mid \vx)}\left[ \frac{x_1^d - x^d}{1 - t} \right] \nonumber \\
    &= \frac{\mathbb{E}_{p_t(\vx_1 \mid \vx)}[x_1^d] - x^d}{1 - t} \nonumber \\
    &\overset{\cref{eq:equality_proof}}{=} \frac{\mathbb{E}_{p_t(x_1^d \mid \vx)}[x_1^d] - x^d}{1 - t} \nonumber \\
    &= u_t\left( x^d \mid \mathbb{E}_{p_t(x_1^d \mid \vx)}[x_1^d] \right) \: .
\end{align}
Here, step (1) follows from the linearity assumption, which ensures that the conditional velocity at $x^d$ depends only on $x_1^d$.

In other words, the expected value under the posterior at an individual feature $d$ does not depend on the other features and has an influence only on the $d$-th dimension of the conditional vector field. This flexibility allows each dimension’s approximate posterior to be chosen from a potentially different distributional family, as long as the first moment exists and is correctly parameterized.

\subsection{Marginal field derivation in source-conditioned VFM}\label{app: marg_field_deriv_cond_vfm}

When applying source conditioning to VFM, the marginal conditional vector field given a source $\vx_0$ is:
\begin{align}\label{eq: app-source_cond_posterior}
    u_t(\vx \mid \vx_0) 
    &= \int u_t(\vx \mid \vx_1)\, \frac{p_t(\vx \mid \vx_1)\, \pi(\vx_1 \mid \vx_0)}{p_t(\vx \mid \vx_0)} \, \rd\vx_1
\end{align}
where the $p_t(\vx \mid \vx_1)$ is a probability path interpolating observations $\vx_1$ with noise. Note that we omit $\vx_0$ from the probability path and conditional velocity as they are fully determined by $\vx_1$ under linear conditional probability paths. Furthermore, we can rewrite the marginal as an expectation:
\begin{equation}
    \int u_t(\vx \mid \vx_1)\, \frac{p_t(\vx \mid \vx_1)\, \pi(\vx_1 \mid \vx_0)}{p_t(\vx \mid \vx_0)} \, \rd\vx_1 = \E_{p_t(\vx_1 \mid \vx, \vx_0)} \left[ u_t(\vx \mid \vx_1) \right] \: ,
\end{equation}
where we used that $p_t(\vx \mid \vx_1)=p_t(\vx \mid \vx_0, \vx_1)$. 

\subsection{Gaussian and Laplacian Hybrid VFM}\label{app: objective_derivation}

We define a hybrid Variational Flow Matching (VFM) model using a fully factorized variational distribution over individual points in the target microenvironment \( \mathcal{M}^1 \). Following the mean-field assumption, the variational distribution factorizes over spatial and feature dimensions:
\begin{align}
    q_t^{\theta}(\mathcal{M}^1 \mid \mathcal{M}, \mathcal{M}^0)
    &= \prod_{(\vc_1, \vx_1) \in \mathcal{M}^1} q_t^{\theta}(\vc_1, \vx_1 \mid \mathcal{M}, \mathcal{M}^0) \\
    &= \prod_{(\vc_1, \vx_1) \in \mathcal{M}^1} 
    \left( \prod_{k=1}^2 f_t^{\theta}(c_1^k \mid \mathcal{M}, \mathcal{M}^0) \cdot 
           \prod_{d=1}^D r_t^{\theta}(x_1^d \mid \mathcal{M}, \mathcal{M}^0) \right) \: .
\end{align}
In line with \citet{eijkelboom2024variational}, using FM with straight paths enables us to efficiently simulate the marginal generating field using the \textit{first moment} of the posterior distribution. In other words, for a fully factorized posterior, we only need to parameterize a mean predictor. In our setting, the mean prediction is a neural network $\mu_t^{\theta}$ as a time-condition function of a noisy microenvironment \( \mathcal{M} \) and a source \( \mathcal{M}^0 \) with outputs: 
$$
(\bar{\vf}_t^{\theta}, \bar{\vr}_t^{\theta}) = \mu_t^{\theta}(\mathcal{M}, \mathcal{M}^0) \: ,  
$$
where $\bar{f}_t^{\theta, k}$ and  $\bar{r}_t^{\theta, d}$ are the expected values for the $k^{\text{th}}$ coordinate and $d^{\text{th}}$ cell feature. Then, we choose a parameterization for the variational factors at time $t \in [0,1]$ as follows:
\begin{align}
    x_1^d &\sim \mathcal{N}(\bar{r}_t^{\theta, d}, 1), \\
    c_1^k &\sim \text{Laplace}(\bar{f}_t^{\theta, k}, 1),
\end{align}

Substituting into the negative log-likelihood yields:

\begin{align}
    &- \log (q_t^{\theta}(\mathcal{M}^1 \mid \mathcal{M}, \mathcal{M}^0)) \\
    &= -\log \left( \prod_{(\vc_1, \vx_1) \in \mathcal{M}^1} 
    \left( \prod_{k=1}^2 f_t^{\theta}(c_1^k \mid \mathcal{M}, \mathcal{M}^0) \cdot 
           \prod_{d=1}^D r_t^{\theta}(x_1^d \mid \mathcal{M}, \mathcal{M}^0) \right) \right) \\
    &= \sum_{(\vc_1, \vx_1) \in \mathcal{M}^1} \Bigg( 
        \sum_{k=1}^2 \left( \log 2 + |c_1^k - \bar{f}_t^{\theta, k}| \right) \nonumber  + \sum_{d=1}^D \left( \frac{1}{2}\log (2\pi) + \frac{1}{2}\left(x_1^d - \bar{r}_t^{\theta, d}\right)^2 \right) \Bigg) \\
    &= \sum_{(\vc_1, \vx_1) \in \mathcal{M}^1}
        \left( \| \vc_1 - \bar{\vf}_t^{\theta} \|_1 
        + \frac{1}{2} \| \vx_1 - \bar{\vr}_t^{\theta} \|_2^2 \right)
        + \text{const w.r.t. } \theta
\end{align}
This results in a loss consisting of an \( \ell_1 \) error on spatial coordinates and a mean squared error on gene expression features, consistent with the hybrid variational design.

\newpage
\section{Additional results}

\subsection{Additional comparisons with moscot on embryonic development}
We propose a similar analysis as presented in \cref{sec: bio_eval}. 
\begin{figure}[H]
    \centering    \includegraphics[width=0.75\linewidth]{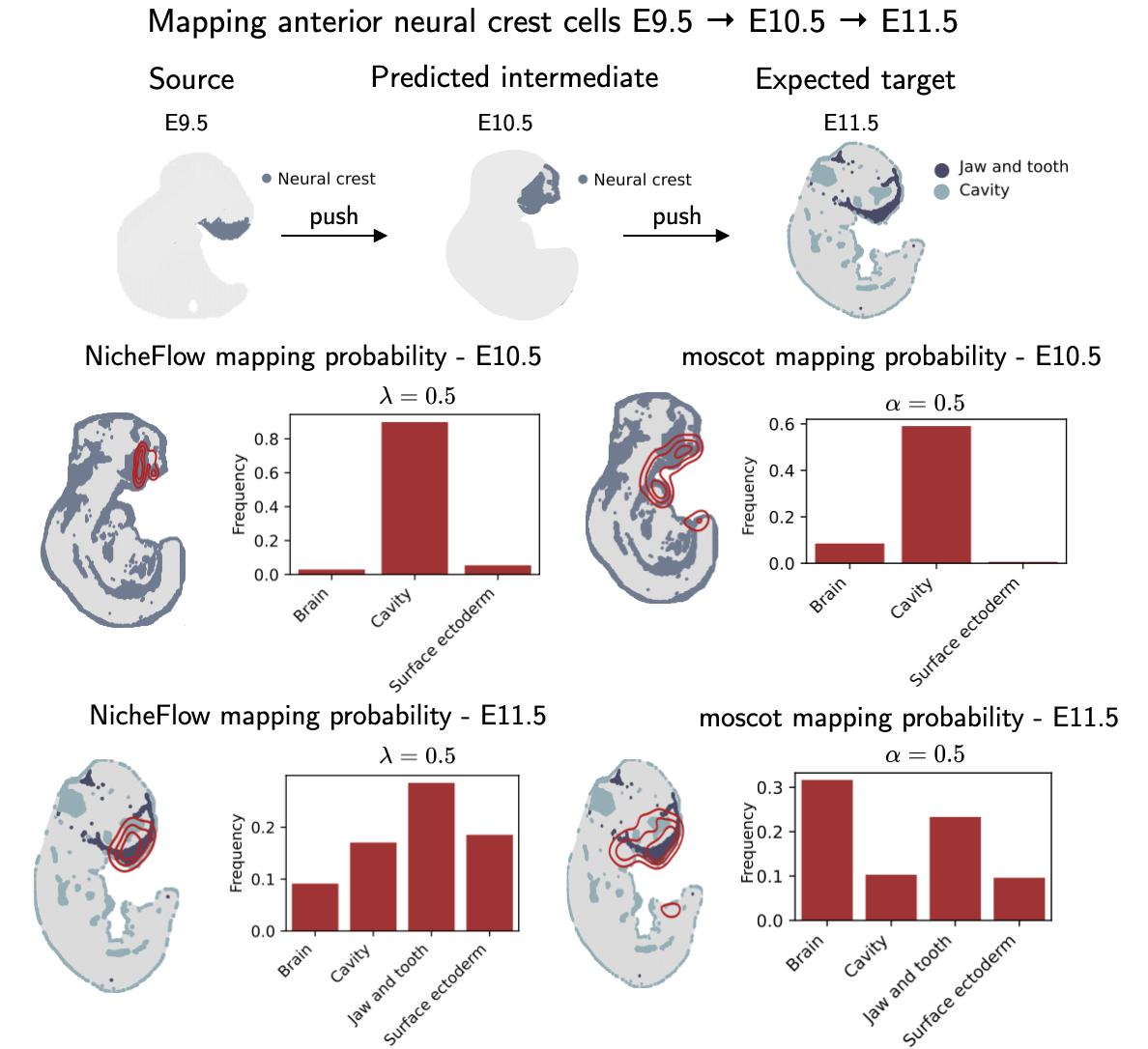}
    \caption{Comparison of NicheFlow and moscot on the prediction of the anterior neural crest cells' fate. For both models, we take source facial neural crest cells at E9.5, push them to time point E10.5, and show the compositional and density predictions in the middle panel. Then, the predictions at 10.5 are used as a source for a second trajectory prediction operation from 10.5 to 11.5, for which we inspect again the cell density over the target slide and the cell type probabilities. 
}
    \label{fig: jaw_teeth_double_push}
\end{figure}
In \cref{fig: jaw_teeth_double_push}, we compare the ability of NicheFlow to predict an entire spatial structure trajectory by pushing an initial source cloud through all the developmental stages. To this end, the trajectory of an initial point cloud is first pushed to the next time point, and the model's prediction is used as a source for predicting the subsequent time point. An accurate niche trajectory reconstruction signifies that our model can be used to sequentially predict microenvironment evolution by treating its intermediate predictions as inputs, corroborating their accurate reflection of real point clouds. 

In \cref{fig: jaw_teeth_double_push}, we show that pushing anterior neural crest cells twice from E9.5 to E11.5 through the flow generates realistic target point clouds with a cell composition reflecting the expected cranial structure, mostly made of cavity cells, jaw and teeth (arising at E11.5 for the first time) and surface ectoderm. Doing the same with moscot oversamples regions outside of the cranial structure, thereby incorrectly mapping most of the neural crest density to brain cells. 

\begin{figure}[H]
    \centering    \includegraphics[width=0.60\linewidth]{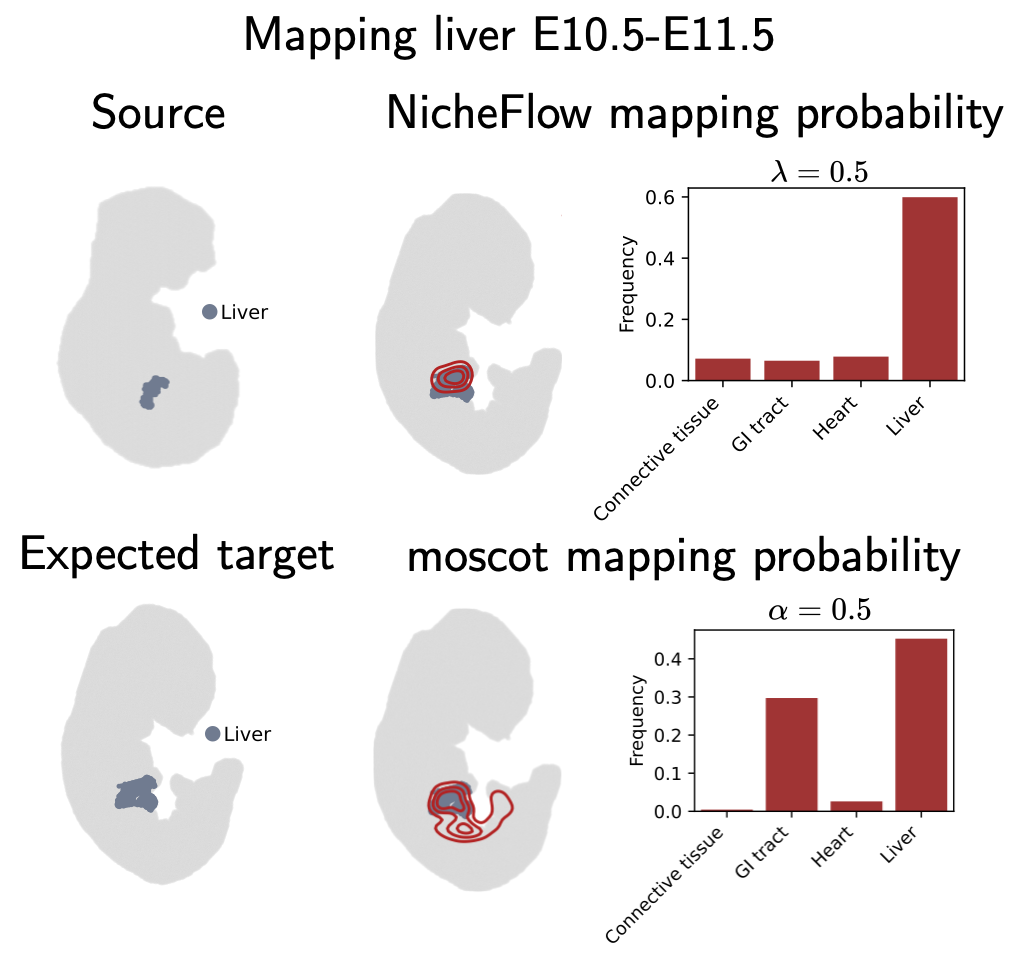}
    \caption{Comparison between moscot and NicheFlow on mapping the liver structure from E10.5 to E11.5. The liver at time E10.5 is used as a source for trajectory prediction using the different models. The left column shows the source and expected target regions highlighted on the respective E10.5 and E11.5 embryos. The middle column displays the density of the prediction obtained by transporting niches from the source to the target slide. On the right, the aggregated cell type proportions according to the density in the middle column (see \cref{app: biological experiments}).
}
    \label{fig: liver_double_push}
\end{figure}

In \cref{fig: liver_double_push}, we also show that NicheFlow is more accurate than moscot at transporting mass from defined organs like the liver across development. More specifically, while density leaks from the liver to the GI tract in the mapping produced by moscot, the prediction computed by NicheFlow more accurately retrieves the liver structure at the later time point. Together with previous evidence, our results underscore the importance of accounting for spatial correlations between cells during OT-based trajectory inference to buffer out the noise resulting from single-cell-based predictions. 

\subsection{$\alpha$ parameter sweep in moscot}\label{sec: moscot_sweep}
The comparison with moscot assesses how well each model captures compositional changes in anatomical structures or migratory patterns, depending on the use case. Qualitatively, we found that the spatial component in the OT problem considered by moscot and regulated by the hyperparameter $\alpha$ is comparable to our spatial term $\alpha$ across the 0.1–0.9 range in the \cref{fig: bio_analysis} analysis.

In \cref{tab:moscot_sweep}, we report the proportion of source density mapped to the correct cell type across $\alpha$ values, to support our choice for the results in \cref{sec: bio_eval}. For the spinal cord (\cref{fig: bio_analysis}, left), values above 0.75 yielded better qualitative and quantitative results. For neural crest cells (\cref{fig: bio_analysis}, right), $\alpha=0.5$ performs best.

More specifically, when mapping fixed structures over time, values below 0.75 caused excessive density dispersion outside the anatomical region. For migration, no value led to generally accurate transitions, though $\alpha=0.5$ mapped the highest density to the expected cell type.

\begin{table}[h]
\centering
\caption{Effect of the parameter $\alpha$ balancing spatial and cell state preservation in moscot. The results in the table indicate the percentage of source density mapped to the correct cell type from the source anatomical structure (the higher, the better).}
\label{tab:moscot_sweep}
\resizebox{0.75\textwidth}{!}{%
    \begin{tabular}{lccccc}
        \toprule
        Tissue & $\alpha = 0.1$ & $\alpha = 0.25$ & $\alpha = 0.5$ & $\alpha = 0.75$ & $\alpha = 0.9$ \\
        \midrule
        Spinal Cord & 0.146 & 0.190 & 0.712 & 0.729 & 0.725 \\
        Neural Crests & 0.159 & 0.160 & 0.162 & 0.155 & 0.112 \\
        \bottomrule
    \end{tabular}%
}
\end{table}

\subsection{Additional experiments on the axolotl brain development and aging datasets}
\label{appendix: qualitative-results}
\begin{figure}[H]
    \centering
    \includegraphics[width=\linewidth]{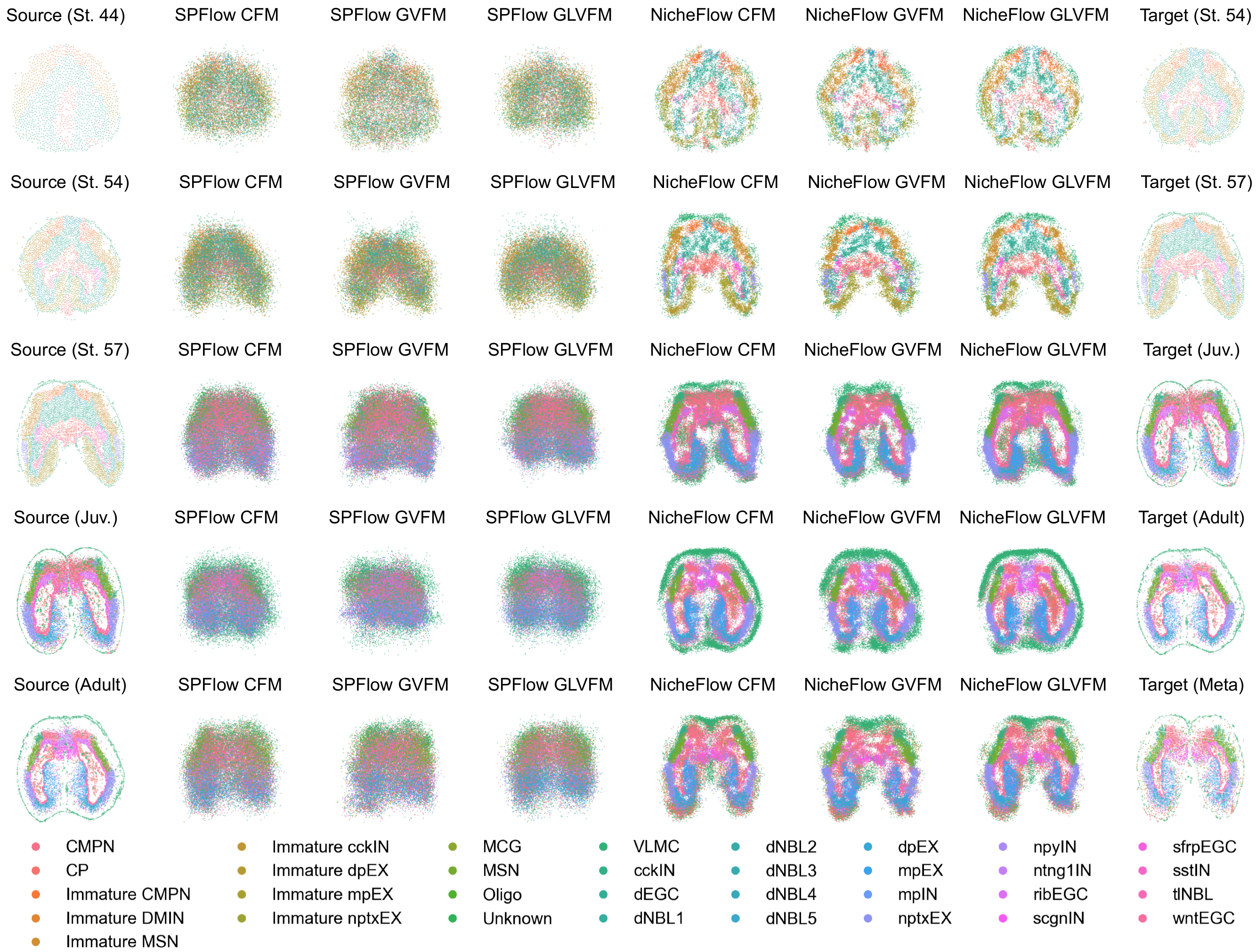}
    \caption{Qualitative comparison of generated samples on the axolotl brain development dataset (Stage 44, 54, 57, Juvenile, Adult, Meta). We show source and target samples alongside predictions from SPFlow and \modelname{} with different objectives. \modelname{} captures the spatial structure and cell-type organization more faithfully across developmental stages.}
    \label{fig: axolotl-figure}
\end{figure}

We provide additional visualizations of the generated samples on the axolotl brain development dataset, presented in \cref{fig: axolotl-figure}. As can be seen from the figure, NicheFlow correctly retrieves the spatial and anatomical characteristics of the brain, including hemisphere formation and cavity. 

\begin{figure}[H]
    \centering
    \includegraphics[width=0.8\linewidth]{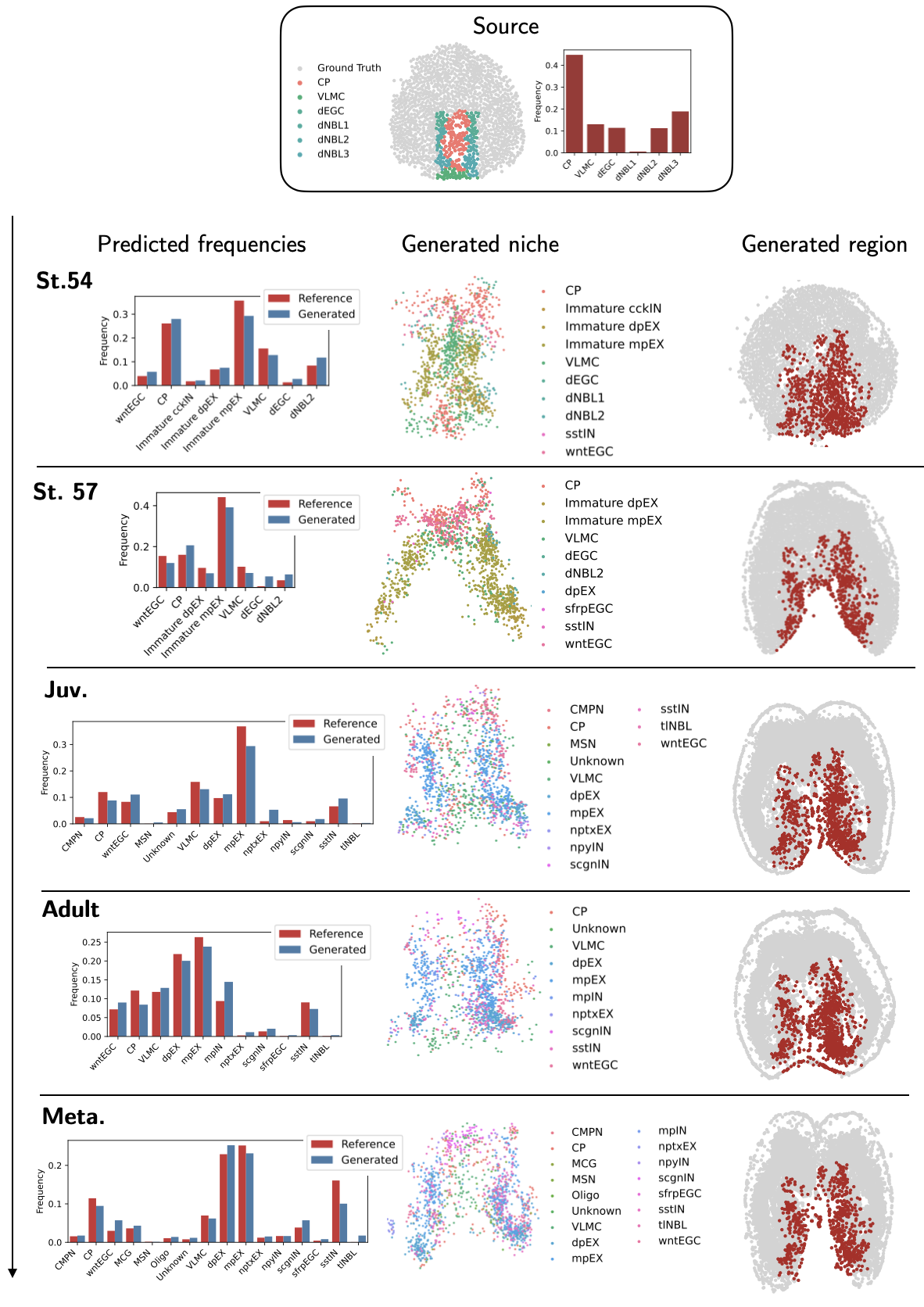}
    \caption{Prediction of hemisphere formation on the axolotl brain development. We evaluate the ability of \modelname{} to generate the anatomical splitting of central brain structure upon the formation of the right and left brain regions. The top panel shows the reference source region and cell type composition. For each stage (St.54, St.57, Juvenile, Adult, and Meta), we show: (1) predicted vs. reference cell type frequencies, (2) the generated niche visualized via 2D embedding colored by cell type, and (3) spatial projection of the generated region onto the anatomical reference (right). We set $\lambda=0.5$.}

    \label{fig: axolotl-anatomical-middle-sep}
\end{figure}

\begin{figure}[H]
    \centering
    \includegraphics[width=0.8\linewidth]{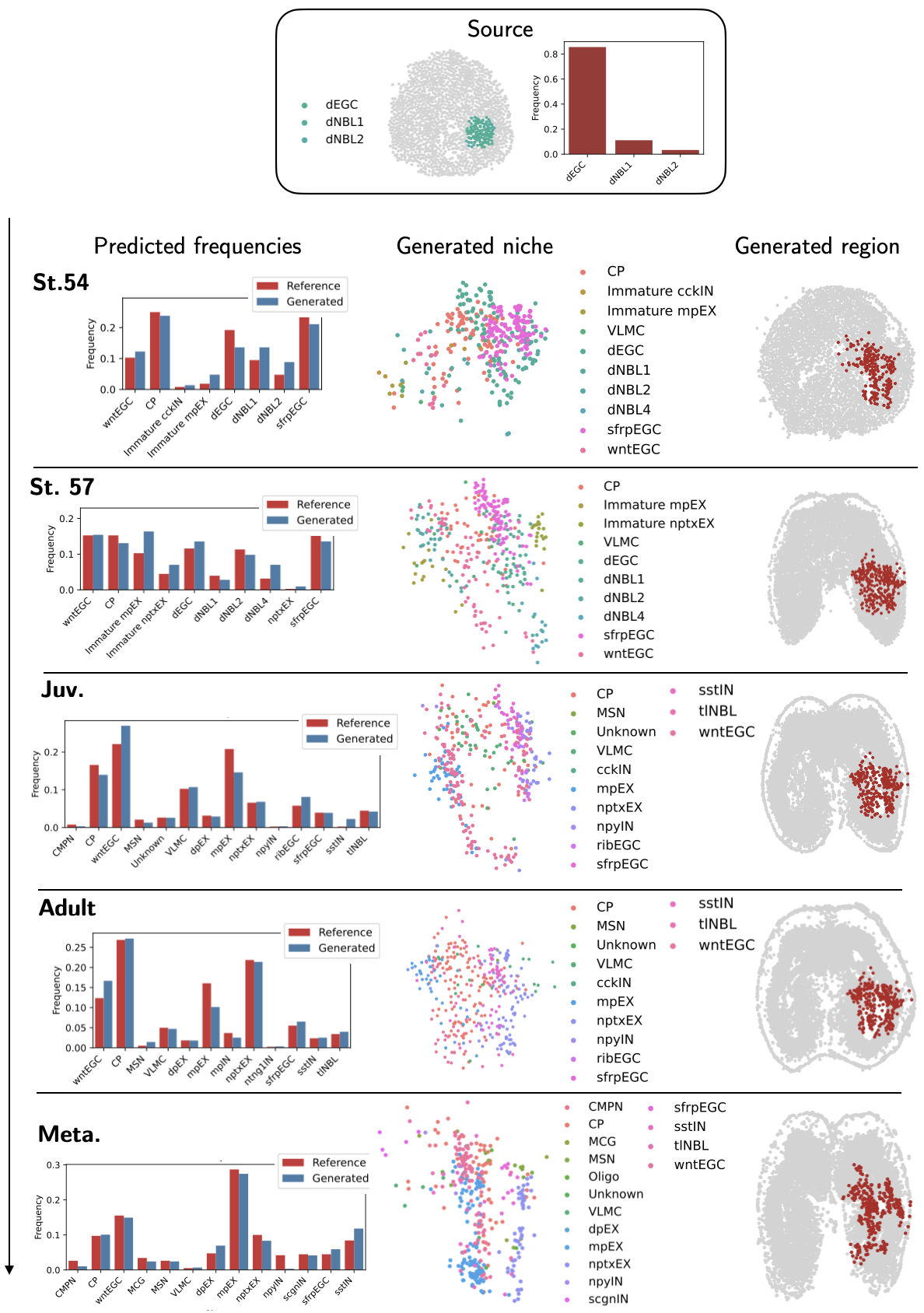}
    \caption{Prediction of cavity formation in the left brain lobe during axolotl development. We assess the ability of \modelname{} to model the emergence of a cavity structure within the left lobe of the axolotl brain. The top panel shows the reference source region and its cell type composition. For each developmental stage (St.54, St.57, Juvenile, Adult, and Meta), we display: (1) predicted versus reference cell type frequencies, (2) the generated niche visualized in 2D embedding space, colored by cell type, and (3) spatial projection of the generated region onto the anatomical brain reference (right). The progression illustrates the model's ability to recapitulate the asymmetric cavity formation localized to the left lobe. We set $\lambda=0.5$.}

    \label{fig: axolotl-anatomical-cavity}
\end{figure}

To further support our result, we propose a more in-depth qualitative analysis of the NicheFlow application on the axolotl brain development dataset. More specifically, in \cref{fig: axolotl-anatomical-middle-sep} we demonstrate that our model predicts the formation of crucial anatomical structures like the left and right lobes both spatially and compositionally. This vouches for flexibility in NichFlow's performance, which extends to non-trivial topology changes and simultaneously accounts for accurate cell state and coordinate generation in time. Similar results can be observed in \cref{fig: axolotl-anatomical-cavity}, where we showcase the correct prediction of the formation of a left lobe cavity, predicting trajectories from an immature brain region.

\begin{figure}[H]
    \centering
    \includegraphics[width=\linewidth]{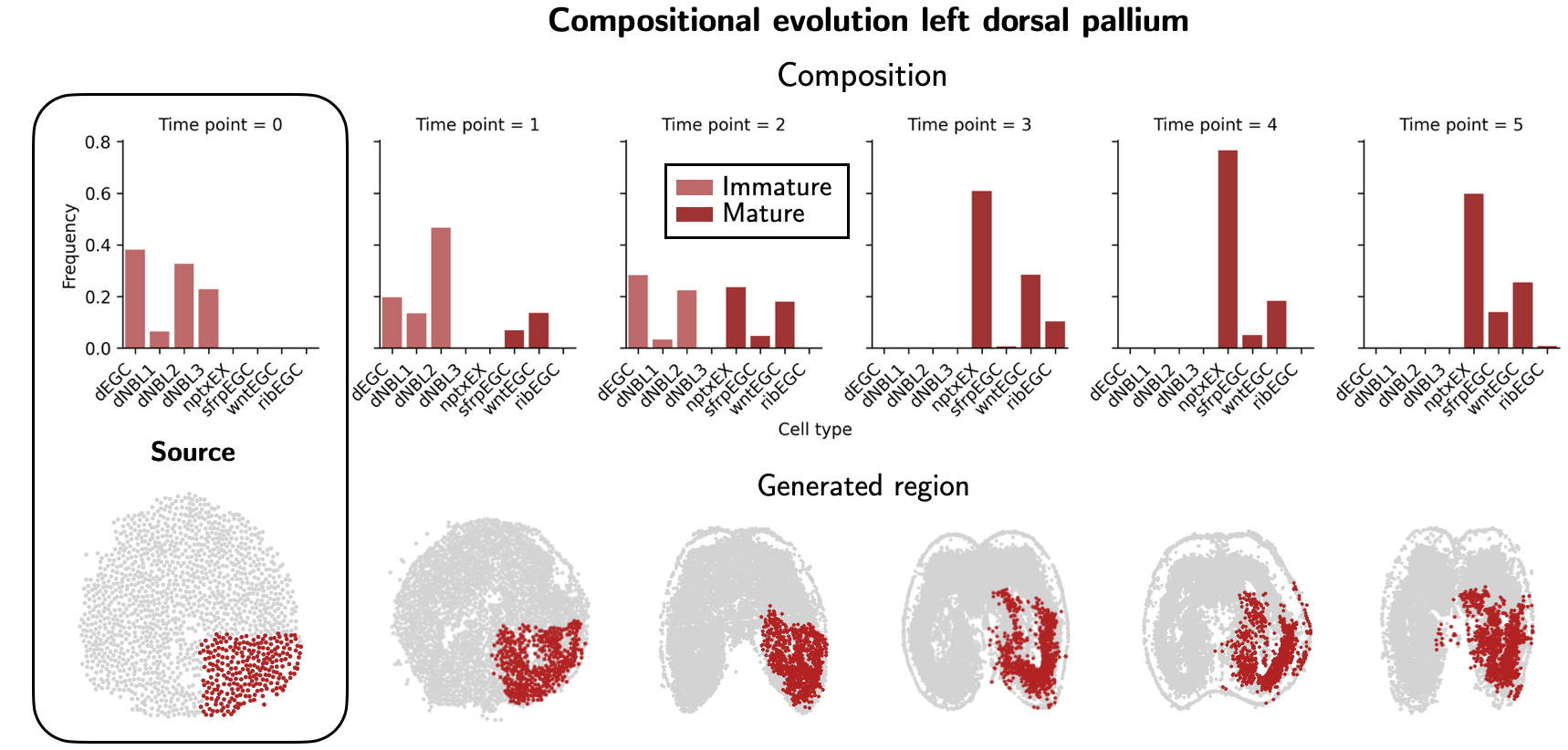}
    \caption{Structural and compositional prediction of the left dorsal pallium during the axolotl brain development. We use the left dorsal pallium region at the St.44 developmental stadium (highlighted on the left) and predict its trajectory over time. In the top row, we show the proportion of different cell types in the predicted region, colored as immature (light red) and mature (dark red). At the bottom, we show the structural prediction for the left dorsal pallium overlaid on the true slide. We set $\lambda=0.5$.}
    \label{fig: axolotl-anatomical-left-dorsal}
\end{figure}

Moreover, in \cref{fig: axolotl-anatomical-left-dorsal} we predict the compositional and structural time evolution of the left dorsal pallium in the axolotl brain development. Following Wei et al. \cite{axolotl}, we know that early time points populate the dorsal pallium of immature cell types like ependymoglial cells (EGC), neuroblasts (NBL), and immature neurons. In the left dorsal pallium, these disappear at the juvenile stages (the 3$^{\textrm{rd}})$ and lead to differentiation into mature neurons (nptxEX) and later EGC (WntEGC, sfrpEGC). This fixed structural development was accurately predicted by NicheFlow when pushing left dorsal pallium cells forward across the trajectory. 

\begin{figure}[H]
    \centering
    \includegraphics[width=\linewidth]{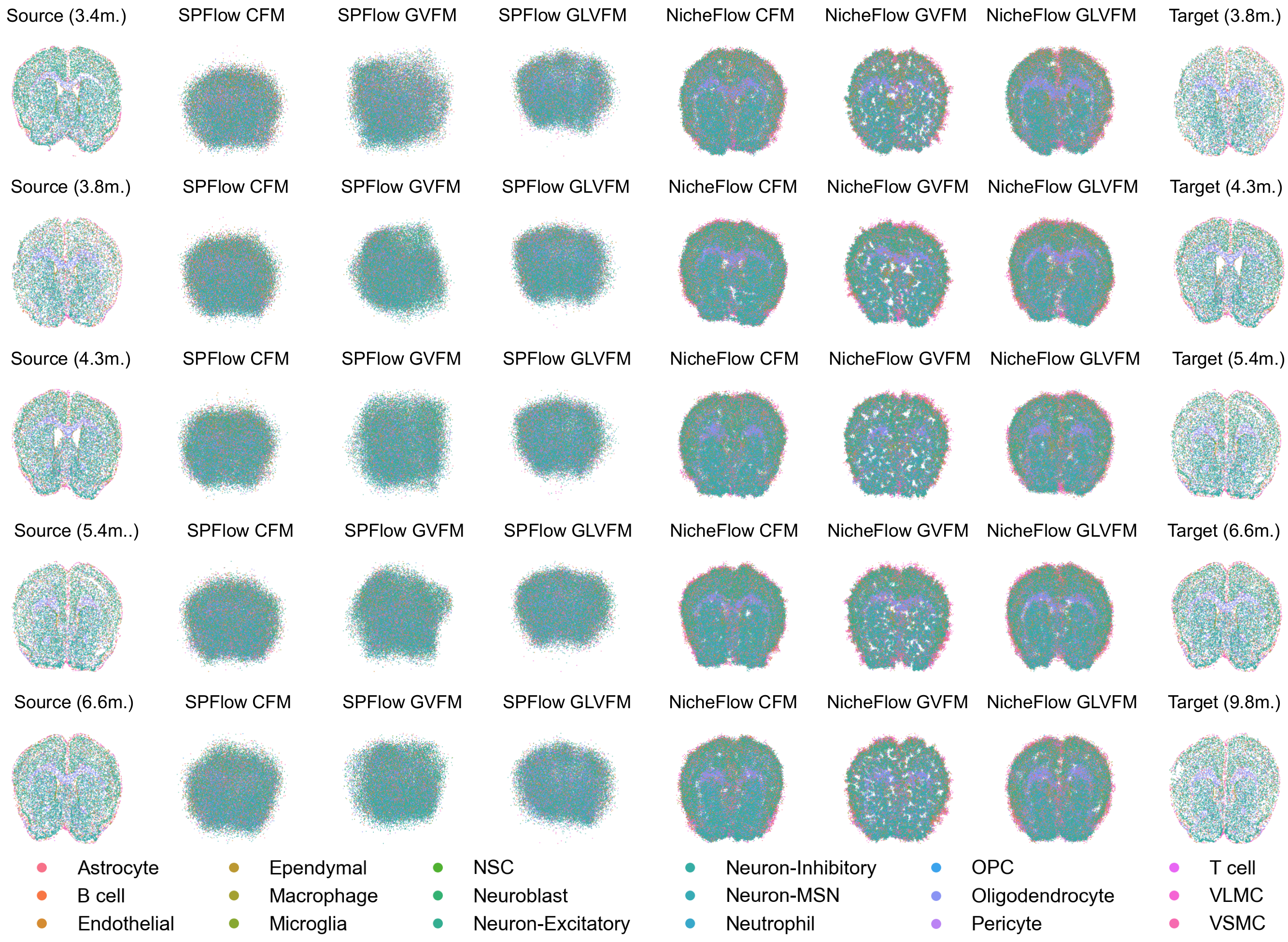}
    \caption{Qualitative comparison of generated samples on the mouse brain aging dataset (3.4, 3.8, 4.3, 5.4, 6.6, 9.8 months). We show source and target samples alongside predictions from SPFlow and \modelname{} with different objectives. \modelname{} captures the spatial structure and cell-type organization more faithfully across developmental stages.}
    \label{fig: aging-figure}
\end{figure}

Finally, similar to \cref{fig: moscot-figure} and \cref{fig: axolotl-figure}, in \cref{fig: aging-figure}, we qualitatively show that NicheFlow with microenvironment sampling strategy and mixed-factorized VFM is the best approach for reconstructing mouse brain trajectories in time. 

\newpage

\newpage
\newpage
\subsection{Conditional generation}
\label{appendix: conditioanl-generation}
\begin{figure}[!htb]
    \centering
    \includegraphics[width=0.9\linewidth]{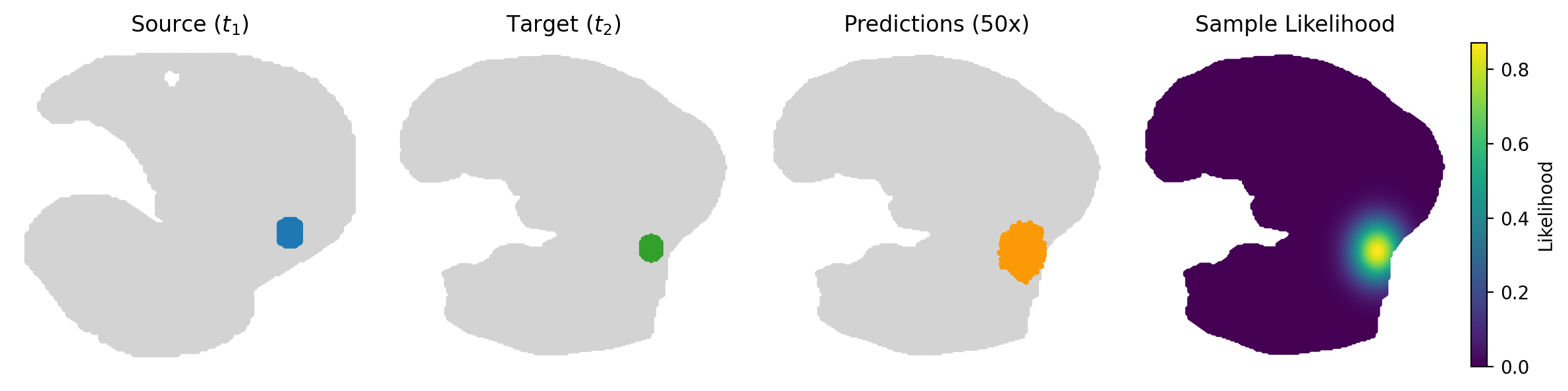}
    \includegraphics[width=0.9\linewidth]{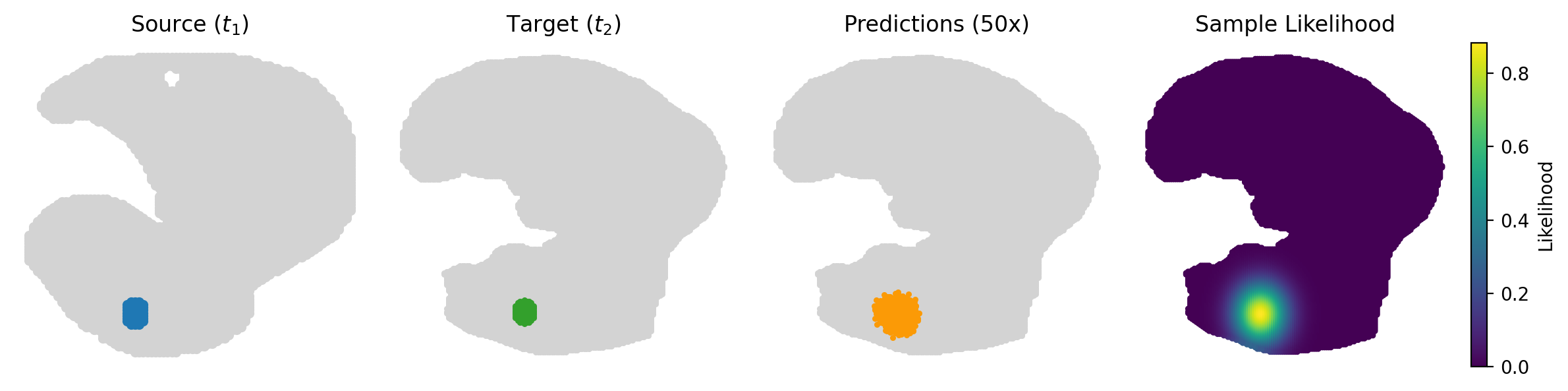}
    \includegraphics[width=0.9\linewidth]{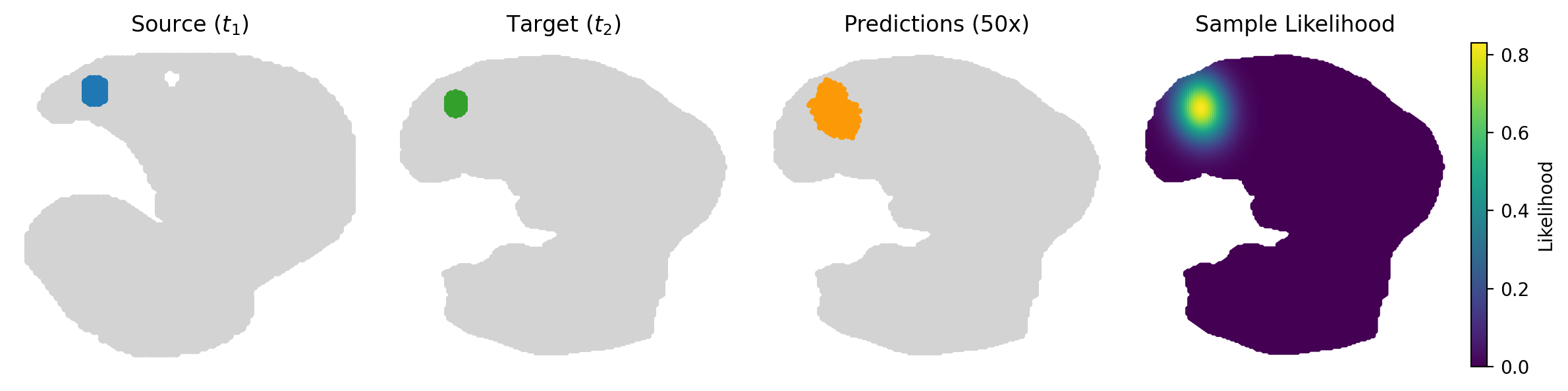}
    \includegraphics[width=0.9\linewidth]{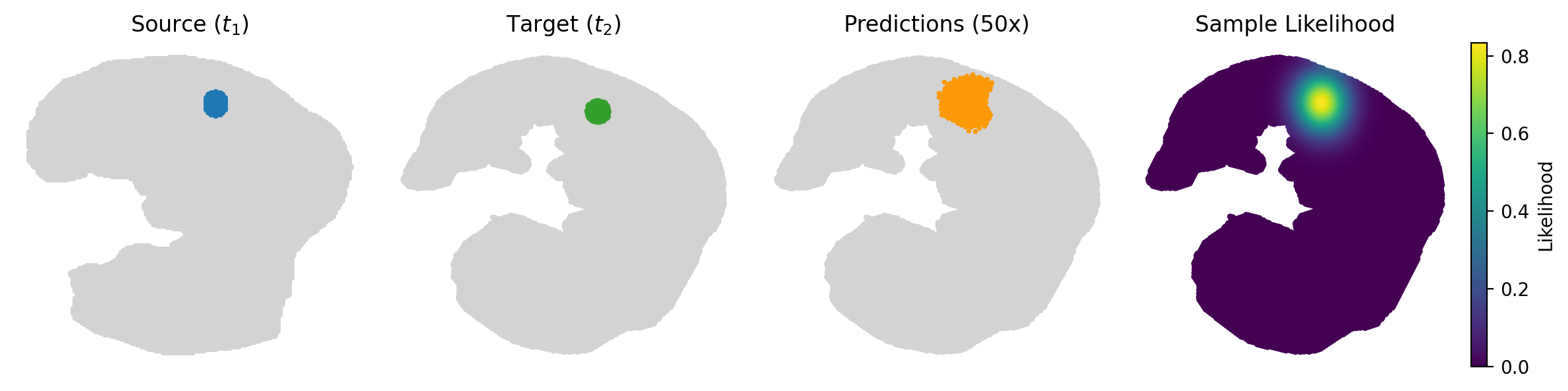}
    \includegraphics[width=0.9\linewidth]{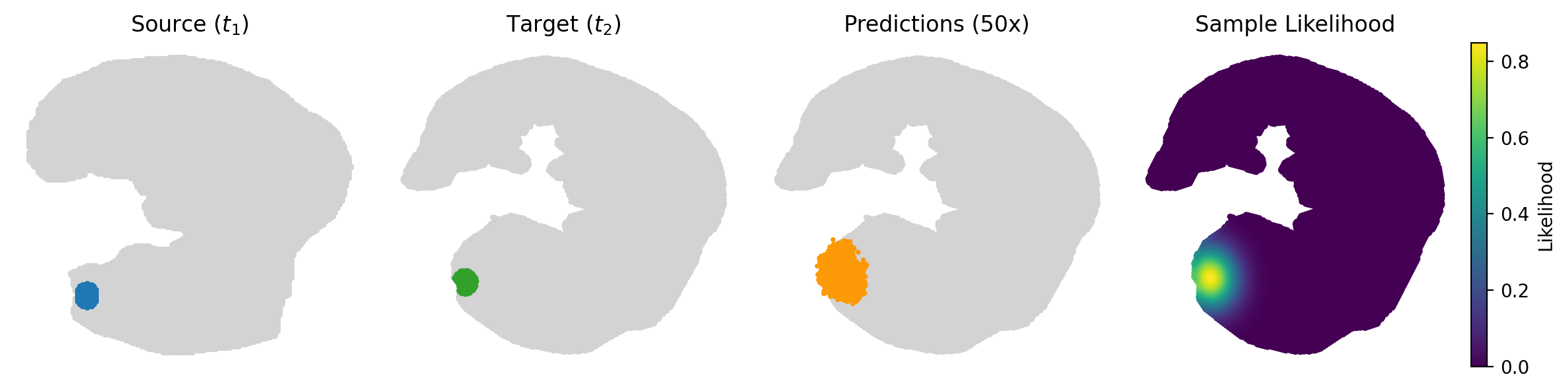}
    \includegraphics[width=0.9\linewidth]{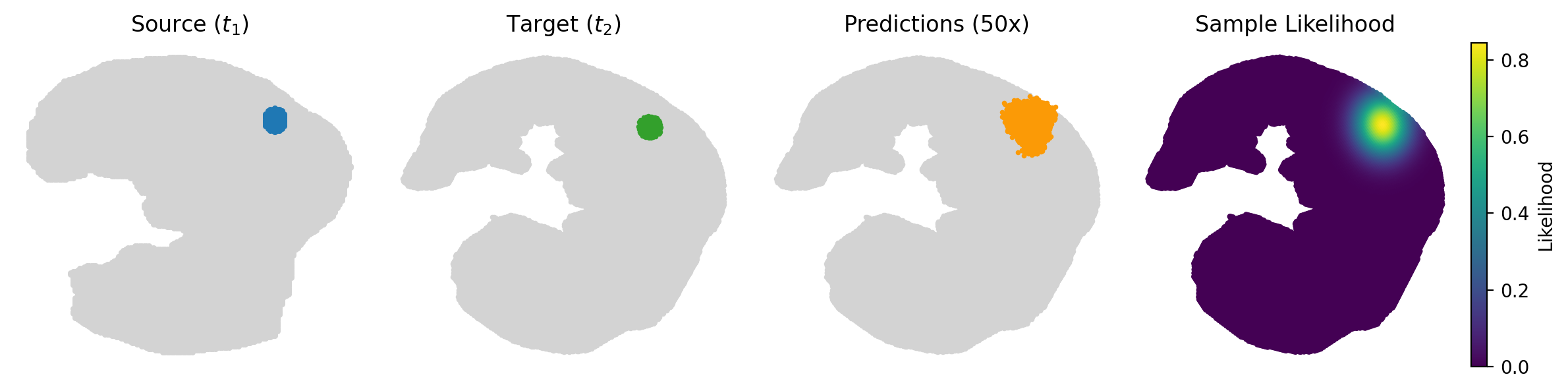}
    \caption{
        Qualitative evaluation of conditional generation with \modelname{} on the embryonic development dataset. Each row corresponds to a different source microenvironment at time $t_1$ (blue), shown alongside the ground truth target microenvironment at time $t_2$ (green). The third column displays 50 samples (orange) generated conditionally by the model, while the fourth column visualizes a kernel density estimate of the sample likelihood over the spatial domain.
    }
    \label{fig: conditioning-appendix}
\end{figure}
To assess whether the model accurately conditions on input microenvironments, we visualize the spatial distribution of generated samples given a fixed source microenvironment. \Cref{fig: conditioning-appendix} illustrates such cases on the embryonic development dataset, showing the input region at time $t_1$, the corresponding target at time $t_2$, and 50 independently generated samples from \modelname{}.
Directly computing the likelihood of ground truth cell coordinates under our generative model is intractable, as it would require evaluating the density of an implicitly defined distribution over point clouds. Instead, we approximate the spatial likelihood using kernel density estimation (KDE) over Monte Carlo samples drawn from the model. Given a set of generated coordinates $\{ \hat{\vc_i}\}_{i=1}^N$, we estimate the likelihood at a ground truth location $\vc$ as:
\begin{equation}
    \hat{p}(\vc) = \frac{1}{N} \sum_{i=1}^{N} \exp\left( -\frac{\| \vc - \hat{\vc}_i \|^2}{2\sigma^2} \right),
\end{equation}
where $\sigma$ is a fixed bandwidth parameter. The resulting KDE heatmap visualizes the spatial concentration of samples, allowing us to qualitatively assess whether the model produces consistent predictions conditioned on the source microenvironment.

\subsection{OT Ablation Study}
\label{appendix: ot-ablation-study}

We begin by emphasizing that the optimal choice of the OT feature-weighting parameter $\lambda$ in \cref{eq: ot-weighted} depends on the downstream application. This parameter determines the relative importance assigned to gene expression versus spatial coordinates during OT. For developmental processes such as organogenesis or regeneration, higher values of $\lambda$ prioritize transcriptional similarity, facilitating the reconstruction of continuous differentiation trajectories and capturing fate-driven transitions, which may span large spatial distances. In contrast, for applications that aim to monitor changes within a fixed spatial region, such as shifts in cell-type composition over time, a lower $\lambda$ is more appropriate. In such cases, emphasizing spatial locality helps avoid spurious long-range transport assignments caused by molecular noise.

\begin{table}[!htbp]
    \caption{
        Ablation study on the feature-coordinate trade-off parameter $\lambda$ in \cref{eq: ot-weighted} across mouse embryonic development (MED), axolotl brain development (ABD), and mouse brain aging (MBA). We report 1NN-F1, PSD, and SPD for each training objective (CFM, GVFM, GLVFM) and model variant (RPCFlow, NicheFlow). All results are averaged over five evaluation runs. Bold indicates the best value per dataset and metric.
    }
    \label{table: lambda-full-ablation}
    \centering
    \resizebox{\textwidth}{!}{%
        \begin{tabular}{lll
        ccc
        ccc
        ccc}
        \toprule
        \multicolumn{3}{l}{} &
        \multicolumn{3}{c}{MED} &
        \multicolumn{3}{c}{ABD} &
        \multicolumn{3}{c}{MBA} \\
        \cmidrule(lr){4-6} \cmidrule(lr){7-9} \cmidrule(lr){10-12}
        Model & Obj. & $\lambda$ & 1NN-F1~$\uparrow$ & PSD~$\downarrow$ & SPD~$\downarrow$
        & 1NN-F1~$\uparrow$ & PSD~$\downarrow$ & SPD~$\downarrow$
        & 1NN-F1~$\uparrow$ & PSD~$\downarrow$ & SPD~$\downarrow$ \\
        \midrule

        RPCFlow & CFM & $0.10$ & 0.546 {\scriptsize{± 0.0012}} & 0.981 {\scriptsize{± 0.0024}} & 0.564 {\scriptsize{± 0.0015}} & 0.524 {\scriptsize{± 0.0020}} & 2.051 {\scriptsize{± 0.0039}} & 1.015 {\scriptsize{± 0.0036}} & 0.271 {\scriptsize{± 0.0004}} & 1.543 {\scriptsize{± 0.0016}} & 0.810 {\scriptsize{± 0.0010}} \\
        RPCFlow & CFM & $0.25$ & 0.545 {\scriptsize{± 0.0004}} & 0.958 {\scriptsize{± 0.0040}} & 0.570 {\scriptsize{± 0.0011}} & 0.511 {\scriptsize{± 0.0015}} & 2.079 {\scriptsize{± 0.0053}} & 1.024 {\scriptsize{± 0.0046}} & 0.273 {\scriptsize{± 0.0005}} & \textbf{1.535 {\scriptsize{± 0.0009}}} & 0.818 {\scriptsize{± 0.0013}} \\
        RPCFlow & CFM & $0.50$ & 0.554 {\scriptsize{± 0.0009}} & 0.988 {\scriptsize{± 0.0031}} & 0.562 {\scriptsize{± 0.0011}} & 0.507 {\scriptsize{± 0.0011}} & 2.077 {\scriptsize{± 0.0065}} & 1.023 {\scriptsize{± 0.0032}} & 0.273 {\scriptsize{± 0.0006}} & 1.546 {\scriptsize{± 0.0010}} & 0.810 {\scriptsize{± 0.0013}} \\
        RPCFlow & CFM & $0.75$ & 0.537 {\scriptsize{± 0.0007}} & 0.981 {\scriptsize{± 0.0020}} & 0.595 {\scriptsize{± 0.0022}} & 0.517 {\scriptsize{± 0.0005}} & 2.033 {\scriptsize{± 0.0072}} & 1.026 {\scriptsize{± 0.0050}} & 0.275 {\scriptsize{± 0.0003}} & 1.553 {\scriptsize{± 0.0008}} & 0.801 {\scriptsize{± 0.0010}} \\

        \midrule
        
        RPCFlow & GLVFM & $0.10$ & 0.586 {\scriptsize{± 0.0016}} & 0.979 {\scriptsize{± 0.0021}} & 0.586 {\scriptsize{± 0.0012}} & 0.554 {\scriptsize{± 0.0007}} & 2.053 {\scriptsize{± 0.0044}} & 1.038 {\scriptsize{± 0.0025}} & 0.265 {\scriptsize{± 0.0004}} & 1.723 {\scriptsize{± 0.0015}} & 0.779 {\scriptsize{± 0.0011}} \\
        RPCFlow & GLVFM & $0.25$ & 0.593 {\scriptsize{± 0.0003}} & 0.924 {\scriptsize{± 0.0019}} & 0.575 {\scriptsize{± 0.0017}} & 0.555 {\scriptsize{± 0.0013}} & 2.076 {\scriptsize{± 0.0057}} & 1.036 {\scriptsize{± 0.0024}} & 0.267 {\scriptsize{± 0.0003}} & 1.728 {\scriptsize{± 0.0016}} & 0.777 {\scriptsize{± 0.0005}} \\
        RPCFlow & GLVFM & $0.50$ & 0.586 {\scriptsize{± 0.0013}} & 0.934 {\scriptsize{± 0.0019}} & 0.569 {\scriptsize{± 0.0014}} & 0.551 {\scriptsize{± 0.0008}} & 2.038 {\scriptsize{± 0.0045}} & 1.032 {\scriptsize{± 0.0034}} & 0.269 {\scriptsize{± 0.0002}} & 1.715 {\scriptsize{± 0.0014}} & 0.783 {\scriptsize{± 0.0006}} \\
        RPCFlow & GLVFM & $0.75$ & 0.593 {\scriptsize{± 0.0009}} & 0.948 {\scriptsize{± 0.0035}} & 0.570 {\scriptsize{± 0.0011}} & 0.561 {\scriptsize{± 0.0008}} & \textbf{2.014 {\scriptsize{± 0.0056}}} & 1.035 {\scriptsize{± 0.0047}} & 0.268 {\scriptsize{± 0.0005}} & 1.675 {\scriptsize{± 0.0019}} & 0.784 {\scriptsize{± 0.0013}} \\

        \midrule
        
        RPCFlow & GVFM & $0.10$ & 0.503 {\scriptsize{± 0.0013}} & 1.155 {\scriptsize{± 0.0044}} & 0.578 {\scriptsize{± 0.0007}} & 0.477 {\scriptsize{± 0.0008}} & 2.260 {\scriptsize{± 0.0077}} & 1.036 {\scriptsize{± 0.0031}} & 0.249 {\scriptsize{± 0.0003}} & 1.753 {\scriptsize{± 0.0020}} & 0.784 {\scriptsize{± 0.0010}} \\
        RPCFlow & GVFM & $0.25$ & 0.520 {\scriptsize{± 0.0010}} & 1.223 {\scriptsize{± 0.0044}} & 0.566 {\scriptsize{± 0.0011}} & 0.478 {\scriptsize{± 0.0014}} & 2.364 {\scriptsize{± 0.0065}} & 1.030 {\scriptsize{± 0.0042}} & 0.246 {\scriptsize{± 0.0005}} & 1.710 {\scriptsize{± 0.0013}} & 0.787 {\scriptsize{± 0.0014}} \\
        RPCFlow & GVFM & $0.50$ & 0.521 {\scriptsize{± 0.0012}} & 1.185 {\scriptsize{± 0.0032}} & 0.569 {\scriptsize{± 0.0009}} & 0.480 {\scriptsize{± 0.0008}} & 2.360 {\scriptsize{± 0.0036}} & 1.025 {\scriptsize{± 0.0015}} & 0.245 {\scriptsize{± 0.0004}} & 1.756 {\scriptsize{± 0.0025}} & 0.774 {\scriptsize{± 0.0007}} \\
        RPCFlow & GVFM & $0.75$ & 0.514 {\scriptsize{± 0.0013}} & 1.202 {\scriptsize{± 0.0014}} & 0.573 {\scriptsize{± 0.0008}} & 0.471 {\scriptsize{± 0.0012}} & 2.476 {\scriptsize{± 0.0082}} & 1.037 {\scriptsize{± 0.0036}} & 0.248 {\scriptsize{± 0.0003}} & 1.831 {\scriptsize{± 0.0015}} & 0.780 {\scriptsize{± 0.0012}} \\

        \midrule
        
        NicheFlow & CFM & $0.10$ & 0.609 {\scriptsize{± 0.0030}} & 0.979 {\scriptsize{± 0.0228}} & 0.402 {\scriptsize{± 0.0036}} & 0.604 {\scriptsize{± 0.0018}} & 2.086 {\scriptsize{± 0.0058}} & 0.568 {\scriptsize{± 0.0030}} & 0.283 {\scriptsize{± 0.0003}} & 1.557 {\scriptsize{± 0.0014}} & 0.556 {\scriptsize{± 0.0028}} \\
        NicheFlow & CFM & $0.25$ & 0.569 {\scriptsize{± 0.0031}} & 0.973 {\scriptsize{± 0.0074}} & 0.425 {\scriptsize{± 0.0062}} & 0.586 {\scriptsize{± 0.0013}} & 2.106 {\scriptsize{± 0.0072}} & \textbf{0.565 {\scriptsize{± 0.0039}}} & 0.281 {\scriptsize{± 0.0006}} & 1.546 {\scriptsize{± 0.0029}} & 0.612 {\scriptsize{± 0.0032}} \\
        NicheFlow & CFM & $0.50$ & 0.551 {\scriptsize{± 0.0009}} & 1.051 {\scriptsize{± 0.0496}} & 0.471 {\scriptsize{± 0.0110}} & 0.585 {\scriptsize{± 0.0013}} & 2.089 {\scriptsize{± 0.0066}} & 0.591 {\scriptsize{± 0.0026}} & 0.283 {\scriptsize{± 0.0002}} & 1.588 {\scriptsize{± 0.0033}} & 0.604 {\scriptsize{± 0.0061}} \\
        NicheFlow & CFM & $0.75$ & 0.519 {\scriptsize{± 0.0038}} & 1.103 {\scriptsize{± 0.0323}} & 0.515 {\scriptsize{± 0.0101}} & 0.571 {\scriptsize{± 0.0012}} & 2.126 {\scriptsize{± 0.0110}} & 0.592 {\scriptsize{± 0.0044}} & 0.278 {\scriptsize{± 0.0003}} & 1.566 {\scriptsize{± 0.0027}} & 0.616 {\scriptsize{± 0.0041}} \\

        \midrule
        
        NicheFlow & GVFM & $0.10$ & 0.596 {\scriptsize{± 0.0027}} & 0.991 {\scriptsize{± 0.0137}} & 0.406 {\scriptsize{± 0.0025}} & 0.574 {\scriptsize{± 0.0015}} & 2.220 {\scriptsize{± 0.0107}} & 0.594 {\scriptsize{± 0.0046}} & 0.268 {\scriptsize{± 0.0003}} & 1.661 {\scriptsize{± 0.0033}} & \textbf{0.531 {\scriptsize{± 0.0010}}} \\
        NicheFlow & GVFM & $0.25$ & 0.563 {\scriptsize{± 0.0027}} & 1.051 {\scriptsize{± 0.0117}} & 0.491 {\scriptsize{± 0.0105}} & 0.571 {\scriptsize{± 0.0009}} & 2.343 {\scriptsize{± 0.0136}} & 0.619 {\scriptsize{± 0.0061}} & 0.265 {\scriptsize{± 0.0006}} & 1.599 {\scriptsize{± 0.0032}} & 0.590 {\scriptsize{± 0.0018}} \\
        NicheFlow & GVFM & $0.50$ & 0.533 {\scriptsize{± 0.0053}} & 1.034 {\scriptsize{± 0.0452}} & 0.800 {\scriptsize{± 0.0401}} & 0.556 {\scriptsize{± 0.0016}} & 2.283 {\scriptsize{± 0.0126}} & 0.742 {\scriptsize{± 0.0134}} & 0.269 {\scriptsize{± 0.0005}} & 1.607 {\scriptsize{± 0.0029}} & 0.605 {\scriptsize{± 0.0022}} \\
        NicheFlow & GVFM & $0.75$ & 0.526 {\scriptsize{± 0.0028}} & 1.121 {\scriptsize{± 0.0112}} & 0.870 {\scriptsize{± 0.0336}} & 0.556 {\scriptsize{± 0.0013}} & 2.166 {\scriptsize{± 0.0091}} & 0.684 {\scriptsize{± 0.0128}} & 0.263 {\scriptsize{± 0.0004}} & 1.613 {\scriptsize{± 0.0031}} & 0.731 {\scriptsize{± 0.0040}} \\

        \midrule
        
        NicheFlow & GLVFM & $0.10$ & \textbf{0.664 {\scriptsize{± 0.0014}}} & 0.883 {\scriptsize{± 0.0094}} & 0.398 {\scriptsize{± 0.0023}} & \textbf{0.628 {\scriptsize{± 0.0013}}} & 2.079 {\scriptsize{± 0.0043}} & 0.576 {\scriptsize{± 0.0055}} & \textbf{0.285 {\scriptsize{± 0.0003}}} & 1.554 {\scriptsize{± 0.0021}} & 0.532 {\scriptsize{± 0.0009}} \\
        NicheFlow & GLVFM & $0.25$ & 0.629 {\scriptsize{± 0.0033}} & 0.923 {\scriptsize{± 0.0109}} & \textbf{0.394 {\scriptsize{± 0.0035}}} & 0.618 {\scriptsize{± 0.0014}} & 2.102 {\scriptsize{± 0.0028}} & 0.577 {\scriptsize{± 0.0051}} & 0.284 {\scriptsize{± 0.0004}} & 1.599 {\scriptsize{± 0.0035}} & 0.549 {\scriptsize{± 0.0012}} \\
        NicheFlow & GLVFM & $0.50$ & 0.610 {\scriptsize{± 0.0036}} & 0.909 {\scriptsize{± 0.0107}} & 0.417 {\scriptsize{± 0.0060}} & 0.610 {\scriptsize{± 0.0008}} & 2.136 {\scriptsize{± 0.0025}} & 0.579 {\scriptsize{± 0.0014}} & 0.283 {\scriptsize{± 0.0005}} & 1.600 {\scriptsize{± 0.0039}} & 0.546 {\scriptsize{± 0.0009}} \\
        NicheFlow & GLVFM & $0.75$ & 0.592 {\scriptsize{± 0.0019}} & \textbf{0.879 {\scriptsize{± 0.0054}}} & 0.472 {\scriptsize{± 0.0150}} & 0.603 {\scriptsize{± 0.0014}} & 2.106 {\scriptsize{± 0.0060}} & 0.592 {\scriptsize{± 0.0073}} & 0.281 {\scriptsize{± 0.0004}} & 1.573 {\scriptsize{± 0.0008}} & 0.595 {\scriptsize{± 0.0026}} \\
        
        \bottomrule
        \end{tabular}%
    }
\end{table}

\begin{figure}[!htb]
    \centering
    \includegraphics[width=0.9\linewidth]{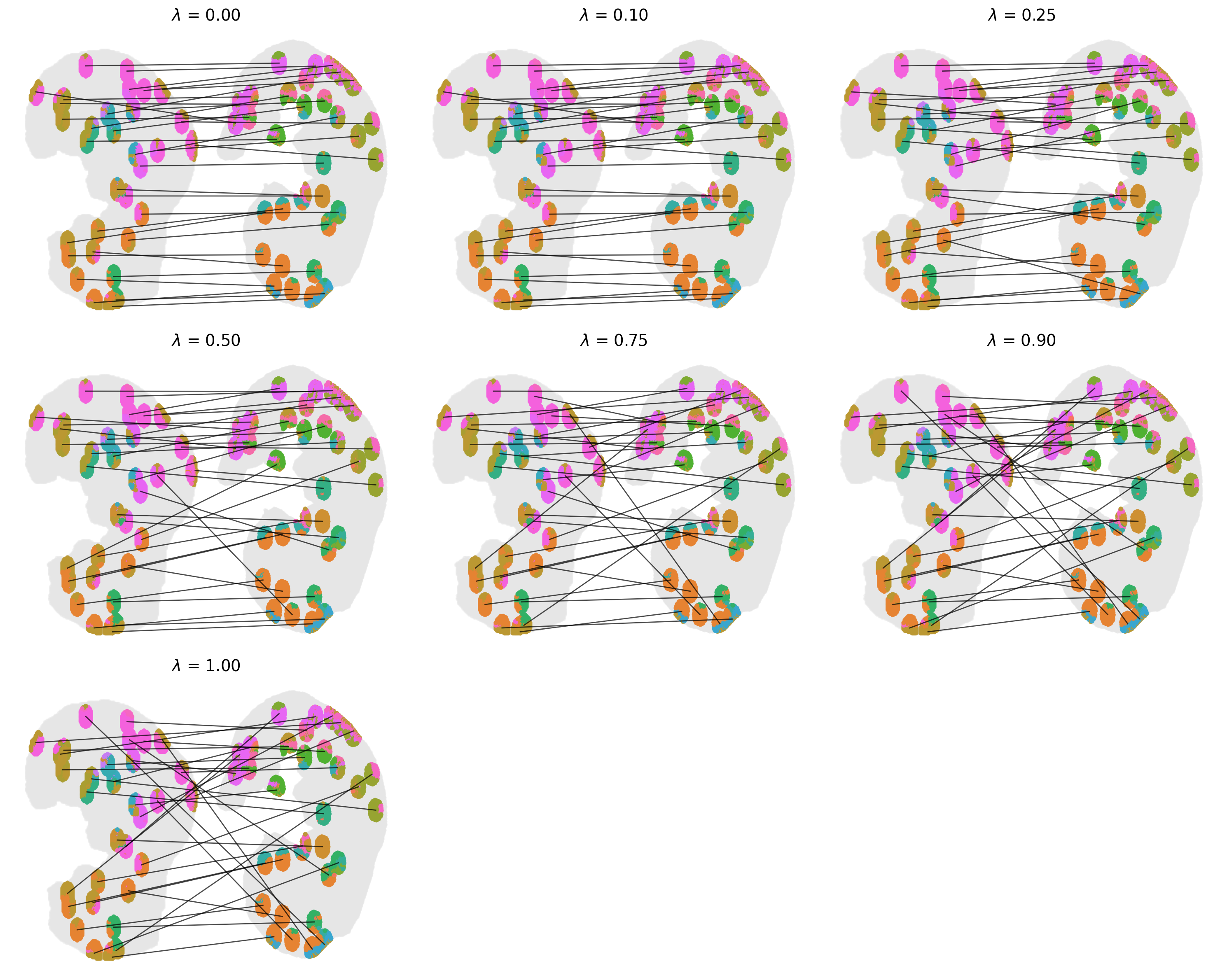}
\caption{
        Visualization of OT couplings computed under varying values of the pooling parameter $\lambda$ in eq. (\ref{eq: ot-weighted}), which balances spatial coordinates and gene expression in microenvironment matching. Lower $\lambda$ values prioritize spatial proximity, resulting in more dispersed and less structured alignments, while intermediate values 
        yield tighter, biologically consistent mappings. Very high $\lambda$ settings ignore spatial context and may lead to implausible long-range matches.
    }
    \label{fig: moscot-ot-lambdas}
\end{figure}
In \cref{table: lambda-full-ablation}, we sweep the value of $\lambda$ across multiple models and training strategies on the same task as in \cref{sec: quant_eval}, showing consistent results across settings as reported in \cref{table: main-results}. 

Furthermore, \Cref{fig: moscot-ot-lambdas} visualizes the impact of varying $\lambda$ on the OT plans described in \cref{sec: pc_ot}, where intermediate values yield more coherent and biologically plausible couplings.

\subsection{$K$-Means regions ablation study}
\label{appendix: kmeans-ablation-study}

To ensure diverse and spatially distributed sampling during training, we partition each tissue section into $K$ spatial regions using $K$-Means clustering over the 2D cell coordinates (see \cref{sec: experimental-setup}). At each training step, microenvironments are sampled uniformly from within these regions, encouraging broad spatial coverage and preventing oversampling of densely populated areas. In this ablation study, we investigate how varying the number of spatial regions, $K$, affects model performance.

\begin{table}[!htbp]
    \caption{
        Ablation study of the number of spatial regions $K$ defined over the datasets. We evaluate \modelname{} with the GLVFM objective across three datasets: mouse embryonic development (MED), axolotl brain development (ABD), and mouse brain aging (MBA). Results are reported as mean ± standard deviation over five evaluation runs.
    }
    \label{table: k-regions-ablations-multidataset}
    \centering
    \resizebox{\textwidth}{!}{%
    \begin{tabular}{l
        ccc
        ccc
        ccc
    }
        \toprule
        \multirow{2}{*}{$K$} &
        \multicolumn{3}{c}{MED} &
        \multicolumn{3}{c}{ABD} &
        \multicolumn{3}{c}{MBA} \\
        \cmidrule(lr){2-4} \cmidrule(lr){5-7} \cmidrule(lr){8-10}
        & 1NN-F1~$\uparrow$ & PSD~$\downarrow$ ($10^2$) & SPD~$\downarrow$ ($10^2$) 
        & 1NN-F1~$\uparrow$ & PSD~$\downarrow$ ($10^2$) & SPD~$\downarrow$ ($10^2$) 
        & 1NN-F1~$\uparrow$ & PSD~$\downarrow$ ($10^2$) & SPD~$\downarrow$ ($10^2$) \\
        \midrule

        8 & 0.640 {\scriptsize{± 0.0043}} & \textbf{0.876 {\scriptsize{± 0.0055}}} & 0.441 {\scriptsize{± 0.0058}} & 0.617 {\scriptsize{± 0.0017}} & 1.954 {\scriptsize{± 0.0057}} & 0.631 {\scriptsize{± 0.0073}} & 0.283 {\scriptsize{± 0.0005}} & 1.561 {\scriptsize{± 0.0018}} & 0.571 {\scriptsize{± 0.0019}} \\
        16 & 0.661 {\scriptsize{± 0.0033}} & 0.881 {\scriptsize{± 0.0068}} & 0.393 {\scriptsize{± 0.0056}} & \textbf{0.633 {\scriptsize{± 0.0008}}} & \textbf{1.936 {\scriptsize{± 0.0027}}} & 0.628 {\scriptsize{± 0.0104}} & 0.283 {\scriptsize{± 0.0003}} & 1.564 {\scriptsize{± 0.0035}} & 0.538 {\scriptsize{± 0.0021}} \\
        32 & 0.659 {\scriptsize{± 0.0025}} & 0.899 {\scriptsize{± 0.0120}} & \textbf{0.391 {\scriptsize{± 0.0029}}} & 0.622 {\scriptsize{± 0.0005}} & 1.968 {\scriptsize{± 0.0036}} & 0.640 {\scriptsize{± 0.0124}} & 0.279 {\scriptsize{± 0.0005}} & 1.600 {\scriptsize{± 0.0040}} & 0.537 {\scriptsize{± 0.0015}} \\
        64 & \textbf{0.664 {\scriptsize{± 0.0014}}} & 0.883 {\scriptsize{± 0.0094}} & 0.398 {\scriptsize{± 0.0023}} & 0.628 {\scriptsize{± 0.0013}} & 2.079 {\scriptsize{± 0.0043}} & \textbf{0.576 {\scriptsize{± 0.0055}}} & \textbf{0.285 {\scriptsize{± 0.0003}}} & \textbf{1.554 {\scriptsize{± 0.0021}}} & \textbf{0.532 {\scriptsize{± 0.0009}}} \\
        \bottomrule
    \end{tabular}%
    }
\end{table}


        
        
%

\begin{figure}[!htb]
    \centering
    \includegraphics[width=0.55\linewidth]{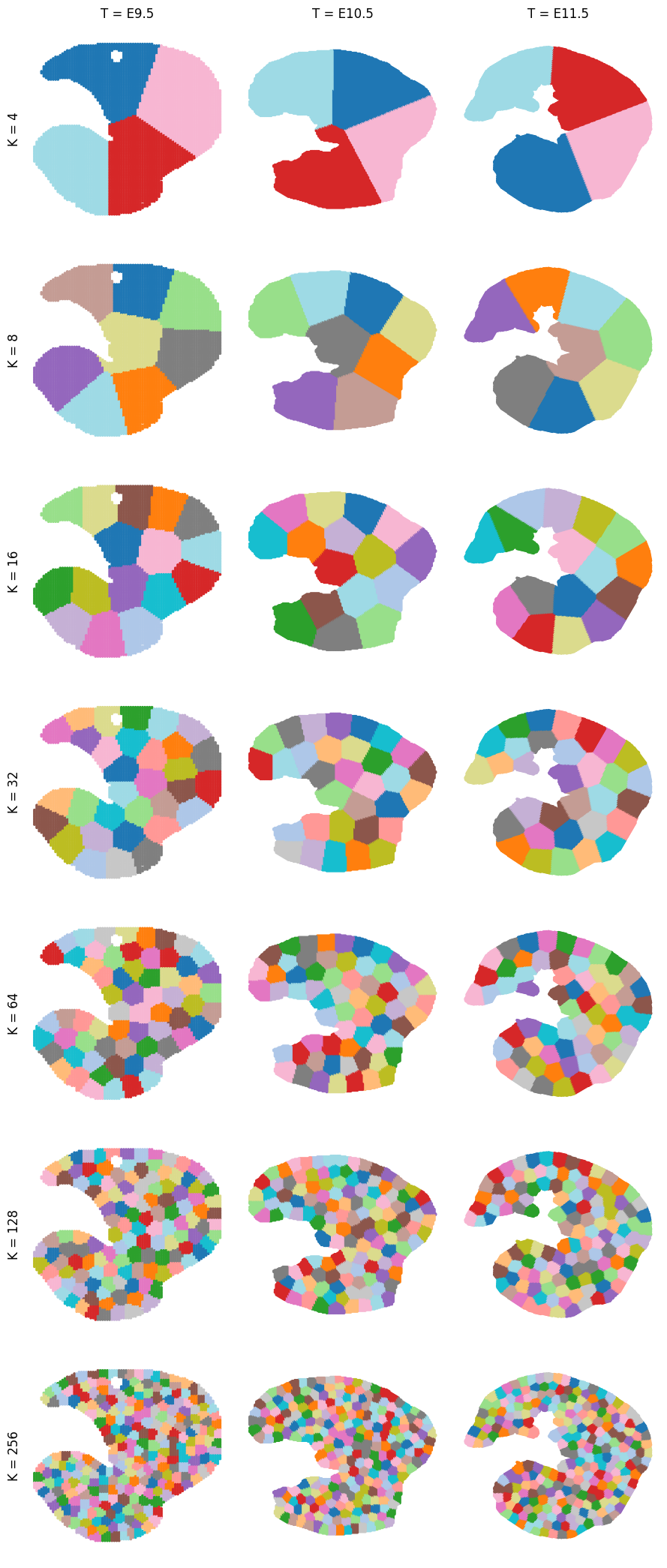}
    \caption{
        Visualization of spatial partitions obtained via KMeans clustering with $K = \{4, 8, 16, 32, 64, 128, 256\}$ on the embryonic development dataset. For low $K$, each region covers large heterogeneous areas; for high $K$, regions become small, dense, and highly overlapping, potentially degrading the diversity and utility of sampled microenvironments.
    }
    \label{fig: kmeans-regions}
\end{figure}

As shown in \Cref{fig: kmeans-regions}, increasing $K$ leads to increasingly fine-grained spatial partitions. While moderate values of $K$ help improve spatial resolution, excessively high values (e.g., $K = 128$ or $256$) result in overly small and fragmented regions. This can cause significant overlap between sampled microenvironments and reduce sampling diversity. Moreover, in sparsely populated tissue sections, high $K$ values may yield regions with insufficient cells, degrading both representativeness and stability.

Conversely, using too few regions (e.g., $K = 4$) results in broad spatial partitions that may encompass multiple heterogeneous tissue compartments. This undermines the locality assumptions of our model and increases intra-region variability, which can impair the model’s ability to learn.

Given the trade-offs outlined above, we focus our evaluation on $K \in \{8, 16, 32, 64\}$, which spans a range of granularities that preserve both spatial interpretability and sampling robustness. \cref{table: k-regions-ablations-multidataset} summarizes the quantitative results for this ablation across three datasets: mouse embryonic development (MED), axolotl brain development (ABD), and mouse brain aging (MBA). We observe that using $K = 64$ consistently yields strong performance, achieving the highest 1NN-F1 scores on both the MED and MBA datasets, while also performing competitively on ABD. These findings indicate that $K = 64$ offers an effective balance between spatial resolution and stability, and we adopt it as the default configuration in our main experiments.

\subsection{Justifying the choice of \modelname{} over RPCFlow}
\label{appendix: stcmflow-vs-rpcflow}
\begin{figure}[!htb]
    \centering
    \includegraphics[width=0.9\linewidth]{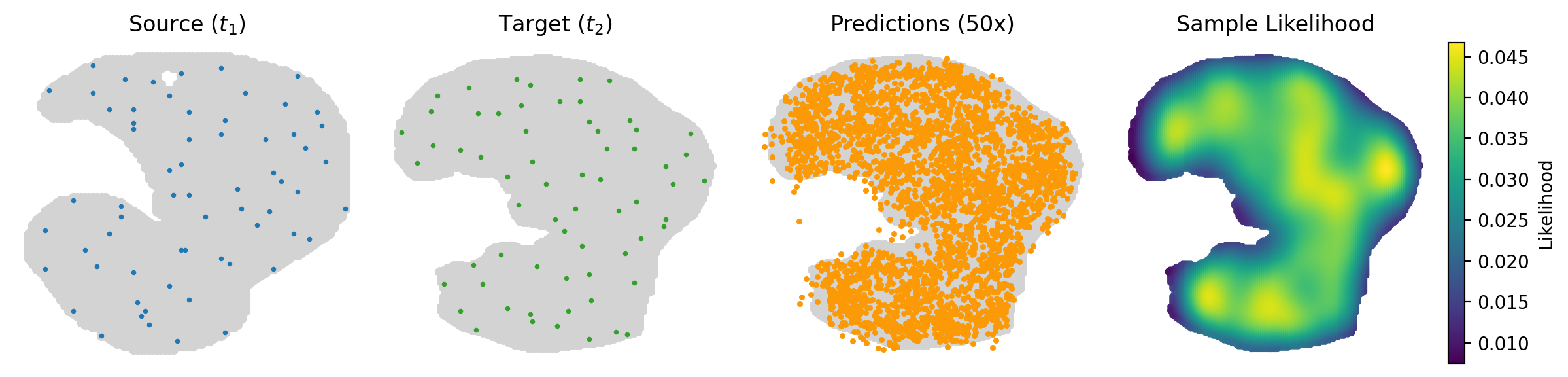}
    \includegraphics[width=0.9\linewidth]{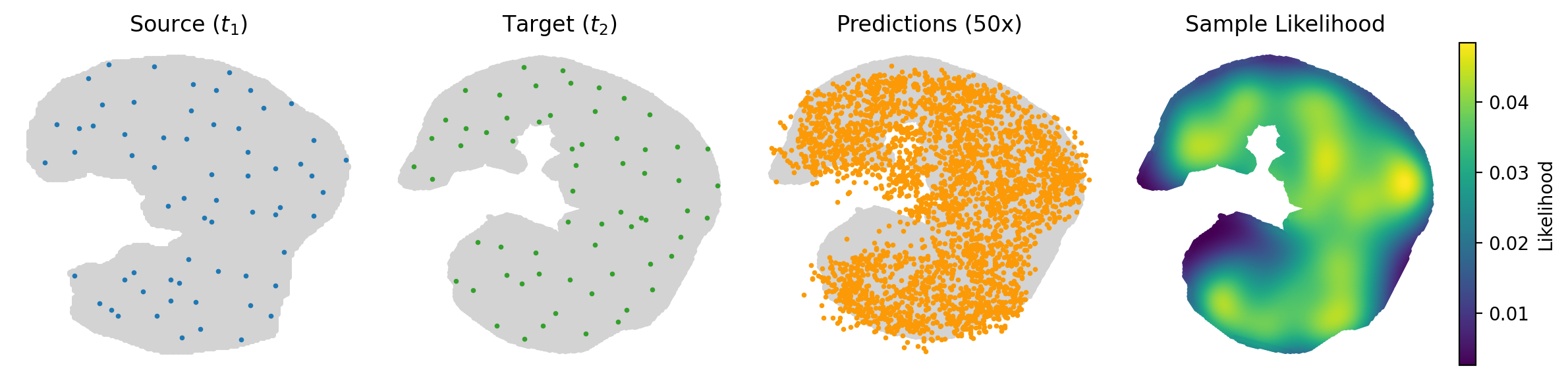}
    \caption{
        RPCFlow fails to perform meaningful conditional generation: the model generates a diffuse reconstruction resembling the entire target tissue rather than conditioning on the provided source points. Each row shows a different source-target pair.
    }
    \label{fig: rpcflow-whole-target}
\end{figure}

Although RPCFlow achieves competitive performance on spatial metrics such as PSD and SPD, it lacks the essential capability of meaningful conditional generation. In RPCFlow, conditioning is performed using randomly sampled point clouds from the spatial regions without explicit microenvironment structure. As shown in \cref{fig: rpcflow-whole-target}, when conditioned on such random sources, RPCFlow tends to reconstruct the entire target tissue, rather than capturing local dynamics driven by the source input. This undermines its ability to model spatiotemporal evolution in a biologically grounded manner.

\begin{figure}[!htb]
    \centering

    \begin{subfigure}{0.9\linewidth}
        \label{fig: microenv-stcmflow-1}
        \centering
        \includegraphics[width=\linewidth]{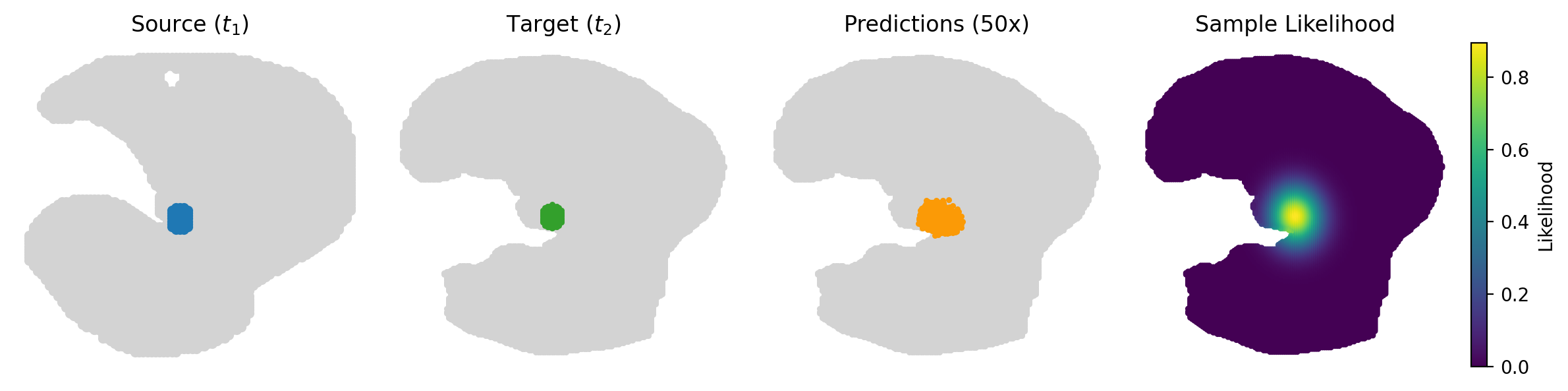}
        \caption{\modelname{} – Sample 1}
    \end{subfigure}

    \vspace{1em} 

    \begin{subfigure}{0.9\linewidth}
        \label{fig: microenv-stcmflow-2}
        \centering
        \includegraphics[width=\linewidth]{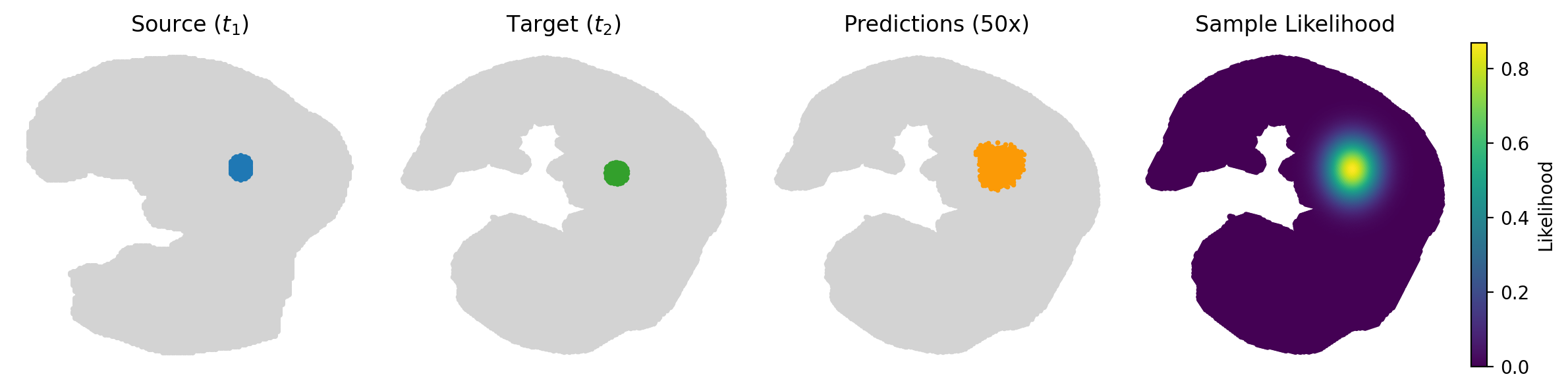}
        \caption{\modelname{} – Sample 2}
    \end{subfigure}

    \vspace{1em}

    \begin{subfigure}{0.9\linewidth}
        \label{fig: microenv-rpcflow-1}
        \centering
        \includegraphics[width=\linewidth]{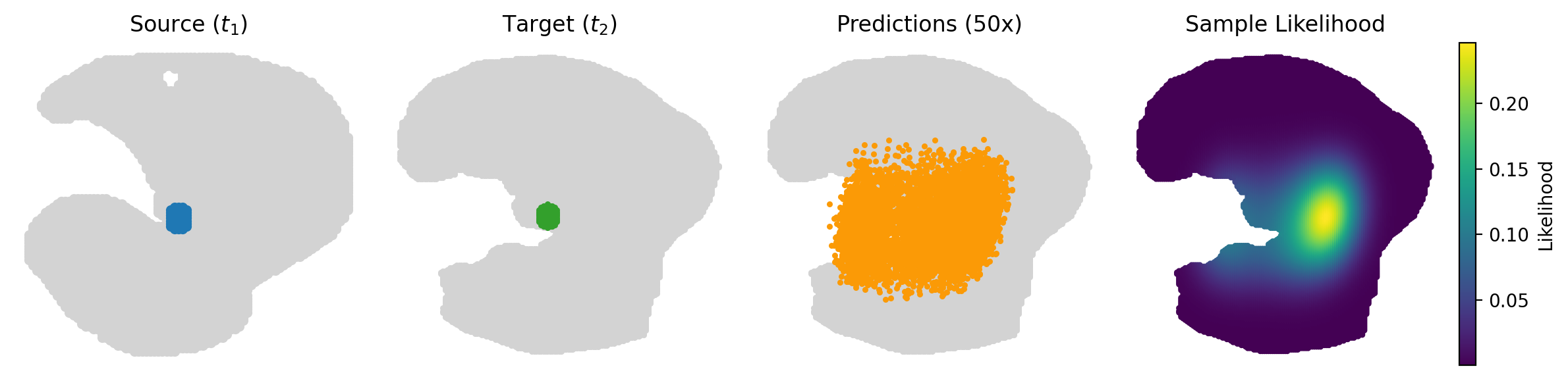}
        \caption{RPCFlow – Sample 1}
    \end{subfigure}

    \vspace{1em}

    \begin{subfigure}{0.9\linewidth}
        \label{fig: microenv-rpcflow-2}
        \centering
        \includegraphics[width=\linewidth]{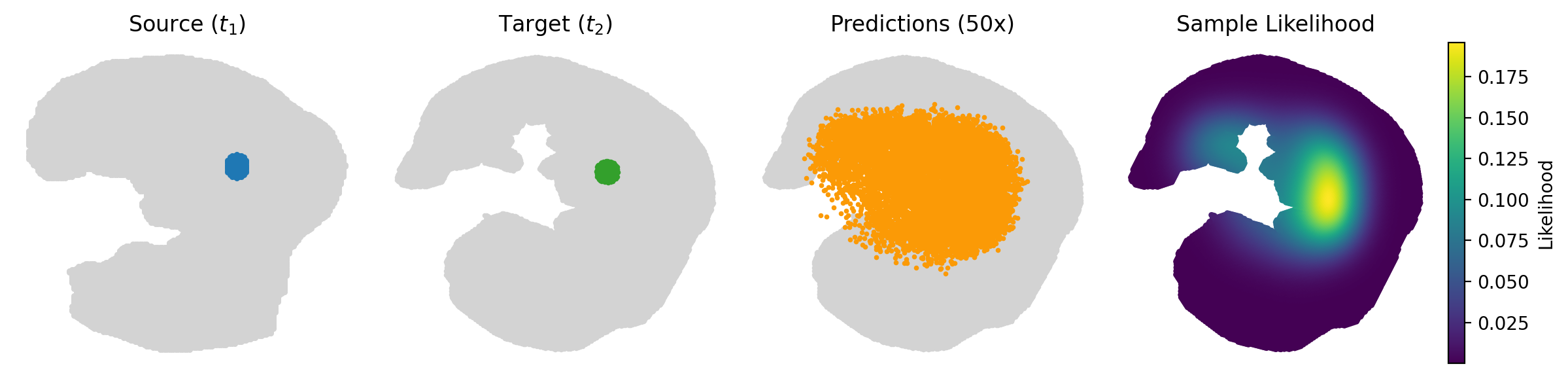}
        \caption{RPCFlow – Sample 2}
    \end{subfigure}

    \caption{
        Comparison between \modelname{} and RPCFlow in a fixed-source microenvironment setting. Each row shows input (source), ground truth target, model prediction, and KDE-estimated likelihood. While \modelname{} (top) produces well-localized samples consistent with the input, RPCFlow (bottom) generates less structured and spatially inaccurate predictions.
    }
    \label{fig: rpcflow-vs-stcmflow}
\end{figure}

In contrast, \modelname{} is explicitly designed to push localized microenvironments through time. By conditioning on fixed-radius neighborhoods centered around specific spatial positions, the model learns how cellular contexts evolve, preserving both spatial coherence and transcriptional identity. \Cref{fig: rpcflow-vs-stcmflow} visualizes this distinction: while \modelname{} consistently generates well-localized predictions aligned with the input microenvironment, RPCFlow fails to retain spatial specificity, often diffusing the prediction across broader regions.

From a biological perspective, predicting the fate of a local tissue region over time is far more relevant than mapping random point sets. Microenvironments encode structured cellular contexts, such as signaling interactions or niche-specific cell states, that are crucial for downstream analysis (e.g., lineage fate prediction, microenvironment-based intervention simulation). Because RPCFlow lacks this interpretability and fails to enable such downstream tasks, it cannot serve as a practical generative model in biological settings.

In summary, while RPCFlow may appear performant under some aggregate spatial metrics, only \modelname{} enables localized, conditionally consistent generation of evolving tissue regions. This makes it not only superior for evaluation but also for practical use in biological modeling.

\subsection{Structure-aware evaluation}
While we refer to our primary evaluation metrics as PSD and SPD, they correspond to the two asymmetric directions of the Chamfer Distance (CD) - a widely used metric in point cloud generation \citep{chamferdistance}. PSD measures how closely the predicted points adhere to the ground truth structure (fidelity), while SPD captures how well the prediction covers the full extent of the target (coverage). We compute both metrics over collections of microenvironments tiled across the tissue, providing an indirect assessment of global morphological accuracy.

To directly assess the preservation of internal structure within generated microenvironments, we additionally evaluate Gromov-Wasserstein (GW) and Fused Gromov-Wasserstein (FGW) distances \citep{gw,vayer2020fused}. These metrics compare the intra-point pairwise distances in predicted and ground truth point clouds, capturing latent structural relationships. FGW further incorporates transcriptomic similarity, offering a holistic measure of structural and feature-based alignment.

Given the computational cost of GW and FGW, we restrict their evaluation to microenvironment-level generations. For each source niche, we sample predictions from the trained model and compute GW/FGW against the corresponding real microenvironment, averaging results across multiple runs. The source–target pairs are obtained by computing OT between all microenvironments at time $t$ and $t+1$, yielding a one-to-one coupling used for conditioning and evaluation.

As SPFlow generates individual points and RPCFlow produces randomly sampled, unstructured clouds, these models lack coherent internal structure and are not amenable to structural comparison via GW or FGW. We therefore report these metrics only for models that generate full microenvironments.

The results in \cref{table:gw-fgw} show that \modelname{}, particularly when trained with the GLVFM objective, consistently achieves lower GW and FGW distances compared to baseline models. These findings underscore the structural fidelity of our predictions and further validate the modeling advantages of microenvironment-aware generative training.
\begin{table}[!htbp]
    \caption{
        Gromov–Wasserstein (GW) and Fused Gromov–Wasserstein (FGW) distances between generated and real microenvironments on mouse embryonic development (MED), axolotl brain development (ABD), and mouse brain aging (MBA). All models are trained with a fixed $\lambda = 0.1$. Results are averaged over five generation runs. Bold indicates the best (lowest) value per column.
    }
    \label{table:gw-fgw}
    \centering
    \resizebox{\textwidth}{!}{%
    \begin{tabular}{llcccccc}
        \toprule
        \multicolumn{2}{l}{} & \multicolumn{2}{c}{MED} & \multicolumn{2}{c}{ABD} & \multicolumn{2}{c}{MBA} \\
        \cmidrule(lr){3-4} \cmidrule(lr){5-6} \cmidrule(lr){7-8}
        Model & Objective  & GW~$(10^2)$ & FGW  & GW~$(10^2)$ & FGW  & GW~$(10^2)$ & FGW \\
        \midrule
        NicheFlow & CFM & 0.315 {\scriptsize{± 0.003}} & 3.273 {\scriptsize{± 0.003}} & 0.783 {\scriptsize{± 0.002}} & 3.543 {\scriptsize{± 0.004}} & 0.315 {\scriptsize{± 0.001}} & 3.531 {\scriptsize{± 0.000}} \\
        
        NicheFlow & GVFM & 0.463 {\scriptsize{± 0.004}} & 3.167 {\scriptsize{± 0.007}} & 0.797 {\scriptsize{± 0.006}} & \textbf{3.414 {\scriptsize{± 0.003}}} & 0.339 {\scriptsize{± 0.000}} & 3.387 {\scriptsize{± 0.000}} \\
        
        NicheFlow & GLVFM & \textbf{0.224 {\scriptsize{± 0.001}}} & \textbf{3.147 {\scriptsize{± 0.007}}} & \textbf{0.720 {\scriptsize{± 0.004}}} & 3.420 {\scriptsize{± 0.003}} & \textbf{0.262 {\scriptsize{± 0.000}}} & \textbf{3.367 {\scriptsize{± 0.000}}} \\
    \bottomrule
    \end{tabular}%
    }
\end{table}

\subsection{Wasserstein metrics}
To complement the local structural assessments with GW and FGW, we additionally include global evaluation metrics that are widely adopted in spatial transcriptomics \citep{wasserstein-ex-trajectorynet,wasserstein-ex-mioflow,wasserstein-ex-deepruot}. Specifically, we report the 1-Wasserstein ($\mathcal{W}_1$) and 2-Wasserstein ($\mathcal{W}_2$) distances between predicted and real cell distributions, computed separately over spatial coordinates and gene expression features. These metrics are calculated at the level of individual cell types and averaged across all timepoints and generated samples. This provides a more global perspective on how well the model reconstructs tissue-wide spatial and transcriptional distributions.

\begin{table}[!htbp]
    \caption{
        Comparison of Wasserstein distances between generated and real tissue structures on the mouse embryonic development (MED) dataset. We compute $\mathcal{W}_1$ and $\mathcal{W}_2$ distances separately for spatial coordinates (Pos.) and gene expression features (Genes), averaged across all generated cell types and timepoints. All models are trained with a fixed value of $\lambda = 0.1$, and results are averaged over five generation runs. Bold indicates the best (lowest) value per column.
    }
    \label{table:wasserstein-med}
    \centering
    \resizebox{0.75\textwidth}{!}{%
    \begin{tabular}{llcccc}
        \toprule
        Model & Obj. & $\mathcal{W}_1$ Pos. & $\mathcal{W}_1$ Genes & $\mathcal{W}_2$ Pos. & $\mathcal{W}_2$ Genes \\
        \midrule
        SPFlow & CFM & 0.634 {\scriptsize{± 0.004}} & 6.545 {\scriptsize{± 0.004}} & 0.773 {\scriptsize{± 0.005}} & 5.585 {\scriptsize{± 0.005}} \\
        SPFlow & GVFM & 0.612 {\scriptsize{± 0.005}} & 6.517 {\scriptsize{± 0.002}} & 0.743 {\scriptsize{± 0.005}} & 5.609 {\scriptsize{± 0.113}} \\
        SPFlow & GLVFM & 0.613 {\scriptsize{± 0.001}} & 6.457 {\scriptsize{± 0.003}} & 0.738 {\scriptsize{± 0.001}} & 5.546 {\scriptsize{± 0.108}} \\
        \midrule
        
        RPCFlow & CFM & 0.290 {\scriptsize{± 0.006}} & 6.303 {\scriptsize{± 0.004}} & 0.460 {\scriptsize{± 0.007}} & 5.386 {\scriptsize{± 0.004}} \\
        RPCFlow & GVFM & 0.258 {\scriptsize{± 0.004}} & 6.134 {\scriptsize{± 0.004}} & 0.398 {\scriptsize{± 0.006}} & 5.292 {\scriptsize{± 0.114}} \\
        RPCFlow & GLVFM & 0.221 {\scriptsize{± 0.003}} & 6.087 {\scriptsize{± 0.002}} & 0.365 {\scriptsize{± 0.007}} & 5.244 {\scriptsize{± 0.113}} \\
        \midrule
        
        NicheFlow & CFM & 0.253 {\scriptsize{± 0.002}} & 6.177 {\scriptsize{± 0.001}} & 0.420 {\scriptsize{± 0.003}} & 5.314 {\scriptsize{± 0.111}} \\
        NicheFlow & GVFM & 0.222 {\scriptsize{± 0.004}} & 5.985 {\scriptsize{± 0.005}} & 0.352 {\scriptsize{± 0.006}} & 5.173 {\scriptsize{± 0.122}} \\
        NicheFlow & GLVFM & \textbf{0.212 {\scriptsize{± 0.004}}} & \textbf{5.930 {\scriptsize{± 0.003}}} & \textbf{0.342 {\scriptsize{± 0.007}}} & \textbf{5.031 {\scriptsize{± 0.003}}} \\
        \bottomrule
    \end{tabular}}
\end{table}

\begin{table}[!htbp]
    \caption{
        Comparison of Wasserstein distances between generated and real tissue structures on the axolotl brain development (ABD) dataset. We report $\mathcal{W}_1$ and $\mathcal{W}_2$ distances separately for spatial positions (Pos.) and gene expression features (Genes), averaged across all generated cell types and timepoints. All models use $\lambda = 0.1$, and results are averaged over five generation runs. Bold indicates the best (lowest) value per column.
    }
    \label{table:wasserstein-abd}
    \centering
    \resizebox{0.7\textwidth}{!}{%
    \begin{tabular}{llcccc}
        \toprule
        Model & Obj. & $\mathcal{W}_1$ Pos. & $\mathcal{W}_1$ Genes & $\mathcal{W}_2$ Pos. & $\mathcal{W}_2$ Genes \\
        \midrule
        SPFlow & CFM & 0.474 {\scriptsize{± 0.003}} & 6.711 {\scriptsize{± 0.002}} & 0.573 {\scriptsize{± 0.003}} & 6.096 {\scriptsize{± 0.081}} \\
        
        SPFlow & GVFM & 0.538 {\scriptsize{± 0.002}} & 6.681 {\scriptsize{± 0.003}} & 0.639 {\scriptsize{± 0.003}} & 6.209 {\scriptsize{± 0.099}} \\
        
        SPFlow & GLVFM & 0.489 {\scriptsize{± 0.002}} & 6.533 {\scriptsize{± 0.002}} & 0.586 {\scriptsize{± 0.003}} & 6.090 {\scriptsize{± 0.166}} \\
        \midrule
        
        RPCFlow & CFM & \textbf{0.234 {\scriptsize{± 0.003}}} & 6.363 {\scriptsize{± 0.004}} & 0.374 {\scriptsize{± 0.005}} & 6.031 {\scriptsize{± 0.089}} 
        \\
        
        RPCFlow & GVFM & 0.247 {\scriptsize{± 0.002}} & 6.093 {\scriptsize{± 0.006}} & \textbf{0.369 {\scriptsize{± 0.005}}} & 5.921 {\scriptsize{± 0.051}} \\

        RPCFlow & GLVFM & 0.254 {\scriptsize{± 0.002}} & \textbf{6.070 {\scriptsize{± 0.004}}} & 0.406 {\scriptsize{± 0.003}} & \textbf{5.884 {\scriptsize{± 0.042}}} \\
        \midrule
        
        NicheFlow & CFM & 0.267 {\scriptsize{± 0.002}} & 6.347 {\scriptsize{± 0.003}} & 0.428 {\scriptsize{± 0.005}} & 6.107 {\scriptsize{± 0.003}} \\
        
        NicheFlow & GVFM & 0.284 {\scriptsize{± 0.004}} & 6.081 {\scriptsize{± 0.004}} & 0.448 {\scriptsize{± 0.006}} & 5.917 {\scriptsize{± 0.036}} \\
    
        NicheFlow & GLVFM & 0.279 {\scriptsize{± 0.006}} & 6.085 {\scriptsize{± 0.003}} & 0.444 {\scriptsize{± 0.011}} & 5.923 {\scriptsize{± 0.034}} \\
        \bottomrule
    \end{tabular}}
\end{table}

\begin{table}[!htbp]
    \caption{
        Comparison of Wasserstein distances between generated and real tissue structures on the mouse brain aging (MBA) dataset. We compute $\mathcal{W}_1$ and $\mathcal{W}_2$ distances for spatial coordinates (Pos.) and gene expression features (Genes), averaging per cell type and timepoint. All models are trained with $\lambda = 0.1$, and results are averaged across five generation runs. Bold marks the lowest value in each column.
    }
    \label{table:wasserstein-mba}
    \centering
    \resizebox{0.7\textwidth}{!}{%
    \begin{tabular}{llcccc}
        \toprule
        Model & Obj. & $\mathcal{W}_1$ Pos. & $\mathcal{W}_1$ Genes & $\mathcal{W}_2$ Pos. & $\mathcal{W}_2$ Genes \\
        \midrule
        
        SPFlow & CFM & 0.515 {\scriptsize{± 0.001}} & 7.544 {\scriptsize{± 0.004}} & 0.600 {\scriptsize{± 0.002}} & 5.237 {\scriptsize{± 0.056}} \\

        SPFlow & GVFM & 0.564 {\scriptsize{± 0.001}} & 7.461 {\scriptsize{± 0.004}} & 0.646 {\scriptsize{± 0.001}} & 4.961 {\scriptsize{± 0.114}} \\
        
        SPFlow & GLVFM & 0.552 {\scriptsize{± 0.002}} & 7.431 {\scriptsize{± 0.003}} & 0.632 {\scriptsize{± 0.002}} & \textbf{4.117 {\scriptsize{± 0.196}}} \\
        \midrule
       
        RPCFlow & CFM & \textbf{0.385 {\scriptsize{± 0.003}}} & 7.318 {\scriptsize{± 0.006}} & \textbf{0.480 {\scriptsize{± 0.003}}} & 6.659 {\scriptsize{± 0.098}} \\
        
        RPCFlow & GVFM & 0.416 {\scriptsize{± 0.003}} & 7.202 {\scriptsize{± 0.003}} & 0.510 {\scriptsize{± 0.003}} & 6.346 {\scriptsize{± 0.061}} \\
        \midrule
        
        NicheFlow & CFM & 0.416 {\scriptsize{± 0.008}} & 7.321 {\scriptsize{± 0.012}} & 0.510 {\scriptsize{± 0.008}} & 6.487 {\scriptsize{± 0.084}} \\
        
        NicheFlow & GVFM & 0.412 {\scriptsize{± 0.004}} & \textbf{7.102 {\scriptsize{± 0.001}}} & 0.503 {\scriptsize{± 0.004}} & 6.403 {\scriptsize{± 0.087}} \\
        
        NicheFlow & GLVFM & 0.431 {\scriptsize{± 0.008}} & 7.121 {\scriptsize{± 0.008}} & 0.527 {\scriptsize{± 0.007}} & 6.298 {\scriptsize{± 0.029}} \\
        \bottomrule
    \end{tabular}}
\end{table}

We evaluate all model variants and training objectives across the three datasets (MED, ABD, MBA) and report the results in \cref{table:wasserstein-med,table:wasserstein-abd,table:wasserstein-mba}. \modelname{} with the GLVFM objective consistently achieves the lowest Wasserstein distances across nearly all dimensions in the embryonic development dataset (MED), outperforming both baseline models and alternative objectives.

In the ABD and MBA datasets, which exhibit more gradual spatial evolution or less distinct cell-type boundaries, we observe that RPCFlow sometimes matches or marginally outperforms \modelname{} in isolated metrics. However, as we discussed in \cref{appendix: stcmflow-vs-rpcflow}, RPCFlow is not a conditionally consistent model: it does not produce meaningful trajectories given a source microenvironment and lacks biological interpretability. Therefore, even when RPCFlow achieves slightly lower Wasserstein distances in aggregate, these gains are not actionable in practice, as the model cannot be used to infer biologically meaningful transitions or trace the evolution of specific spatial regions.

Taken together, Wasserstein metrics offer a complementary, global validation of the structural and semantic accuracy of \modelname{}. They reinforce the quantitative evidence already provided by Chamfer-based metrics (PSD/SPD) and Gromov-based structural evaluations (GW/FGW), confirming that our model accurately captures the tissue-scale organization of both spatial and gene expression patterns.

\section{Algorithms}
\label{appendix: algorithms}
Here, we present the algorithmic procedures underlying \modelname{}. To streamline the exposition, we first define compact notation for representing noisy microenvironments and their interpolations.

We denote a single noisy microenvironment (simplifying \cref{eq: pc_sampling}) as:
\begin{equation}
    \gM^z \sim \gN(\mathbf{0}, \mathbf{I}_{D+2})^{1 \times k} = \left\{(\vc_i^z, \vz_i) \mid \vc_i^z \sim \gN(\mathbf{0}, \mathbf{I}_2),\ \vz_i \sim \gN(\mathbf{0}, \mathbf{I}_D),\ \forall i = 1,\dots,k \right\},
\end{equation}
where $k$ denotes the number of spatial points per microenvironment, $\vc_i^z$ are 2D cell coordinates, and $\vz_i$ are associated feature vectors in $\mathbb{R}^D$.
For a collection of $N$ microenvironments, we define the corresponding set of noisy microenvironments
\begin{equation}
    \boldsymbol{\gM^z} \sim \gN(\mathbf{0}, \mathbf{I}_{D+2})^{N \times k}  = \left\{ \gM^z_i \sim \gN(\mathbf{0}, \mathbf{I}_{D+2})^{1 \times k} \mid \forall i = 1, \dots, N \right\}.
\end{equation}
Given a noisy sample $\gM^z$ and a corresponding ground-truth microenvironment $\gM^1$, we define their linear interpolation at time $t \in [0, 1]$ as:
\(\gM^t = (1 - t) \gM^z + t \gM^1\),
where the interpolation is performed element-wise across the matrix rows. For batched data, this generalizes to:
\begin{equation}
    \boldsymbol{\gM}^t = (1 - t) \boldsymbol{\gM}^z + t \boldsymbol{\gM}^1.
\end{equation}

\begin{algorithm}[H]
    \small
    \caption{\textsc{SampleAndInterpolate}}
    \label{alg:sample_and_interpolate}
    \begin{algorithmic}[1]
        \Require Number of samples $N$, feature dimension $D$, microenvironment size $k$, OT plan $\pi_{\epsilon,\lambda}^*$
        \Ensure Source $(\boldsymbol{\gM}^0)$, target $(\boldsymbol{\gM}^1)$, interpolated $(\boldsymbol{\gM}^t)$, and noisy $(\boldsymbol{\gM}^z)$ microenvironments
        \State $(\boldsymbol{\gM}^0, \boldsymbol{\gM}^1) \gets \text{Sample from $K$-Means regions}$ \Comment{Initial microenvironment pairs}
        \State $(\boldsymbol{\gM}^0, \boldsymbol{\gM}^1) \gets \pi_{\epsilon, \lambda}^*(\boldsymbol{\gM}^0, \boldsymbol{\gM}^1)$ \Comment{Resample with OT plan}
        \State $\smash{\boldsymbol{\gM}^z \sim \gN(\mathbf{0}, \mathbf{I}_{D+2})^{N \times k}}$ \Comment{Noisy initial states}
        \State $t \sim \mathcal{U}(0,1)$ \Comment{Random interpolation time}
        \State $\smash{\boldsymbol{\gM}^t \gets (1 - t)\boldsymbol{\gM}^z + t\boldsymbol{\gM}^1}$ \Comment{Linearly interpolated states}
    \end{algorithmic}
\end{algorithm}

\subsection{OT Conditional Flow Matching}
\label{appendix:otcfm}
The OT Conditional Flow Matching (OT-CFM) algorithm consists of a training and a generation phase. During training, the model learns a time-dependent vector field $u_t^{\theta}$ conditioned on a source microenvironment $\gM^0$, which maps a noisy microenvironment $\gM^z$—sampled from Gaussian noise—to its corresponding target microenvironment $\gM^1$. The supervision is provided via source-target pairs $(\gM^0, \gM^1)$ obtained through EOT over pooled microenvironment representations. At generation time, the learned vector field is integrated starting from $\gM^z$, conditioned on a given source $\gM^0$, to generate the predicted target microenvironment $\gM^1$.

We optimize the following loss:
\begin{equation}
    \label{eq:otcfm_loss}
    \gL(\boldsymbol{\gM}^0, \boldsymbol{\gM}^1, \boldsymbol{\gM}^t, \boldsymbol{\gM}^z; \theta)  = \frac{1}{2} \sum_{\substack{
            \gM^0 \in \boldsymbol{\gM}^0 \\
            \gM^z \in \boldsymbol{\gM}^z \\
            \gM^t \in \boldsymbol{\gM}^t \\
            \gM^1 \in \boldsymbol{\gM}^1
        }} 
        \left\|
            u_t^\theta(\gM^t, \gM^0)
            - (\gM^1 - \gM^z)
        \right\|_2^2
\end{equation}

The full pseudocode for both phases is provided in \cref{alg:otcfm_training,alg:otcfm_inference}.

\begin{algorithm}[H]
    \small
    \caption{OT CFM --- Training}
    \label{alg:otcfm_training}
    \begin{algorithmic}[1]
        \Require Number of samples $N$, feature dimension $D$, microenvironment size $k$, OT plan $\pi_{\epsilon,\lambda}^*$,  conditional velocity field $u_t^\theta$
        \Ensure Trained parameters $\theta$ of $u_t^\theta$
        \State $(\boldsymbol{\gM}^0, \boldsymbol{\gM}^1, \boldsymbol{\gM}^t, \boldsymbol{\gM}^z) \gets \Call{SampleAndInterpolate}{N, D, k, \pi_{\epsilon,\lambda}^*}$ \Comment{Microenvironments (\cref{alg:sample_and_interpolate})}
        \State $\theta \gets \nabla_\theta\gL(\boldsymbol{\gM}^0, \boldsymbol{\gM}^1, \boldsymbol{\gM}^t, \boldsymbol{\gM}^z; \theta)$ \Comment{Compute loss (\cref{eq:otcfm_loss}) \& update parameters \(\theta\)}
    \end{algorithmic}
    \end{algorithm}

\begin{algorithm}[H]
    \small
    \caption{OT CFM --- Generation}
    \label{alg:otcfm_inference}
    \begin{algorithmic}[1]
        \Require Source microenvironment $\gM^0$, learned conditional velocity field $u_t^\theta$
        \Ensure Generated microenvironment $\gM^1$
        \State $\smash{\gM^z \sim \gN(\mathbf{0}, \mathbf{I}_{D+2})^{1 \times k}}$ \Comment{Sample noisy sample}
        \State $\smash{\gM^1 \gets \gM^z + \int_0^1 u_t^\theta(\gM^t, \gM^0)\, \mathrm{d}t}$ \Comment{Solve ODE}
    \end{algorithmic}
\end{algorithm}

\subsection{OT Gaussian Variational Flow Matching}
\label{appendix:otgvfm}
The OT Gaussian Variational Flow Matching (OT-GVFM) algorithm adopts a variational perspective on Flow Matching. Instead of directly learning a time-dependent conditional velocity field, the model learns a factorized variational posterior $q_t^\theta(\mathcal{M}^1 \mid \mathcal{M}^t, \mathcal{M}^0)$ over target microenvironments $\mathcal{M}^1$, conditioned on an interpolated microenvironment $\mathcal{M}^t$ and a source microenvironment $\mathcal{M}^0$. The predicted velocity field is then computed as the difference between the posterior mean 
$\mu_t^\theta(\mathcal{M}^t, \mathcal{M}^0)$ and the current state $\mathcal{M}^1$.

The training objective minimizes the discrepancy between ground-truth targets and the predicted posterior means $(\bar{\vf}_t^{\theta}, \bar{\vr}_t^{\theta})$:
\begin{equation}
    \label{eq:otgvfm_loss}
    \gL(\boldsymbol{\gM}^0, \boldsymbol{\gM}^1, \boldsymbol{\gM}^t; \mu_t^\theta) = \frac{1}{2} \sum_{\substack{
        \gM^0 \in \boldsymbol{\gM}^0 \\
        \gM^t \in \boldsymbol{\gM}^t \\
        \gM^1 \in \boldsymbol{\gM}^1
    }} 
    \sum_{\substack{
        (\vc_1, \vx_1) \in \gM^1 \\
        (\bar{\vf}_t^{\theta}, \bar{\vr}_t^{\theta}) \in \mu_t^{\theta}(\gM^t, \gM^0)
    }} 
    \left(
        \| \vc_1 - \bar{\vf}_t^{\theta} \|_2^2 
        + \| \vx_1 - \bar{\vr}_t^{\theta} \|_2^2
    \right).
\end{equation}
At generation time, trajectories are generated by integrating the vector field $\mu_t^\theta(\mathcal{M}^t, \mathcal{M}^0) - \mathcal{M}^t$, starting from a noise-sampled microenvironment $\mathcal{M}^z$ and conditioned on a given source $\mathcal{M}^0$. To ensure stability near $t = 1$, the vector field is scaled by a time-dependent denominator, as in prior VFM formulations.

The pseudocode for both phases is provided in \cref{alg:otgvfm_training,alg:otgvfm_inference}.

\begin{algorithm}[H]
    \small
    \caption{OT-GVFM --- Training}
    \label{alg:otgvfm_training}
    \begin{algorithmic}[1]
        \Require Number of samples $N$, feature dimension $D$, microenvironment size $k$, OT plan $\pi_{\epsilon,\lambda}^*$, source-conditioned posterior mean predictor $\mu_t^\theta$
        \Ensure Trained parameters $\theta$ of $\mu_t^\theta$
        \State $(\boldsymbol{\gM}^0, \boldsymbol{\gM}^1, \boldsymbol{\gM}^t, \boldsymbol{\gM}^z) \gets \Call{SampleAndInterpolate}{N, D, k, \pi_{\epsilon,\lambda}^*}$ \Comment{Microenvironments (\cref{alg:sample_and_interpolate})}
        \State $\theta \gets \nabla_\theta\gL(\boldsymbol{\gM}^0, \boldsymbol{\gM}^1, \boldsymbol{\gM}^t; \mu_t^{\theta})$ \Comment{Compute loss (\cref{eq:otgvfm_loss}) \& update parameters \(\theta\)}
    \end{algorithmic}
\end{algorithm}

\begin{algorithm}[H]
    \small
    \caption{OT-GVFM --- Generation}
    \label{alg:otgvfm_inference}
    \begin{algorithmic}[1]
        \Require Source microenvironment $\gM^0$, learned source-conditioned posterior mean predictor $\mu_t^\theta$
        \Ensure Generated microenvironment $\gM^1$
        \State $\gM^z \sim \gN(\mathbf{0}, \mathbf{I}_{D+2})^{1 \times k}$ \Comment{Sample noisy sample}
        \State $\gM^1 \gets \gM^z + \int_0^1 \frac{\mu_t^\theta(\gM^t, \gM^0) - \gM^t}{1 - t + \epsilon} \mathrm{d}t$ \Comment{Solve ODE}
    \end{algorithmic}
\end{algorithm}

\subsection{\modelname{}: OT Gaussian-Laplace Variational Flow Matching}
\label{appendix:otglvfm}
\modelname{} extends OT-GVFM by modifying the variational posterior: it assumes a Gaussian distribution for gene expression features and a Laplace distribution for spatial coordinates. This change leads to a hybrid loss that combines an $L^2$ loss on gene expression and an $L^1$ loss on spatial locations:

\begin{equation}
    \label{eq:otglvfm_loss}
    \gL(\boldsymbol{\gM}^0, \boldsymbol{\gM}^1, \boldsymbol{\gM}^t; \theta) = \sum_{\substack{
        \gM^0 \in \boldsymbol{\gM}^0 \\
        \gM^t \in \boldsymbol{\gM}^t \\
        \gM^1 \in \boldsymbol{\gM}^1
    }} 
    \sum_{\substack{
        (\vc_1, \vx_1) \in \gM^1 \\
        (\bar{\vf}_t^{\theta}, \bar{\vr}_t^{\theta}) \in \mu_t^{\theta}(\gM^t, \gM^0)
    }} 
    \left(
        \| \vc_1 - \bar{\vf}_t^{\theta} \|_1 
        + \frac{1}{2} \| \vx_1 - \bar{\vr}_t^{\theta} \|_2^2
    \right)
\end{equation}
At generation time, the model integrates the velocity field defined by the difference between the predicted mean $\mu_t^\theta(\mathcal{M}^t, \mathcal{M}^0)$ and the current state $\mathcal{M}^1$, starting from noise and conditioned on the source $\mathcal{M}^0$, identical to the OT-GVFM procedure.
\section{Experimental setup}
\label{appendix: experimental-setup}

\subsection{Dataset description}\label{appendix: dataset_description}
We evaluate our model on three publicly available, time-resolved spatial transcriptomics datasets spanning development and aging processes. Each dataset provides single-cell resolution profiles with matched spatial coordinates and curated cell type annotations. A detailed summary of the organism, tissue, number of time points, cell types, total number of cells, and acquisition technology for each dataset is provided in \Cref{tab: dataset-summary}.

\begin{table}[h]
\centering
\caption{Overview of the time-resolved spatial transcriptomics datasets used in our experiments. MED: Mouse Embryonic Development, ABD: Axolotl Brain Development, MBA: Mouse Brain Aging. Each dataset varies in organism, tissue type, number of timepoints, cell types, total number of cells, and spatial transcriptomics technology.}
\label{tab: dataset-summary}
\resizebox{0.9\textwidth}{!}{%
    \begin{tabular}{lcccccc}
        \toprule
        Dataset & Organism & Tissue & Timepoints & Cell Types & Cells & Technology \\
        \midrule
        MED & Mouse & Whole embryo & 3 & 24 & 54k & Stereo-seq \\
        ABD & Axolotl & Brain & 6 & 33 & 36k & Stereo-seq \\
        MBA & Mouse & Brain & 20 & 18 & 1.5M & MERFISH \\
        \bottomrule
    \end{tabular}%
}
\end{table}

\subsection{Dataset and microenvironments preprocessing}
\label{appendix: dataset-preprocessing}
\paragraph{Dataset preprocessing.} All datasets used in our study underwent a preprocessing procedure appropriate for spatial transcriptomic analysis, involving total count normalization, logarithmic transformation, and principal component analysis (PCA). For the mouse embryonic development \citep{klein2025mapping} and axolotl brain development \citep{axolotl} datasets, total count normalization, and logarithmic transformation had already been applied; we, therefore, performed PCA ourselves on the transformed data. For the mouse brain aging dataset \citep{Sun2025aging}, we applied all three steps: we first normalized raw count matrices so that each cell had the same total expression, followed by a natural logarithm transformation of the form $\log(X + 1)$ to stabilize variance and mitigate the influence of large values. We then computed PCA on the log-transformed data.

To reduce computational overhead due to high cell counts in the aging dataset, we subsampled the data by a factor of 0.2.

Finally, we standardized PCA components across all cells to ensure consistent scaling across time points. Spatial coordinates were standardized independently for each time point by subtracting the per-time-point mean and dividing by the standard deviation. This standardization preserves the relative spatial configuration while accounting for scale and position differences over developmental time.

\paragraph{Microenvironments preprocessing.} To facilitate efficient training and enable consistent microenvironment construction, we precompute all fixed-radius neighborhoods for each dataset using a radius $r$ chosen based on spatial resolution. To reduce variability in the number of neighbors and improve batching efficiency, we fix the number of nodes per microenvironment to the most frequent neighbor count observed within each slide. This standardization ensures structural comparability across microenvironments while significantly reducing computational overhead during training, as costly radius or nearest-neighbor queries are avoided at runtime.

\subsection{Quantitative evaluation metrics}\label{app: quantitative_evaluation_metrics}
We assess spatial fidelity using two complementary asymmetric distance measures. The \textit{point-to-shape distance} (PSD) captures how much predicted cell positions diverge from the actual tissue layout, computed as the average squared distance from each simulated point to its nearest neighbor in the ground truth. Conversely, the \textit{shape-to-point distance} (SPD) quantifies how comprehensively the predicted distribution spans the target structure by averaging the squared distance from each ground truth point to its closest generated counterpart.

Let $ \mathcal{G}^t $ denote the set of generated coordinates and $ \mathcal{R}^t $ the set of ground truth coordinates at time $t$. Define $ \text{NN}_{\text{ref}}^t(\vc_i) $ as the nearest neighbor of $ \vc_i \in \mathcal{G}^t $ in $ \mathcal{R}^t $, and $ \text{NN}_{\text{gen}}^t(\vc_i) $ as the nearest neighbor of $ \vc_j \in \mathcal{R}^t $ in $ \mathcal{G}^t $. Then, the two metrics are given by:
\begin{equation}
    \text{PSD} = \frac{1}{|\mathcal{G}|} \sum_{\mathcal{G}^t \in \mathcal{G}} \sum_{\vc_i \in \mathcal{G}^t} \| \vc_i - \text{NN}_{\text{ref}}^t(\vc_i) \|_2^2,
\end{equation}
\begin{equation}
    \text{SPD} = \frac{1}{|\mathcal{R}|} \sum_{\mathcal{R}^t \in \mathcal{R}} \sum_{\vc_j \in \mathcal{R}^t} \| \vc_j - \text{NN}^t_{\text{gen}}(\vc_j) \|_2^2.
\end{equation}
where $ \mathcal{G} := \cup_{t \in \mathcal{T}} \mathcal{G}^t $ and $ \mathcal{R} := \cup_{t \in \mathcal{T}} \mathcal{R}^t $.

\subsection{Cell-type classification for evaluation}
\label{appendix: cell-type-classifier}
To evaluate cell-type fidelity of generated microenvironments, we train a supervised classifier to assign cell type labels based on gene expression profiles. We apply the same preprocessing steps used for training our generative models: total count normalization, log-transformation, and PCA (see \cref{appendix: dataset-preprocessing}). The resulting low-dimensional embeddings are used as input features for a simple multilayer perceptron (MLP), trained to predict discrete cell type labels.

The classifier consists of a two-layer feedforward network with ReLU activations and a final linear projection to the number of cell types. It is trained using cross-entropy loss and optimized with the AdamW optimizer. We report performance using the weighted F1-score.

We obtain strong classification results across all datasets. On the mouse embryonic development dataset, the classifier achieves a weighted F1-score of 0.85; on axolotl brain development, 0.80; and on the aging dataset, 0.97. These results correlate with the number of input genes and the variance retained in the PCA-reduced space. The aging dataset contains only 300 genes, and 50 principal components explain sufficient variance to accurately distinguish most cell types. In contrast, the embryonic development dataset contains approximately 2,000 genes, and the axolotl brain development dataset includes over 12,700 genes, making classification more challenging due to higher gene expression variability.

\subsection{Discretized microenvironments}
\label{appendix: discretized-microenvironments}

To ensure consistent and reproducible evaluation across methods and datasets, we construct a fixed set of evaluation microenvironments by discretizing the spatial domain of each tissue section. For each time point, we define a regular 2D grid over the tissue and select the closest cell to each grid point as the centroid of a microenvironment. Each centroid is then used to construct a fixed-radius neighborhood, following the microenvironment definition in Section~\ref{sec: single_cell_microenv}. This procedure ensures full spatial coverage by verifying that every cell belongs to at least one microenvironment.

\begin{figure}[!htb]
    \centering
    \includegraphics[width=0.9\linewidth]{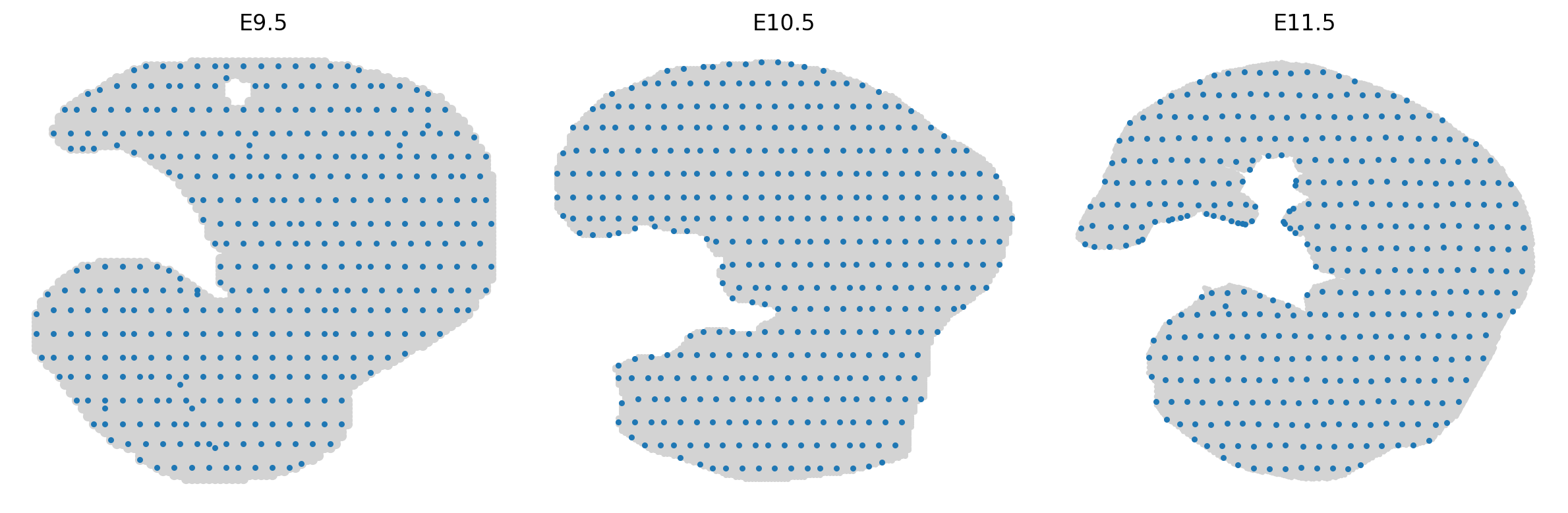}
    \caption{
        Discretized grid of microenvironments for the mouse embryonic development dataset. Each blue point denotes a centroid around which a microenvironment is constructed. To ensure consistent coverage across tissue sections, additional centroids are randomly sampled such that each slide contains the same number of microenvironments.
    }
    \label{fig: discretized-grid}
\end{figure}

Figure~\ref{fig: discretized-grid} illustrates this discretization for the mouse embryonic development dataset. Each blue dot corresponds to a selected centroid. In cases where the number of grid-based centroids falls below a target threshold, additional centroids are randomly sampled to match a fixed total count per slide. This augmentation ensures that all slides contribute equally to the evaluation and prevents bias from sparse regions.

We apply the same discretization procedure to all three datasets used in our experiments: mouse embryogenesis, axolotl brain development, and mouse brain aging. By standardizing the evaluation regions spatially and deterministically, we eliminate the need for stochastic region sampling during evaluation, which would otherwise lead to nondeterministic and irreproducible results.

\subsection{Microenvironment Transformer}
\label{appendix: microenvironment-transformer}

To model the spatiotemporal evolution of local cellular neighborhoods, we design a permutation-invariant transformer-based architecture tailored to structured point cloud data. Our Microenvironment Transformer processes local microenvironments---sets of cells with gene expression features and spatial coordinates---and predicts time-dependent outputs such as velocity fields or future states.

The model operates on a source $\gM^0$ and noisy $\gM^z$ microenvironments with per-cell features $\vx_i \in \mathbb{R}^D$ and 2D coordinates $\vc_i \in \mathbb{R}^2$. The architecture consists of the following components:

\begin{enumerate}
    \item \textbf{Input Embeddings:}
    \begin{enumerate}
        \item \textbf{Feature Embedding:} A linear transformation is applied to the input features $\vx_i$ of each cell.
        \item \textbf{Coordinate Embedding:} Spatial coordinates $\vc_i$ are linearly projected and concatenated to the feature embedding.
        \item \textbf{Time Embedding:} For the noisy microenvironment only, time $t \in [0,1]$ is encoded using sinusoidal functions $\cos(\omega t)$ and $\sin(\omega t)$, followed by a linear projection and concatenation with the input embedding.
    \end{enumerate}

    \item \textbf{Transformer Encoder (Source Microenvironment):}
    \begin{enumerate}
        \item \textbf{Self-Attention:} A stack of transformer encoder blocks with multi-head self-attention processes the embedded source microenvironment.
        \item \textbf{No Time Embedding:} Time information is \emph{not} provided to the encoder, as it encodes the source $\mathcal{M}^0$.
        \item \textbf{Residual Feedforward:} Each block contains a feedforward subnetwork with LeakyReLU activation and a residual connection.
        \item \textbf{Layer Normalization:} Applied after both the attention and feedforward layers.
        \item \textbf{Masking:} Binary masks are used to ignore padding in variable-length microenvironments.
    \end{enumerate}

    \item \textbf{Transformer Decoder (Noisy Microenvironment):}
    \begin{enumerate}
        \item \textbf{Time Embedding:} Temporal context is injected into the decoder by embedding the time $t$ and concatenating it to the target point embedding.
        \item \textbf{Cross-Attention:} Decoder layers apply cross-attention between the noisy microenvironment and the encoded source microenvironment.
        \item \textbf{Self-Attention and Feedforward:} Each decoder block includes standard self-attention and residual feedforward layers.
        \item \textbf{Layer Normalization and Masking:} As with the encoder, normalization, and masking are applied throughout.
    \end{enumerate}

    \item \textbf{Final Output Projection:}
    \begin{enumerate}
        \item \textbf{Prediction Head:} A linear layer maps the decoder outputs to the desired dimensionality.
    \end{enumerate}
\end{enumerate}

This architecture allows for flexible and expressive modeling of temporal dynamics in cellular point clouds. By encoding only the source and decoding the temporally conditioned target, the model supports variational and flow-based training objectives with explicit temporal conditioning.

\subsection{Hyperparameters and Computational Costs}
\label{appendix: hyperparameters-comp-costs}

\paragraph{Model hyperparameters.} For all experiments, we use the same configuration for the Microenvironment Transformer architecture. The full set of hyperparameters is as follows:

\begin{itemize}
    \item \textbf{Input feature dimension:} 50 PCA-based gene expression features concatenated with a one-hot encoding of the time-point, resulting in a total dimensionality of $50 + |\mathcal{T}|$, where $|\mathcal{T}|$ is the number of slides (timepoints) in the dataset.
    \item \textbf{Input coordinate dimension:} 2
    \item \textbf{Embedding dimension:} 128
    \item \textbf{MLP hidden dimension:} 256
    \item \textbf{Number of attention heads:} 4
    \item \textbf{Number of encoder layers:} 2
    \item \textbf{Number of decoder layers:} 2
    \item \textbf{Dropout rate:} 0.1
    \item \textbf{Output dimension:} 52 (gene expressions features + coordinates)
\end{itemize}

\paragraph{OT and mini-batching.} To ensure spatial diversity and computational tractability, we uniformly sample 256 source–target microenvironment pairs from the $K$ spatial clusters obtained via $K$-Means. We then compute the entropic OT plan between these sampled pairs and resample 64 source-target pairs from this plan to define a single training instance. During training, we process 16 such instances per batch.
\paragraph{Optimization.} All models are trained using the AdamW optimizer with a learning rate of $2 \cdot 10^{-4}$ and a weight decay of $1 \cdot 10^{-5}$. We train each model until convergence.
\paragraph{Computational cost.} All models were trained on a single NVIDIA GeForce GTX 1080 Ti GPU with 11GB of memory. Depending on the dataset and training objective (e.g., CFM or VFM), training takes approximately 12--16 hours per model.

\subsection{Comparison with moscot}\label{app: biological experiments}

Here, we describe how we conduct the comparisons with moscot, as shown in \cref{fig: bio_analysis}, \cref{fig: jaw_teeth_double_push}, and \cref{fig: liver_double_push}. For both \modelname{} and moscot, we select a source microenvironment that we want to track over time. In the case of \modelname{}, this corresponds to an aggregate of point clouds. For moscot, it refers to a group of single cells spatially located within the region of interest. The same set of cells is used for both methods.

\textbf{NicheFlow.} To generate contour plots over the spatial slide, we push forward the selected region and assign the generated points to their nearest real neighbors based on spatial coordinates. We then compute a probability value for each real position by normalizing the number of assigned generated points. In other words, the more generated points that are close to a given real point, the higher the probability assigned to that location in the contour plot. Cell type proportions are computed as the aggregated frequencies of the generated cell types across the slide. Each plot considers 10 independent generation runs from the same niche. 

\textbf{moscot.} This baseline is not a generative model, but rather a standard discrete OT framework using a Fused Gromov-Wasserstein cost. As such, it does not generate new features or coordinates. Instead, it outputs a transition matrix that assigns matching probabilities between each source slide cell and each target slide cell. We use these transition probabilities to compute contour plots over the target slide and to aggregate cell type probabilities for the bar plots. For the latter, moscot provides a custom method called \texttt{cell\_transition()}. 

For the Appendix figures \cref{fig: jaw_teeth_double_push},\cref{fig: axolotl-anatomical-middle-sep}, \cref{fig: axolotl-anatomical-cavity}, and \cref{fig: axolotl-anatomical-left-dorsal}, we propagate the initial source region across multiple time steps. In \modelname{}, this is achieved by using the simulated point cloud from step $t$ to predict the next state at $t+1$, and then feeding this output as the input for the following trajectory step from $t+1$ to $t+2$, and so on. Ground truth points are not used as intermediate sources during this process. For moscot, pushing the source across time points can be automatically done by setting non-subsequent time points in the \texttt{cell\_transition()} function.

\end{document}